\documentclass[lettersize,journal]{IEEEtran}
\usepackage{amsmath,amsfonts}
\usepackage{algorithmic}
\usepackage{algorithm}
\usepackage{array}
\usepackage{textcomp}
\usepackage{stfloats}
\usepackage{verbatim}
\usepackage{graphics} 
\usepackage{epsfig} 
\usepackage{amsmath} 
\usepackage{amssymb}  
\usepackage{times} 
\usepackage{cite} 
\usepackage{graphicx}
\usepackage{subfigure} 
\usepackage{picinpar}
\usepackage{url}
\usepackage{flushend}
\usepackage{colortbl}
\usepackage{soul}
\usepackage{color}
\soulregister{\cite}7 
\soulregister{\citep}7 
\soulregister{\citet}7 
\soulregister{\ref}7 
\soulregister{\pageref}7 

\usepackage{multirow}
\usepackage{pifont}
\usepackage{alltt}
\usepackage{hyperref}
\usepackage{float}
\usepackage{enumerate}
\usepackage{siunitx}
\usepackage{breakurl}
\usepackage{epstopdf}
\usepackage{pbox}
\usepackage{booktabs}
\usepackage{makecell} 
\usepackage{threeparttable} 
\usepackage{balance}
\usepackage[justification=centering,labelsep=period]{caption} 
\usepackage{color} 
\usepackage{gensymb} 
\hypersetup{colorlinks=true, linkcolor=black} 

\usepackage{adjustbox}

\usepackage{bm}
\usepackage{geometry}
\title{\LARGE \bf
    EVI-SAM: Robust, Real-time, Tightly-coupled Event-Visual-Inertial State Estimation and 3D Dense Mapping 
}

\let\oldtwocolumn\twocolumn
\renewcommand\twocolumn[1][]{%
    \oldtwocolumn[{#1}{
    \begin{center}
           \captionsetup{justification=justified}
           \includegraphics[width=\textwidth]{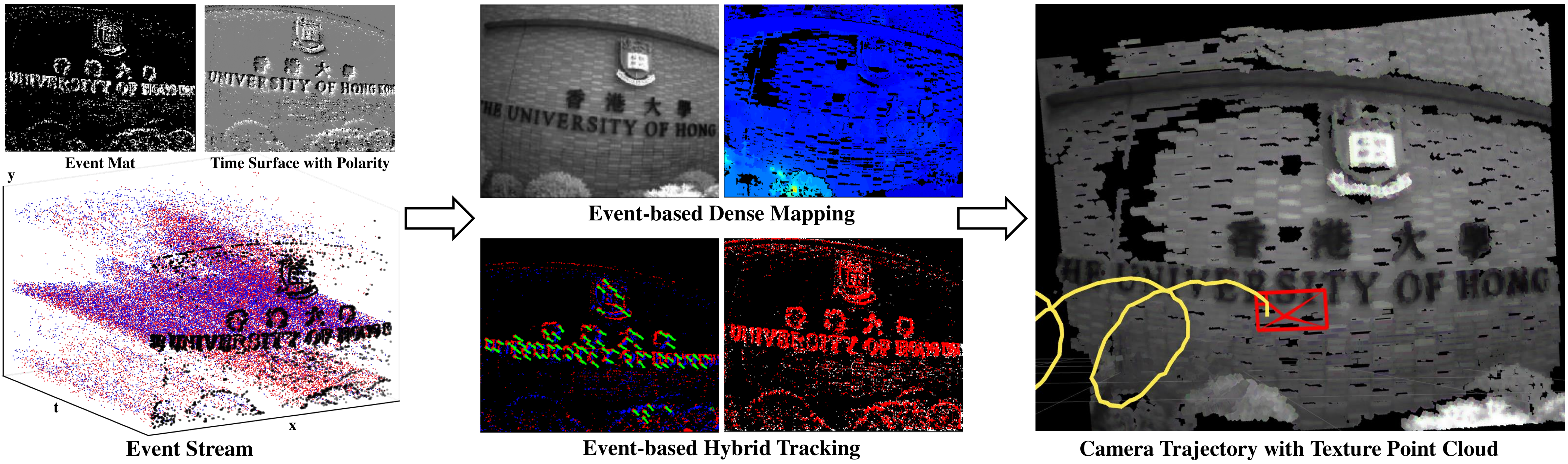}
           \captionof{figure}{         
          Our EVI-SAM enables the recovery of both the camera pose and dense maps of the scene. 
          The tracking module leverages the re-projection constraint from the feature-based method and the relative pose constraints from the direct-based method within an event-based hybrid tracking framework.
          The mapping module represents a pioneering event-based dense mapping framework, distinguished as a non-learning method that can conduct real-time dense and texture mapping on a standard CPU.
           }
           \label{fig:fig1}
        \end{center}
    }]
}

\author{Weipeng~Guan,   Peiyu~Chen,   Huibin~Zhao,   Yu~Wang,   Peng~Lu$^{*}$     
    \thanks{
        $^{*}$corresponding author;
    
        The authors are with the Adaptive Robotic Controls Lab (ArcLab), Department of Mechanical Engineering, The University of Hong Kong, Hong Kong SAR, China.
        E-mail: \{wpguan, chenpyhk, hbzhao, ywang812\}@connect.hku.hk, lupeng@hku.hk.
        }%
    \thanks{
        This work was supported by General Research Fund under Grant 17204222, and in part by the Seed Fund for Collaborative Research and General Funding Scheme-HKU-TCL Joint Research Center for Artificial Intelligence.
        }%
}

\begin{document}   
\maketitle  
\thispagestyle{headings} 
\pagestyle{headings}

\begin{abstract}
Event cameras are bio-inspired, motion-activated sensors that demonstrate substantial potential in handling challenging situations, such as motion blur and high-dynamic range.
In this paper, we introduce EVI-SAM to tackle the problem of 6-DoF pose tracking and 3D dense reconstruction using the monocular event camera.
A novel event-based hybrid tracking framework is designed to estimate the pose, leveraging the robustness of feature matching and the precision of direct alignment.
Specifically, we develop an event-based 2D-2D alignment to construct the photometric constraint, and tightly integrate it with the event-based re-projection constraint. 
The mapping module recovers the dense and colorful depth of the scene through the image-guided event-based mapping method.
Subsequently, the appearance, texture, and surface mesh of the 3D scene can be reconstructed by fusing the dense depth map from multiple viewpoints using truncated signed distance function (TSDF) fusion.
To the best of our knowledge, this is the first non-learning work to realize event-based dense mapping. 
Numerical evaluations are performed on both publicly available and self-collected datasets, which qualitatively and quantitatively demonstrate the superior performance of our method.
Our EVI-SAM effectively balances accuracy and robustness while maintaining computational efficiency, showcasing superior pose tracking and dense mapping performance in challenging scenarios.
\textbf{Video Demo}: \url{https://youtu.be/Nn40U4e5Si8}.
\end{abstract}

\begin{IEEEkeywords} 
Event-based Vision, 6-DoF Pose Tracking, Event Camera, SLAM, Robotics.
\end{IEEEkeywords} 

\section{INTRODUCTION}
\label{INTRODUCTION}

\subsection{Motivations} 
\label{Motivations}
\IEEEPARstart{E}{vent} camera captures high-quality data either in extreme lighting scenes or high-speed motion, which paves the way to tackle challenging scenarios in robotics such as Visual Odometry (VO), Visual Inertial Odometry (VIO), Simultaneous Localization and Mapping (SLAM), etc.
This novel sensor offers many advantages over standard cameras including high temporal resolutions ($\mu$s-level), high dynamic range (HDR, 140 dB), low power consumption, and immunity to motion blur~\cite{GWPHKU:EVENT-SURVEY}. 
However, the event camera exclusively responds to moving edges within the scene, inherently resulting in sparse and asynchronous output.
This poses a challenge in the estimation of dense depth and the recovery of detailed structures for objects and environments, especially in low-contrast regions where events may not be triggered.

Previous works in event-based depth estimation, such as EMVS~\cite{GWPHKU:EMVS}, Refs.~\cite{zhou2018semi} and~\cite{ghosh2022multi}, have shown impressive performance in sparse or semi-dense mapping, while event-based dense mapping using non-learning methods remains a research gap.
Learning-based approaches for event-based dense mapping are present in~\cite{zhu2019unsupervised, EventNeRF, tulyakov2019learning, nam2022stereo, ahmed2021deep, mostafavi2021event}, which rely on extensive training data and provide no guarantees of optimality or generality.
Additionally, these approaches only produce local depth maps and are incapable of reconstructing a globally consistent depth map.
How to effectively exploit the sparse spatial information and abundant temporal cues from asynchronous events to generate dense depth maps remains an unsolved problem.
This motivates us to investigate the complementarity between events and images, exploring the event-based dense mapping approach for inferring dense depth from sparse and incomplete event measurements.


Meanwhile, event-based pose tracking has also gained significant research interest for challenging scenarios where the performance of traditional cameras may be compromised such as under extreme lighting conditions and very fast motion.
Prior studies on event-based pose tracking using direct-based methods (such as EVO~\cite{GWPHKU:EVO}, ESVO~\cite{GWPHKU:ESVO}, EDS~\cite{EDS}) as well as feature-based methods (such as Ultimate-SLAM~\cite{GWPHKU:Ultimate-SLAM}, PL-EVIO~\cite{GWPHKU:PL-EVIO}, ESVIO~\cite{ESVIO}) has made significant progress in state estimation.
Both feature-based and direct-based methods have their strengths and weaknesses in their respective aspects, contributing to mutual complementarity to some degree.
Exploring how to better leverage or integrate the respective advantages of feature-based methods and direct-based methods for event-based pose tracking remains an uncharted area.
This motivates us to combine feature-based and direct methods for a more robust and accurate event-based pose tracker. 

\subsection{Contributions} 
\label{Contributions}
In this work, we introduce the EVI-SAM, a comprehensive SLAM system that integrates events, images, and IMU data seamlessly.
To this end, an event-based hybrid tracking pipeline is designed to enable accurate 6-DoF pose tracking and an event-based dense mapping method is proposed to reconstruct the dense depth of the surrounding environment.
To the best of our knowledge, this is the first framework that employs a non-learning approach to achieve event-based dense and textured 3D reconstruction without GPU acceleration.
Additionally, it is also the first hybrid approach that integrates both photometric and geometric errors within an event-based framework.
Our contributions can be summarized as follows:

\begin{enumerate}

\item 
We present a novel direct event-based pose tracking scheme that employs event-based 2D-2D alignment for photometric optimization. 

\item 
We introduce an innovative event-based hybrid tracking pipeline.
This pipeline efficiently constructs a nonlinear graph optimization framework to jointly incorporate the re-projection constraint from the feature-based method and the relative pose constraints from the direct-based method.

\item 
We propose an event-based dense mapping method. 
It involves segmenting the event-based semi-dense depth map with image guidance and introduces a novel interpolation method to recover dense depth within each segment. 
Additionally, we also leverage image measurements to render texture to the map (i.e., the color of 3D points). 

\item 
We design a TSDF-based map fusion method to integrate local event-based depth, producing a precise and globally consistent 3D map that includes appearance and surface mesh.

\item 
Extensive experimental evaluations, both qualitative and quantitative, demonstrate the accuracy, robustness, generalization ability, and outstanding performance of our approach.

\end{enumerate}

The remainder of the paper is organized as follows:
Section~\ref{Related Works} introduces the related works on event-based pose tracking and depth estimation.
Section~\ref{System Overview} provides a brief overview of our EVI-SAM system, with detailed descriptions of its event-based hybrid tracking and event-based dense mapping presented in Sections~\ref{6-Dof Pose Tracking} and~\ref{3D Mapping}, respectively.
Section~\ref{Evaluation} presents the evaluations.
Finally, the conclusion is given in Section~\ref{CONCLUSIONS}.

\section{Related Works}
\label{Related Works}

\subsection{Event-based Pose Tracking}

Event-based pose tracking can be classified into two primary categories: 
\textit{(i) feature/indirect-based} methods~\cite{GWPHKU:Ultimate-SLAM,GWPHKU:PL-EVIO, ESVIO}, that extract a sparse set of repeatable event-based features from the event data, and then recover the pose and scene geometry through minimizing the re-projection errors based on these feature associations; 
and \textit{(ii) direct-based} methods~\cite{GWPHKU:EVO, GWPHKU:ESVO, EDS}, that directly minimize photometric errors in different event representations without explicit data association. 

\subsubsection{Feature-based Event Pose Tracking}

The first feature-based event poses tracking method was proposed in Ref.~\cite{GWPHKU:Event-based-visual-inertial-odometry} which fuses events with IMU through the Extended Kalman Filter.
After that, nonlinear optimization employing feature-based methods, EIO (event and IMU odometry) and EVIO (event, image, IMU odometry), were introduced in Ref.~\cite{GWPHKU:ETH-EVIO} and Ultimate-SLAM~\cite{GWPHKU:Ultimate-SLAM}, where the image-like event frames were developed to leverage traditional image-based feature detection and tracking. 
Ref.~\cite{GWPHKU:Continuous-time-visual-inertial-odometry-for-event-cameras} introduced an EIO approach by addressing the re-projection errors from the asynchronous events and directly incorporating IMU within a continuous-time framework.
Ref.~\cite{EKLT-VIO} combined the event-based and image-based feature tracker~\cite{GWPHKU:EKLT} as the front end with a filter-based back end.
Ref.~\cite{GWPHKU:MyEVIO} presented a monocular feature-based EIO that employs event-corner features to provide real-time 6-DoF state estimation.
Ref.~\cite{IROS2022_EVIO} extracted features from events-only data and associated it with a spatio-temporal locality scheme based on exponential decay.
Ref.~\cite{liu2022asynchronous} performed the event-only VO by optimizing the camera poses and 3D feature positions jointly.
ESVIO~\cite{ESVIO} proposed the first stereo EIO and EVIO framework to estimate states through temporally and spatially event-corner feature association.
IDOL~\cite{GWPHKU:IDOL}, Refs.~\cite{chamorro2023event} and~\cite{chamorro2022event} explored the event-based line feature in state estimation.
PL-EVIO~\cite{GWPHKU:PL-EVIO} integrated event-based point and line features to perform robust pose estimations, which can be used as onboard pose feedback for the quadrotor to achieve aggressive flip motion.

\subsubsection{Direct-based Event Pose Tracking}

Ref.~\cite{GWPHKU:kim2016real} presented an event-based direct method that leverages photometric relationships between brightness changes and absolute
brightness intensity to associate events with the corresponding pixels in the reference image.
EVO~\cite{GWPHKU:EVO} performed an event-based tracking approach relying on edge-map (2D-3D) model alignment, utilizing the 3D map reconstructed from EMVS\cite{GWPHKU:EMVS}. 
Refs.~\cite{gallego2017event} and~\cite{bryner2019event} presented direct-based tracking of an event camera from a given photometric depth map.
ESVO~\cite{GWPHKU:ESVO} is the first stereo event-based VO method, which follows a parallel tracking and mapping scheme to estimate the ego motion through the 2D-3D edge registration on time surface (TS). 
Building upon the framework of ESVO, DEVO~\cite{DEVO} introduced a direct VO framework that incorporates a depth camera to supply depth information for the event camera.
EDS~\cite{EDS} proposed an event-image alignment algorithm that minimizes the photometric errors between the brightness change from events and the image gradients, enabling 6-DOF pose tracking.
However, most of the aforementioned direct-based event pose tracking methods are limited to small-scale environments and small, dedicated movements.
They struggle to provide reliable state estimations, as they heavily depend on successful direct model alignment and the timely updates of the local 3D map.
While the integration of both direct-based and feature-based methods in event-based pose tracking is still a research gap.

\subsection{Event-based Mapping}

\subsubsection{Monocular Depth Estimation}

Ref.~\cite{GWPHKU:kim2016real} introduced the pioneering concept of purely event-based depth estimation by employing three decoupled probabilistic filters. 
EMVS~\cite{GWPHKU:EMVS} is the first work to achieve semi-dense 3D reconstruction from a single event camera with a known trajectory without requiring any explicit data association or intensity estimation.
The concept of event-based denser mapping was introduced in Ref.~\cite{dong2021standard}, building upon EMVS~\cite{GWPHKU:EMVS}. 
However, it is still in the conceptual stage and does not successfully reconstruct any event-based dense point cloud.
EOMVS~\cite{EOMVS} adopted an omnidirectional event camera in EMVS~\cite{GWPHKU:EMVS} to reconstruct a wider field-of-view semi-dense depth.
Ref.~\cite{gallego2018unifying} used contrast maximization to find the best depth value that fits the event stream.
These methods recover semi-dense 3D reconstructions of scenes by integrating events from a moving camera over a time interval, and they require knowledge of camera motion.
Ref.~\cite{chiavazza2023low} calculated semi-dense depth from optical flow using neuromorphic hardware to process asynchronous events. 
However, its applicability is confined to translational motion only.
Besides, various learning-based methods have gained popularity in tackling the monocular event depth estimation problem, such as convolutional neural network (CNN)~\cite{zhu2019unsupervised, chaney2019learning, hidalgo2020learning}, recurrent neural networks (RNN)~\cite{gehrig2021combining}, and Neural Radiance Field (NeRF)~\cite{EventNeRF,Ev-NeRF,E-nerf, mahbub2023multimodal}.
However, the exploration of non-learning approaches for the monocular event camera to recover dense and textured 3D structures remains an uncharted research area.

\subsubsection{Stereo Depth Estimation}

Ref.~\cite{ieng2018neuromorphic} followed a paradigm of event matching plus triangulation to realize the 3D reconstruction.
Ref.~\cite{zhou2018semi} and ESVO~\cite{GWPHKU:ESVO} tackled a semi-dense reconstruction problem using a pair of temporally-synchronized event cameras in stereo configuration through energy minimization methods.
T-ESVO~\cite{T-ESVO} extended the ESVO~\cite{GWPHKU:ESVO} using TSDF to reconstruct 3D environments and re-estimated the semi-dense depth.
Refs.~\cite{ghosh2022event} and~\cite{ghosh2022multi} extended the EMVS~\cite{GWPHKU:EMVS} into stereo setup to estimate depth by fusing back-projected ray densities.
Meanwhile, learning-based methods, such as Refs.~\cite{tulyakov2019learning, nam2022stereo, ahmed2021deep}, have been applied to stereo event-based depth estimation, where different deep networks were developed to reconstruct event-based dense maps.
Although these learning-based methods can predict dense depth and reveal 3D structures with limited events, they may struggle to handle objects that were not included in training sets, thus leading to uncertainties in their generalization ability.
Furthermore, these learning-based methods only generate localized depth maps. The production of globally consistent depth maps with rich texture information, such as surface mesh, remains an unexplored territory in the field of event-based vision.

\section{System Overview} 
\label{System Overview}

\begin{figure*}[htb]  
        \captionsetup{justification=justified}
        \centering
        \includegraphics[width=2.0\columnwidth]{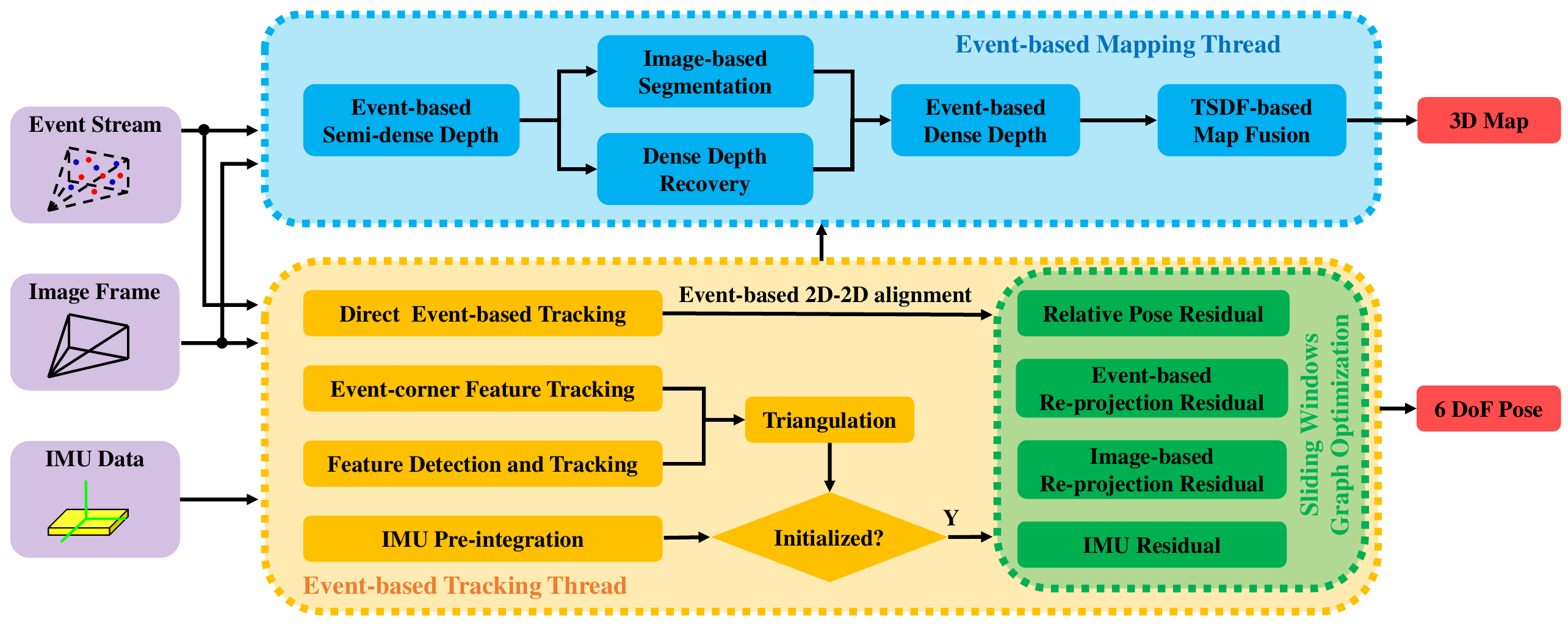}
        \caption{
        System overview. 
        The EVI-SAM algorithm takes events, images, and IMU as inputs, enabling the recovery of both camera pose and dense map of the scene.
        The mapping process takes raw event streams as input, using images for guidance, and produces dense and textured 3D mapping as output.
        The tracking thread takes event, image, and IMU as input, and constructs the feature-based and direct-based constraints to estimate the 6-DoF pose.
        }  
        \label{overview}
\end{figure*}%

The proposed EVI-SAM utilizes inputs from the monocular event camera, including events, images, and IMU, to simultaneously estimate the 6-DoF pose and reconstruct the 3D dense maps of the environment.
The overview of our EVI-SAM pipeline is illustrated in Fig.~\ref{overview}.
Our EVI-SAM consists of tracking (Section~\ref{6-Dof Pose Tracking}) and mapping (Section~\ref{3D Mapping}) modules, which operate in two parallel threads.

The tracking module employs a hybrid framework that combines feature-based and direct-based methods to process events, enabling the estimation of 6-DoF pose.
A sliding window graph-based optimization framework is designed to tightly fuse the event-based geometric errors (re-projection residuals) and event-based photometric errors (relative pose residuals), along with the image-based geometric errors and the IMU pre-integration.

Concurrently, the mapping module initially reconstructs the event-based semi-dense depth using a space-sweep way through the 6-DoF pose obtained from the tracking module. 
Subsequently, it integrates an aligned intensity image as guidance to reconstruct the event-based dense depth and render the texture of the map (i.e., the color of 3D points). 
Finally, the TSDF-based map fusion is designed to generate a 3D global consistent texture map and surface mesh of the environment.

\subsection{Event Representations} 
\label{Event Representations}
Event cameras are motion-activated sensors that capture pixel-wise intensity differences and report them as a continuous stream, rather than capturing the entire scene as an intensity image frame.
An event can be triggered either by moving objects or the ego-motion of the camera and when a large enough intensity change exceeds the pre-defined threshold $T_{threshold}$, it can be represented as follows: 
\begin{equation}
\boldsymbol{e}=\left\{t, u, v, p \right\} \Leftrightarrow \boldsymbol{I}(u, v, t+\bigtriangleup t)-\boldsymbol{I}(u, v, t) \geq p \cdot T_{threshold}
\end{equation}
where $ t $ is the timestamp that the intensity of a pixel $ \boldsymbol{I}\left(u, v \right)$ changes, and $ p $ is the polarity that indicates the direction of the intensity change.
Since a single event lacks sufficient information, it is common to aggregate sets of events within time intervals into synchronous data representations, rather than processing individual events asynchronously.
In the EVI-SAM, the mapping module takes the raw event streams as input, while the tracking module employs two distinct event representations (as shown in the left part of Fig.~\ref{fig:fig1}) as intermediate variables to facilitate the tracking between adjacent observations: the time surface (TS) with polarity (Eq.~\eqref{eq:TS_p}) for the event-corner feature tracking, and the event mat (Eq.~\eqref{eq:event_mat}) for event-based 2D-2D alignment.

The TS with polarity can encode the spatio-temporal constraints of the historical event stream at any given instant. 
Using an exponential decay kernel, it can emphasize recent events compared to past events. 
Assuming $t_{last}(\boldsymbol{x})$ as the timestamp of the last event at each pixel coordinate $\boldsymbol{x}=(u, v)$, the TS with polarity at time $t \geq t_{last}(\boldsymbol{x})$ can be defined by:
\begin{equation}
\boldsymbol{T}_{p}(\boldsymbol{x}, t)=p \cdot \exp(-\frac{t-t_{last}(\boldsymbol{x})}{\eta}) 
\label{eq:TS_p}
\end{equation}
where $\eta$ is the decay rate kernel.

The event mat is generated by aggregating a group of events that occur within a temporal neighborhood onto the event-accumulated image. 
In this process, pixel values are set to 255.0 when events are triggered; otherwise, they are set to zero, as follows:
\begin{equation}
\boldsymbol{E}_{t}=\sum_{t=t_{0}}^{t=t_{0}+\Delta t} \{ e|t_{0} \leq t \leq t_{0}+\Delta t\}
\label{eq:event_mat}
\end{equation}
where $\Delta t$ is the constant temporal window of the observed events.
While the temporal window length for event processing remains consistent with the frequency of the input event stream.

\subsection{Initialization}
The event-based mapping thread and the direct event-based tracking are initiated after the successful bootstrap of the feature-based EVIO.
Adopting from~\cite{GWPHKU:VINS-MONO,GWPHKU:VINS-MONO-initialization}, the feature-based EVIO pipeline commences with a vision-only structure from motion (SfM) to establish the up-to-scale structure of camera pose and event-corner feature positions.
By loosely aligning the SfM with the pre-integrated IMU measurements, it can bootstrap the system from unknown initial states.
\section{Event-based Hybrid Pose Tracking} 
\label{6-Dof Pose Tracking}
Our proposed event-based hybrid tracking module tightly integrates feature-based EVIO and event-based direct alignment within a graph-based nonlinear optimization framework.
The feature-based EVIO (detailed in Section~\ref{Feature-based EVIO Pose Tracking}) consists of event-based landmarks, image-based landmarks, and IMU pre-integration.
The event-based direct alignment (detailed in Section~\ref{Event-based Direct Pose Tracking}) takes continuous event streams as input and calculates the photometric errors at the pixel level to establish relative pose constraints.

As for the feature-based EVIO (more details are available in~\cite{GWPHKU:MyEVIO,GWPHKU:PL-EVIO}), when a new event stream arrives, the first step is to track the existing event-corner features using optical flow on the TS with polarity denoted as $\boldsymbol{T}_{p}(\boldsymbol{x},t)$ (see Eq.~\eqref{eq:TS_p}). Any features that cannot be successfully tracked at the current timestamp are discarded.
Subsequently, new event-corners are detected from the incoming event stream whenever the number of tracked features falls below a certain threshold. 
Meanwhile, the TS with polarity $\boldsymbol{T}_{p}(\boldsymbol{x},t)$ is utilized as a mask to ensure the even distribution of event-corner features.
After undistortion, outliers filtering, and triangulation, the event-based landmark, for which the 3D position has been successfully calculated, is integrated to establish the re-projection constraints.

Regarding the event-based direct alignment, direct refinement is performed on every event mat $\boldsymbol{E}_{t}$ by minimizing the photometric errors to establish relative pose constraints.
After that, a tightly joint optimization scheme (Eq.~\eqref{joint_nonlinear_optimization}) is performed over the sliding window which contains re-projection constraints and the relative pose constraints.
To ensure the co-visibility between the feature-based and direct constraints, the state variables shared by these two constraints are synchronized before optimization. 

\subsection{Hybrid Optimization for Feature-based EVIO and Event-based Direct Tracking} 
\label{Hybrid Optimization for Feature-based EVIO and Event-based Direct Tracking}

Events are more suitable for the direct methods compared to images, and they also inherently overcome challenges associated with direct-based methods.
Firstly, events are triggered by intensity changes and naturally select high-gradient pixels, effectively substituting the gradient selection process in image-based direct methods~\cite{dso}.
Secondly, the high temporal resolution of events results in minimal displacement between two event frames, enabling the attainment of an optimal solution~\cite{event-slam-survey}.
Thirdly, the event camera is immune to photometric distortions/variations, gradual brightness changes, and highly non-convex or non-linear intensity values.
Besides, both feature-based and direct-based methods have their strengths and weaknesses in various aspects, which complement each other to some extent~\cite{svo,envio}. 
The robustness provided by salient visual features makes feature-based methods well-suited for handling irregular scene variations and large inter-frame motions. 
Conversely, direct-based methods exhibit better resilience and superior accuracy in low-textured scenes by leveraging high-gradient sub-pixel alignment.

Therefore, we integrate the feature-based and direct-based methods into an event-based hybrid pose tracking, combining the superior robustness of feature-based event pose tracking and the relatively high accuracy achieved through event-based direct alignment.
We define hybrid optimization as the process of maximizing a posterior based on feature-based and direct-based measurements.
Assuming that these two measurements are independent of each other and the noise with each measurement is zero-mean Gaussian distributed, the hybrid optimization problem can be formulated as the minimization of a series of costs as follows:
\begin{equation}
\begin{aligned}
    \begin{split}
       \boldsymbol{\chi}^{*}
       & = \mathop{\arg\max}_{\boldsymbol{\chi}} p(\boldsymbol{\chi} | \boldsymbol{z}) \\
       & = \mathop{\arg\min}_{\boldsymbol{\chi}} \left\{  ||\boldsymbol{r}_{p}-\boldsymbol{H}_{p}\boldsymbol{\chi}||^{2} + \sum_{i=1}^{n} ||\boldsymbol{r}(\boldsymbol{z}_{i}, \boldsymbol{\chi})||^{2}_{\boldsymbol{P}_{i}}  \right\}  \\ 
       \boldsymbol{\chi}  & =[\boldsymbol{p}^{w}_{b}, \boldsymbol{q}^{w}_{b}, \boldsymbol{v}^{w}_{b}]
    \end{split}
\label{joint_nonlinear_optimization}
\end{aligned}
\end{equation}
where $\boldsymbol{\chi}$ is the estimated state of the system, the state is composed by the position $ \boldsymbol{p}^{w}_{b} $, and the orientation quaternion $ \boldsymbol{q}^{w}_{b} $, and the velocity $\boldsymbol{v}^{w}_{b} $ of the IMU in the world frame.
$\boldsymbol{z}$ stands for the aggregation of the independent feature-based and direct-based measurements ($n=2$ in our framework), and $\left\{ \boldsymbol{r}_{p}, \boldsymbol{H}_{p} \right\}$ encapsulates the prior information or the marginalization~\cite{sibley2010sliding,leutenegger2013keyframe} of the system state.
$\boldsymbol{r}\left( \cdot \right)$ denotes the residual function of each measurement and $||\cdot||_{\boldsymbol{P}}$ is the Mahalanobis norm.
We decompose this optimization problem as two individual factors, the relative pose constraints from direct-based measurements (Eq.~\eqref{relative_pose_constraint}), and the constraints from feature-based EVIO (Eq.~\eqref{feature_based_nonlinear_optimization}).

\subsection{Event-based Direct Pose Tracking} 
\label{Event-based Direct Pose Tracking}

\begin{figure*}[htb]  
        \subfigbottomskip=-1pt 
        \subfigcapskip=-10pt 
        \centering
        \captionsetup{justification=justified}
        \subfigure[Event-based 2D-2D alignment]{
            \begin{minipage}[t]{1.0\columnwidth}
            \centering
            \includegraphics[width=1.0\columnwidth]{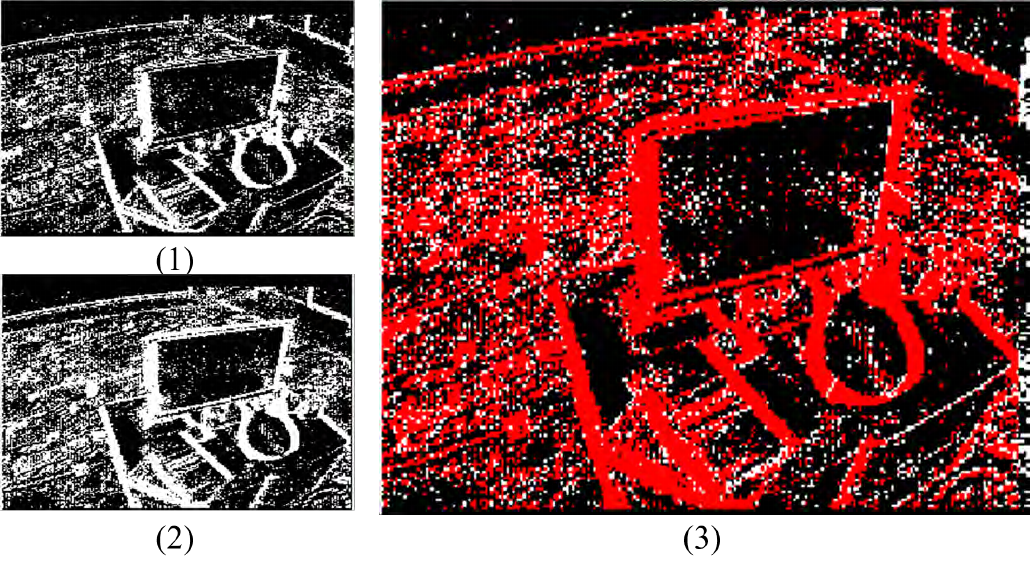}
            \label{fig:event-based 2D-2D alignment}
            \end{minipage}%
        }        
        \subfigure[Event-based 2D-3D alignment]{
            \begin{minipage}[t]{1.0\columnwidth}
            \centering
            \includegraphics[width=1.0\columnwidth]{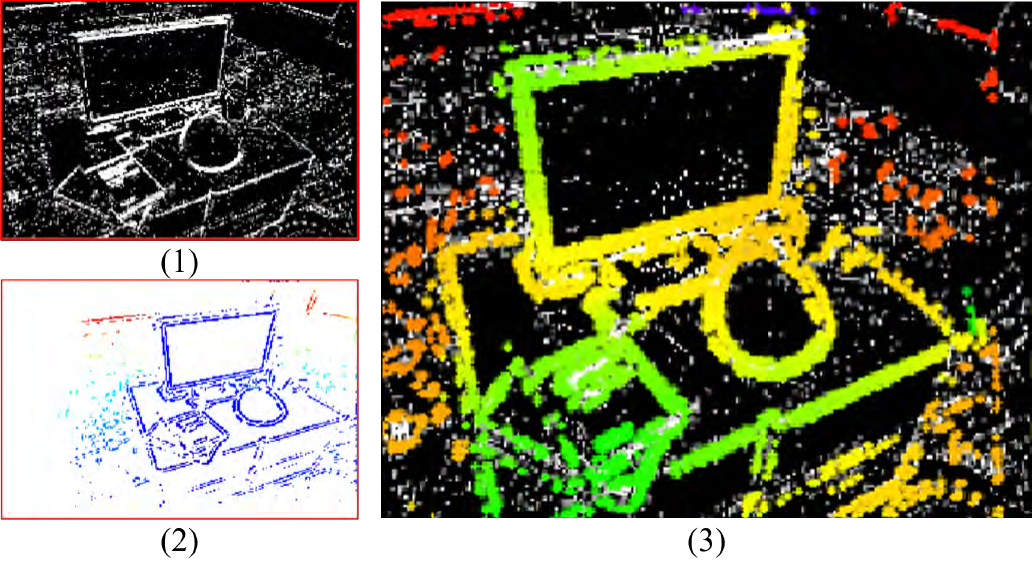}
            \label{fig:event-based 2D-3D alignment}
            \end{minipage}%
        }
        \caption{
        Direct event-based alignment.
        (a) Event-based 2D-2D alignment:
        The 2D event mat in the current timestamp (a1) 
        is warped to 
        the 2D event mat in the previous timestamp
        (a2). 
        The result (a3) is the alignment between the 2D current event mat (white) and the previous 2D event mat (red).
        (b) Event-based 2D-3D alignment:
        The current 2D event mat aggregated through a small number of events(b1)
        is warped to 
        the projected mat recovered from the event-based 3D semi-dense depth in (b2).
        The result (b3) is a good alignment between the 2D current event mat (white) and the projected 3D event-based map (color).
        }  
        \label{fig:direct_event_tracking}
\end{figure*}%

The direct-based measurements of our framework align the observed 2D event mats of the two consecutive timestamps.
The photometric errors can be created by the 2D event mat $\boldsymbol{E}_{t}$ at two timestamps $k$ and $i$:
\begin{equation}
\begin{aligned}
\boldsymbol{e}_{2D-2D}
    & = \boldsymbol{E}_{k}-\boldsymbol{E}_{i} \\
    & = \boldsymbol{E}(u_{k},v_{k}) - \boldsymbol{E}(\Delta \boldsymbol{T}^{k}_{i}(u_{i},v_{i})
\label{2D_2D_Photometric_Error}
\end{aligned}
\end{equation}
The data association of the pixel between these two event mats is completed using the inverse compositional Lucas-Kanade method (see Fig.~\ref{fig:event-based 2D-2D alignment}), which iteratively computes the incremental motion $\Delta \boldsymbol{T}^{k}_{i}$ from the reference pose of the $\boldsymbol{E}_{i}$  at timestamp $i$ to the current pose of the $\boldsymbol{E}_{k}$ at timestamp $k$.
While the Jacobian and Hessian Matrix of $\boldsymbol{e}_{2D-2D}$ are given by:
\begin{equation}
\begin{split}
    J &= \sum_{u,v} \frac{\partial \boldsymbol{e}_{2D-2D}}{\partial \Delta \boldsymbol{T}^{k}_{i}} \cdot \boldsymbol{e}_{2D-2D} \\
    H &=\sum_{u,v} J \cdot J^{T},  \quad  H \cdot \Delta \chi = J
\end{split}
\end{equation}
where $\Delta \chi$ is the update pose vector, the Lie-algebra, representing the incremental change of the pose within each iteration.  
While the incremental motion $\Delta \boldsymbol{T}^{k}_{i}$ from the reference frame to the current frame can be iteratively calculated as:
\begin{equation}
    \Delta \boldsymbol{T}^{k}_{i} \leftarrow exp_{\boldsymbol{SE(3)}}(\Delta \chi) \Delta \boldsymbol{T}^{k}_{i}
\end{equation}

Instead of directly incorporating the photometric errors term in the graph optimization, the pose-pose constraint is leveraged to implicitly contain the relative transforms derived from the direct method.
The relative pose constraint based on the optimized incremental pose $\Delta \boldsymbol{T}^{k}_{i}$ in Eq.~\eqref{2D_2D_Photometric_Error} is given by:
\begin{equation}
\boldsymbol{r}(\boldsymbol{z}_{direct},\boldsymbol{\chi})=\sum_{i,k \in K} ||\Delta \boldsymbol{T}^{k}_{i} \cdot (\boldsymbol{T}^{w}_{b_{i}})^{-1} \cdot \boldsymbol{T}^{w}_{b_{k}}||^{2}
\label{relative_pose_constraint}
\end{equation}
where $\boldsymbol{z}_{direct}$ represents the direct-based measurements. $K$ ($K=10$ in our experiments) is the number of keyframes in the sliding window.
$\boldsymbol{T}^{w}_{b_{i}}$ and $\boldsymbol{T}^{w}_{b_{k}}$ are the pose of the body frame in the world frame in $i^{th}$ and $k^{th}$ keyframe.

An example of the event-based 2D-3D alignment is also illustrated in Fig.~\ref{fig:event-based 2D-3D alignment}. 
Current event-based direct pose tracking methods~\cite{GWPHKU:EVO,GWPHKU:ESVO,EDS} and~\cite{DEVO} rely on the 2D-3D geometric alignment, resulting in relatively lower localization accuracy and even failures in pose tracking.
This may be attributed to the slow convergence of the estimated 3D map or the rapid changes in edge patterns between adjacent event packets.
We posit that the effectiveness of the event-based 2D-3D alignment method relies heavily on the accurate construction of the event-based 3D depth map and its successful alignment with the current 2D event frame.  
This limitation results in a notable lack of robustness in these event-based direct pose tracking methods, particularly in scenarios involving aggressive motion or HDR. 
Conversely, our proposed event-based 2D-2D alignment excels in performance without relying on the 3D depth maps, surpassing the performance of the conventional event-based 2D-3D alignment.

\subsection{Feature-based EVIO Pose Tracking} 
\label{Feature-based EVIO Pose Tracking}
The feature-based tracking module of our system is the same as the feature-based EVIO in our previous works~\cite{GWPHKU:PL-EVIO, GWPHKU:MyEVIO}, where the camera poses are optimized by minimizing a joint nonlinear least-squares problem for the system state $\boldsymbol{\chi}$ as follows:
\begin{equation}
\begin{aligned}
       r(&\boldsymbol{z}_{feature}, \boldsymbol{\chi})=\sum_{k=0}^{K-1}||\boldsymbol{e}^{k}_{imu}||^{2}\\
       &
       +\sum_{k=0}^{K-1}\sum_{l\in \xi}||\boldsymbol{e}^{k,l}_{image}||^{2}+\sum_{k=0}^{K-1}\sum_{l\in \zeta}||\boldsymbol{e}^{k,l}_{event}||^{2}
\label{feature_based_nonlinear_optimization}
\end{aligned}
\end{equation}

Eq.~\eqref{feature_based_nonlinear_optimization} is composed of three components: 
(i) the IMU pre-integration residuals $\boldsymbol{e}^{k}_{imu}$;
(ii) the feature-based image residuals $\boldsymbol{e}^{k,l}_{image}$;
(iii) the feature-based event residuals $\boldsymbol{e}^{k,l}_{event}$.
$\zeta$ and $\xi$ are the set of event-corner features and image features, respectively.
$K$ ($K=10$ in our experiments) is the total number of keyframes in the sliding window.
The detailed mathematical explanation and the keyframe selection scheme can be found in our previous work~\cite{GWPHKU:PL-EVIO}.
Here, we only review the re-projection constraint of the event-corner features. 
Considering the $l^{th}$ event-corner feature that is first observed in the $i^{th}$ keyframe, the residual for its observation in the $k^{th}$ keyframe is defined as:
\begin{equation}
\begin{split}
        \boldsymbol{e}^{k,l}_{event}=&
                \left[
                \begin{array}{c}
                u^{k}_{l}\\
                v^{k}_{l}\\
                \end{array} 
                \right]
                \\
                &
                -\pi_{e}\cdot(\boldsymbol{T}^{b}_{e})^{-1}\cdot \boldsymbol{T}^{b_{k}}_{w}\cdot \boldsymbol{T}^{w}_{b_{i}}\cdot \boldsymbol{T}^{b}_{e}\cdot \pi^{-1}_{e}(\frac{1}{\lambda_{l}},\left[
                        \begin{array}{c}
                        u^{i}_{l}\\
                        v^{i}_{l}\\
                        \end{array} 
                        \right])
 \label{re-projection-point-event}
\end{split}
\end{equation}
where 
$\lambda_{l}$ is the inverse depth of the event-corner features; $[u^{i}_{l}, v^{i}_{l}]^\mathrm{T}$
is the first observation of the $l^{th}$ event-corner feature in the $i^{th}$ keyframe.
$[u^{k}_{l}, v^{k}_{l}]^\mathrm{T}$
is the observation of the same event-corner feature in the $k^{th}$ keyframe, $\pi_{e}$ and $\pi_{e}^{-1}$ denote the event camera projection and inverse projection.
$\boldsymbol{T}^{b}_{e}$ is the extrinsic matrix between the event camera frame and the body frame.
$\boldsymbol{T}^{w}_{b_{i}}$ indicates the movement of the body frame related to the world frame in timestamp $i$, $\boldsymbol{T}^{b_{k}}_{w}$ is the transpose of the pose of the body in the world frame in the $k^{th}$ keyframe.

\section{Event-based Dense Mapping} 
\label{3D Mapping}

The proposed event-based dense mapping module consists of three steps: 
$ \left( i \right)$ computing semi-dense depth maps from event streams (Section~\ref{section: Purely Event-based Semi-dense Mapping}); 
$ \left( ii \right)$ reconstructing dense depth maps from the event-based semi-dense depth with images as guidance  (Section~\ref{section: Image-guided Event-based Dense Mapping});
$ \left( iii \right)$ fusing the estimated local depths into the global 3D map using TSDF-based fusion (Section~\ref{section: TSDF-based Map Fusion}).

\subsection{Purely Event-based Semi-dense Mapping}
\label{section: Purely Event-based Semi-dense Mapping}
We develop the event-based space-sweep algorithm used in~\cite{GWPHKU:EMVS} to perform semi-dense depth estimation using the monocular event camera, which targets a multi-view-stereo (MVS) problem.
It consists of two steps: building a disparity space image (DSI) using a space-sweep method~\cite{space-sweep} from different reference viewpoints (RV), and then identifying the local maxima of the DSI to determine the depth value of the pixel.

\begin{figure}[htb]  
    \captionsetup{justification=justified}
    \centering
    \includegraphics[width=1.0\columnwidth]{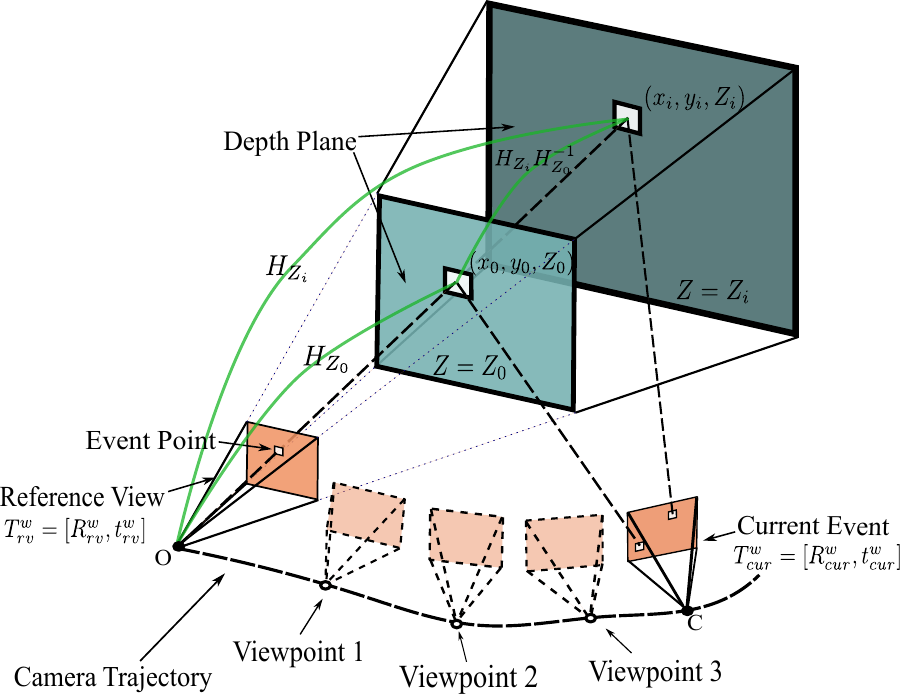}
    \caption{
    The model of event-based back-projection and space-sweep calculations across different depth planes of the DSI.
    }
    \label{Event-based back-projection}
\end{figure}

A new RV would be chosen when the motion of the event camera surpasses a specified threshold. 
Subsequently, all events between two consecutive RVs are back-projected to the front RV.
The creation of an RV triggers the construction of an associated local depth map from that viewpoint.
The back-projection rays from these nearby viewpoints are the perspective projection rays of 2D events into the 3D space (DSI) based on the camera model.
The DSI, shown in Figs.~\ref{Event-based back-projection} and~\ref{fig:fig4.b}, is a 3D voxel grid constructed for each RV.
It describes the distribution of event back-projected rays and the scores stored in each voxel (i.e., the number of back-projected rays passing through each voxel from nearby viewpoints).
The DSI is defined by $N$ depth planes $ \left\{ Z_{i} \right\} ^{N-1}_{i=1}$ with each depth plane discretized into $ w \times h \times N $ cuboid voxels, where $w$ and $h$ are the width and height of the event camera, respectively.
$N$ is equal to 100 in our implementation.
The back-projecting events to the DSI can be discretized to the execution of mapping events to all depth planes.
We can warp an event point $\left( u_{i},v_{i} \right)$ from the current event camera to the canonical plane $Z=Z_{0}$ of RV using the planar homography $\boldsymbol{H}_{Z_{0}}:$ $(u_{i},v_{i},1)^{\mathrm{T}} \to (x(Z_{0}), y(Z_{0}),1)^{\mathrm{T}}$:
\begin{equation}
\boldsymbol{H}^{-1}_{Z_{0}}=\boldsymbol{R}^{cur}_{rv}+\frac{1}{Z_{0}} \boldsymbol{t}^{cur}_{rv} \boldsymbol{e}^{\mathrm{T}}_{3}
\end{equation}
where $\boldsymbol{e}_{3}=(0,0,1)^{\mathrm{T}}$; $\boldsymbol{R}^{cur}_{rv}$ and $\boldsymbol{t}^{cur}_{rv}$ are the rotation matrix and translation vector from the RV to the current event camera, respectively:
\begin{equation}
\boldsymbol{T}^{cur}_{rv}=
    \left[
        \begin{matrix}
            \boldsymbol{R}^{cur}_{rv} & \boldsymbol{t}^{cur}_{rv}\\
            0 & 1
        \end{matrix}
    \right]
    =(\boldsymbol{T}^{rv}_{w} \cdot \boldsymbol{T}^{w}_{cur})^{-1}=(\left(\boldsymbol{T}^{w}_{rv}\right)^{-1} \cdot \boldsymbol{T}^{w}_{cur})^{-1}
\end{equation}
where $\boldsymbol{T}^{w}_{cur}$ and $\boldsymbol{T}^{w}_{rv}$ are the pose of the event camera in the current timestamp and the pose of the RV related to the world frame from the tracking module of EVI-SAM, respectively.
After that, events in the other plane of DSI are transformed with the depth from $Z=Z_{0}$ to $Z=Z_{i}$ (shown in Fig.~\ref{Event-based back-projection}) using the following equation:
\begin{equation}
\boldsymbol{H}_{Z_{i}} \boldsymbol{H}^{-1}_{Z_{0}}=\left( \boldsymbol{R}^{cur}_{rv}+\frac{1}{Z_{i}} \boldsymbol{t}^{cur}_{rv} \boldsymbol{e}^{\mathrm{T}}_{3} \right)^{-1}
                         \left( \boldsymbol{R}^{cur}_{rv}+\frac{1}{Z_{0}} \boldsymbol{t}^{cur}_{rv} \boldsymbol{e}^{\mathrm{T}}_{3} \right)
\end{equation}

The optical center of the event camera in the coordinate frame of the RV can be defined as:
\begin{equation}
    \boldsymbol{X}=(X_{c}, Y_{c}, Z_{c})=-\left( \boldsymbol{R}^{cur}_{rv} \right) ^{\mathrm{T}} \cdot \boldsymbol{t}^{cur}_{rv}
\end{equation}

Then, we can calculate the warped point $(x(Z_{i}), y(Z_{i}))$ in the canonical plane $Z=Z_{i}$ from the point $\left( u_{i},v_{i} \right)$ in the current event camera as follows:
\begin{equation}
    \begin{split}
    & x(Z_{i})= \frac{Z_{0} (Z_{i}-Z_{0})}{Z_{i} (Z_{0}-Z_{c})} x(Z_{0}) +\frac{1}{Z_{i}} (1-\frac{(Z_{i}-Z_{0})}{(Z_{0}-Z_{c})} ) X_{c}  \\
    & y(Z_{i})= \frac{Z_{0} (Z_{i}-Z_{0})}{Z_{i} (Z_{0}-Z_{c})} y(Z_{0}) +\frac{1}{Z_{i}} (1-\frac{(Z_{i}-Z_{0})}{(Z_{0}-Z_{c})} ) Y_{c} 
    \end{split}
\end{equation}

After back-projecting events to DSI, we count the number of back-projection rays that pass through each voxel and determine whether or not a 3D point exists in each DSI voxel.
Based on the theory that the regions where multiple back-projection rays nearly intersect are likely to be a 3D point in the scene, the 3D points can be determined when the scores of the DSI voxels are at a local maximum. 
The aggregation of all these 3D points onto the image plane constitutes the semi-dense depth.

\subsection{Image-guided Event-based Dense Mapping}
\label{section: Image-guided Event-based Dense Mapping}
Events are sparse and mostly respond to moving edges so the event-based semi-dense depth is also incomplete, only capturing depths at edges.
Inspired by Ref.~\cite{ma2019sparse}, it is possible to reconstruct the geometry of an unknown environment using sparse and incomplete depth measurements.
Therefore, in this section, we leverage the complementary strengths of event-based and standard frame-based cameras by incorporating additional intensity images as guidance for event-based dense depth completion.
It builds on the theories that depth discontinuities are commonly correlated with intensity boundaries, whereas homogeneous intensity regions correspond to homogeneous depth parts~\cite{zhang2018probability,telea2004image}. 
Moreover, pixels sharing the same intensity values are likely to belong to the same block in the depth image~\cite{shenshaojie:quadtree}.

To this end, we can interpolate or fill the hole of the event-based semi-dense map based on the surrounding sparse depth information and the edges extracted from intensity images.
We first select the intensity image captured at the RV (Fig.~\ref{fig:fig4.g}) to ensure proper alignment with the event-based semi-dense depth.
Next, we employ the region-growing approach to segment the image, as depicted in Fig.~\ref{fig:fig4.h}.
Other different segmentation or edge extraction operators can also be used, though satisfactory results have already been obtained using such simple selection.
After that, all the semi-dense depth points at the current RV are projected onto the segmentation regions.
Since events are only generated at pixels where brightness changes occur, they naturally align with contours. 
Finally, the vacant depth region can be filled or inpainted by using the weighted sum of the depth-recovered projected event points within the corresponding segmentation region.

\begin{figure}[htb]  
    \captionsetup{justification=justified}
    \centering
    \includegraphics[width=0.9\columnwidth]{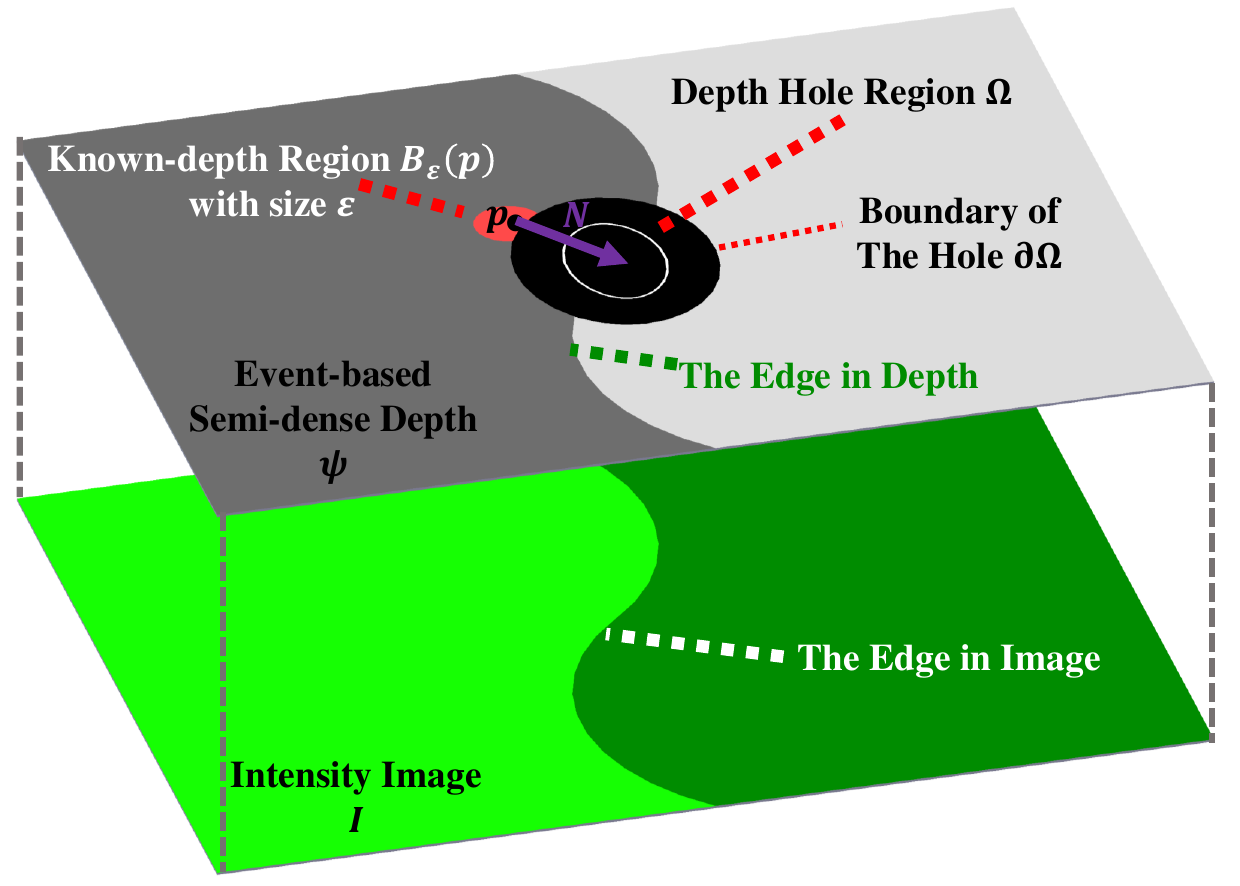}
    \caption{
    The model of our event-based dense mapping incorporates edges derived from the intensity image as guidance.
    The upper layer represents the event-based semi-dense depth. 
    This layer includes areas where the depth is known (regions successfully recovered through semi-dense mapping, marked in red) and areas with unknown depth (marked in black).
    The lower layer represents the intensity image with boundary information after segmentation. 
    Since events are triggered in regions with edges, the semi-dense depth and the intensity image edges at the corresponding locations are consistent.
    }
    \label{inpant_depth}
\end{figure}%

We extend the fast marching method~\cite{telea2004image} to incorporate the intensity image as guidance during the filling or inpainting process for holes in the event-based semi-dense depth, ultimately producing the event-based dense depth.
We firstly estimate the depth of a pixel located on the boundary of unknown regions, as illustrated in Fig.~\ref{inpant_depth}, in which the depth value of point $\boldsymbol{p}$ is unknown situated on the boundary $\partial\boldsymbol{\Omega}$ of the depth-hole region $\boldsymbol{\Omega}$ in the event-based semi-dense depth $\boldsymbol{\psi}$.
Considering a small neighborhood region $\boldsymbol{B}_{\varepsilon}$ with size $\varepsilon$ around $\boldsymbol{p}$, where the depth values are known.
Assuming the depth values have similarities in the local region, the inpainting value of $\boldsymbol{p}$ can be determined by the values of the known-depth region $\boldsymbol{B}_{\varepsilon}$.
If $\varepsilon$ is small enough, we can consider the first order approximation $d(\boldsymbol{p},q), \boldsymbol{q} \in \boldsymbol{B}_{\varepsilon}$, the contribution value of known-depth points $\boldsymbol{q}$ to unknown-depth point $\boldsymbol{p}$ can be denoted as:
\begin{equation}
    d(\boldsymbol{p},\boldsymbol{q})=d(\boldsymbol{q})+\bigtriangledown d(\boldsymbol{q})(\boldsymbol{p}-\boldsymbol{q}) \approx d(\boldsymbol{q})
\end{equation}
where the $\boldsymbol{q}$ is the points with known depth values of region $\boldsymbol{B}_{\varepsilon}$;
$d(\boldsymbol{q})$ is the depth value of point $\boldsymbol{q}$, and $\bigtriangledown d(\boldsymbol{q})$ indicates the depth gradient at the pixel $\boldsymbol{q}$.

Next, the filled value of point $p$ can be calculated by the weighted sum of all the contribution values of points $\boldsymbol{q}$ in the $\boldsymbol{B}_{\varepsilon}(\boldsymbol{p})$, as follows:
\begin{equation}
\begin{aligned}
d(\boldsymbol{p})=d(\boldsymbol{p},\omega)
                & =\frac{\sum_{\boldsymbol{q} \in \boldsymbol{B}_{\varepsilon}(\boldsymbol{p})} \omega_{\boldsymbol{p},\boldsymbol{q}} d(\boldsymbol{p},\boldsymbol{q}) }{\sum_{\boldsymbol{q} \in \boldsymbol{B}_{\varepsilon}(\boldsymbol{p})} \omega_{\boldsymbol{p},\boldsymbol{q}}} \\
                & =\frac{\sum_{\boldsymbol{q} \in \boldsymbol{B}_{\varepsilon}(\boldsymbol{p})} \omega_{\boldsymbol{p},\boldsymbol{q}} d(\boldsymbol{q}) }{\sum_{\boldsymbol{q} \in \boldsymbol{B}_{\varepsilon}(\boldsymbol{p})} \omega_{\boldsymbol{p},\boldsymbol{q}}}
\label{eq_depth_inpainting}
\end{aligned}
\end{equation}

\begin{equation}
  \omega_{\boldsymbol{p},\boldsymbol{q}}= |\omega_{dir} \cdot \omega_{dst} \cdot \omega_{lev} \cdot \omega_{img}|\\
\label{eq:16}
\end{equation}
where $\omega_{\boldsymbol{p},\boldsymbol{q}}$ is the weighting coefficient that measures the similarity of depth value between the point $\boldsymbol{p}$ and $\boldsymbol{q}$.

The term of $\omega_{dir}$ ensures that pixels closer to the normal direction $\boldsymbol{N}(\boldsymbol{p})$ have higher contributions, as defined below:
\begin{equation}
  \omega_{dir}=\frac{\boldsymbol{p}-\boldsymbol{q}}{||\boldsymbol{p}-\boldsymbol{q}||} \cdot \boldsymbol{N}(\boldsymbol{p})
\end{equation}

The weight $\omega_{dst}$ in Eq.~\eqref{eq:16} determines the contribution of pixels based on geometric distances to $\boldsymbol{p}$.
\begin{equation}
  \omega_{dst}= \frac{1}{||\boldsymbol{p}-\boldsymbol{q}||^{2}}
\end{equation}
This term is directly associated with the continuity of depth, and we have empirically observed that its contribution is larger than that of other terms.

The term of $\omega_{lev}$ in Eq.~\eqref{eq:16} represents the level set term, assigning a higher weight to pixels closer to the contour through $p$, as follows:
\begin{equation}
  \omega_{lev}= \frac{1}{1+|T(\boldsymbol{p})-T(\boldsymbol{q})|}
\end{equation}
where $T(\boldsymbol{p})$ is the distance of point $\boldsymbol{p}$ to the initial boundary of depth-hole.
$\bigtriangledown T(\boldsymbol{p})= \boldsymbol{N}(\boldsymbol{p})$ denotes the normal of hole boundary at the point $\boldsymbol{p}$.
The missing data is repaired pixel-wisely and the pixel with smaller $T(\boldsymbol{p})$ has a higher inpainting priority.

Lastly, the weight $\omega_{img}$ in Eq.~\eqref{eq:16} ensures that pixels with similar pixel intensity to $\boldsymbol{I}(p)$ contribute more than others:
\begin{equation}
     \omega_{img}=e^{-\frac{||\boldsymbol{I}(p)-\boldsymbol{I}(q)||^{2}}{2}}
\end{equation}
$\boldsymbol{I}(p)$ and $\boldsymbol{I}(q)$ indicates the intensity value of the image at point $\boldsymbol{p}$ and $\boldsymbol{q}$.
This term allows the weighting function in Eq.~\eqref{eq:16} to integrate intensity information for dense depth recovery.
Pixels with similar intensity values are more likely to be part of the same block in the depth image, resulting in a higher weight.

To fill the whole region $\boldsymbol{\Omega}$ and generate the event-based dense depth, after recovering the depth of pixels on the boundary $\partial\boldsymbol{\Omega}$ of the depth hole $\boldsymbol{\Omega}$, we propagate the depth from $\partial\boldsymbol{\Omega}$ to $\boldsymbol{\Omega}$ through iteratively applying Eq.~\eqref{eq_depth_inpainting}.
Firstly, we set $T(\boldsymbol{p})=0$ for pixels in known regions and progressively generate the distance map $T(\boldsymbol{p})$ while marching into $\boldsymbol{\Omega}$, ensuring $||\bigtriangledown T(\boldsymbol{p})||=1$.
To remove the outliers and preserve the sharpness of depth discontinuities, we employ the bilateral and non-local means filter on the event-based dense depth (Fig.~\ref{fig:fig4.i}).
Finally, we render the texture information from the intensity image onto the event-based 3D point cloud to obtain a textured map (Fig.~\ref{fig:fig4.j}).

\begin{figure*}[htb]  
    \setlength{\abovecaptionskip}{-0.1em}
    \subfigtopskip=0pt 
    \subfigbottomskip=5pt 
    \subfigcapskip=-5pt 
    \captionsetup{justification=justified}
    \centering
    \subfigure[ Event Stream ]{
                \begin{minipage}[t]{0.3\columnwidth}
                \centering
                \includegraphics[width=1.0\columnwidth]{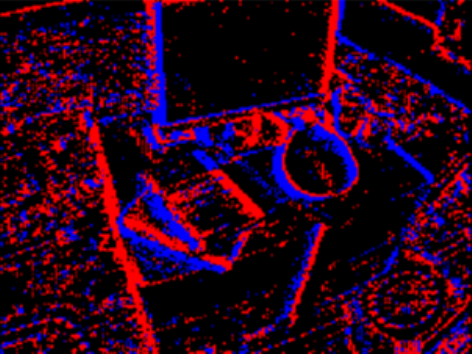}
                \label{fig:fig4.a}
                \end{minipage}%
        }
    \subfigure[ DSI ]{
                \begin{minipage}[t]{0.3\columnwidth}
                \centering
                \includegraphics[width=1.0\columnwidth]{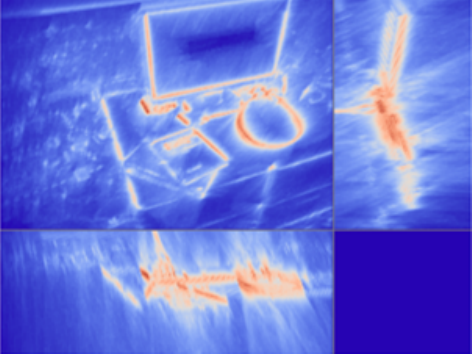}
                \label{fig:fig4.b}
                \end{minipage}%
        }
    \subfigure[ Semi-dense Depth ]{
                \begin{minipage}[t]{0.3\columnwidth}
                \centering
                \includegraphics[width=1.0\columnwidth]{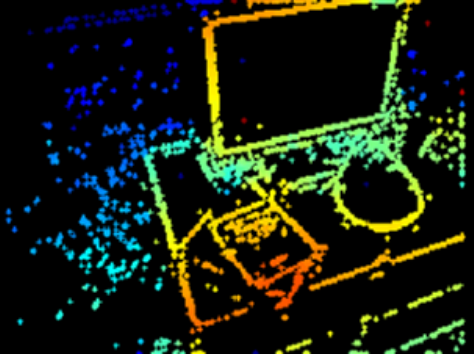}
                \label{fig:fig4.c}
                \end{minipage}%
        }
    \subfigure[ Point Cloud ]{
                \begin{minipage}[t]{0.3\columnwidth}
                \centering
                \includegraphics[width=1.0\columnwidth]{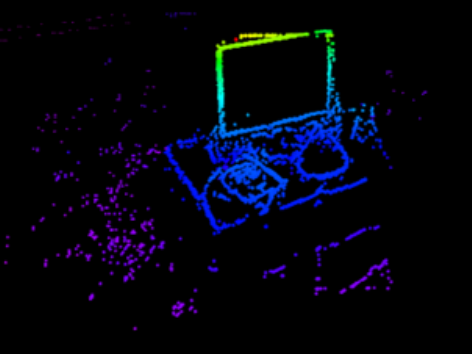}
                \label{fig:fig4.d}
                \end{minipage}%
        }
    \subfigure[ TSDF ]{
                \begin{minipage}[t]{0.3\columnwidth}
                \centering
                \includegraphics[width=1.0\columnwidth]{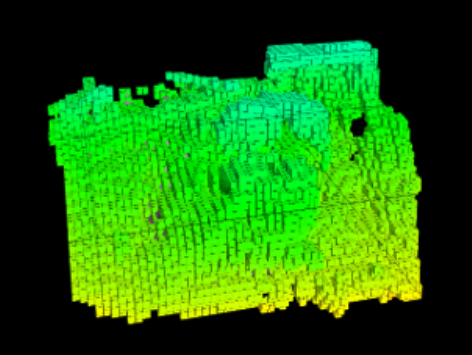}
                \label{fig:fig4.e}
                \end{minipage}%
        }
    \subfigure[ Surface Meshes]{
                \begin{minipage}[t]{0.3\columnwidth}
                \centering
                \includegraphics[width=1.0\columnwidth]{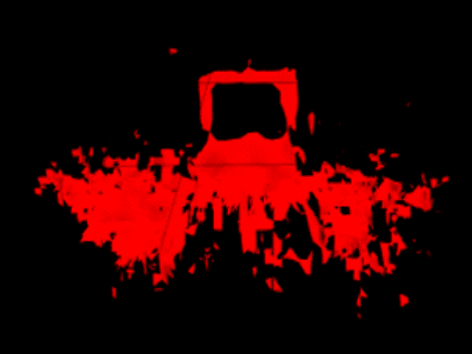}
                \label{fig:fig4.f}
                \end{minipage}%
        }

    \subfigure[ Image ]{
                \begin{minipage}[t]{0.3\columnwidth}
                \centering
                \includegraphics[width=1.0\columnwidth]{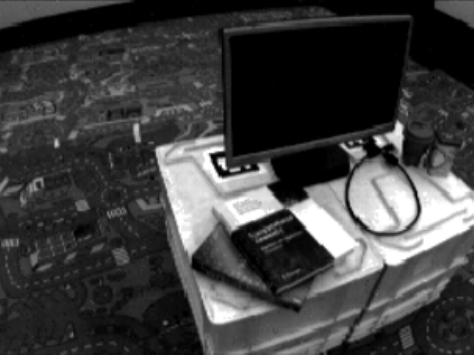}
                \label{fig:fig4.g}
                \end{minipage}%
        }
    \subfigure[ Segmentation ]{
                \begin{minipage}[t]{0.3\columnwidth}
                \centering
                \includegraphics[width=1.0\columnwidth]{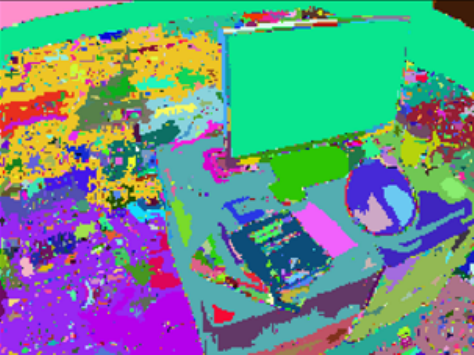}
                \label{fig:fig4.h}
                \end{minipage}%
        }
    \subfigure[ Dense Depth ]{
                \begin{minipage}[t]{0.3\columnwidth}
                \centering
                \includegraphics[width=1.0\columnwidth]{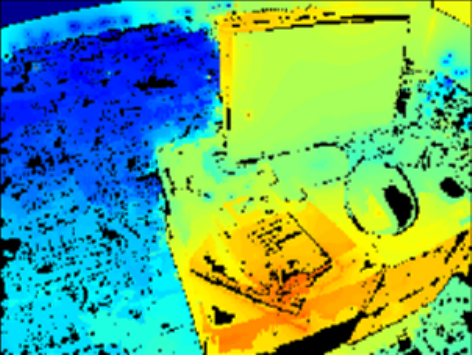}
                \label{fig:fig4.i}
                \end{minipage}%
        }
    \subfigure[ Point Cloud ]{
                \begin{minipage}[t]{0.3\columnwidth}
                \centering
                \includegraphics[width=1.0\columnwidth]{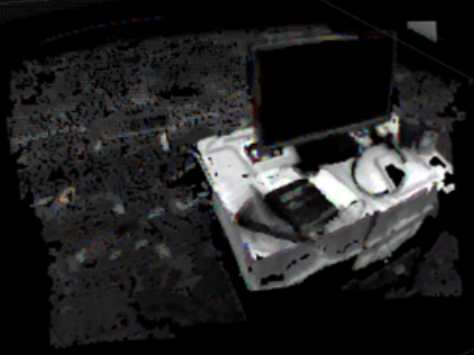}
                \label{fig:fig4.j}
                \end{minipage}%
        }
    \subfigure[ TSDF ]{
                \begin{minipage}[t]{0.3\columnwidth}
                \centering
                \includegraphics[width=1.0\columnwidth]{image/mapping_analysis/TSDF_of_the_Dense_Point_Cloud.pdf}
                \label{fig:fig4.k}
                \end{minipage}%
        }
    \subfigure[ Surface Meshes ]{
                \begin{minipage}[t]{0.3\columnwidth}
                \centering
                \includegraphics[width=1.0\columnwidth]{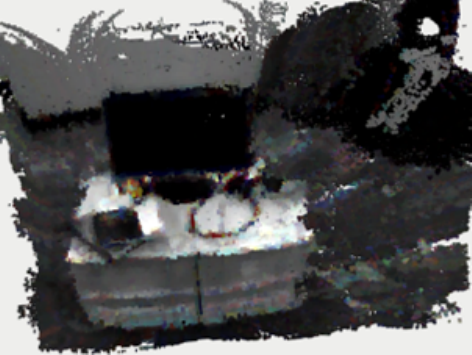}
                \label{fig:fig4.l}
                \end{minipage}%
        }
        
    \caption{
    The event-based semi-dense and dense mapping of our EVI-SAM. 
    (a) The raw event stream and (g) the intensity image from the event camera;
    (b) The disparity space image (DSI) of the reference view (RV) point;
    (c) The purely event-based semi-dense depth generated from our EVI-SAM;
    (d) The point cloud of the event-based semi-dense depth;
    (e) The occupied node of the semi-dense mapping after TSDF-fusion; 
    (f) The surface mesh of the semi-dense mapping; 
    (h) The segmentation on the image;
    (i) The event-based dense depth generated from our EVI-SAM;
    (j) The point cloud of the event-based dense depth with texture information;
    (k) The occupied node of the dense mapping after TSDF-fusion; 
    (l) The surface mesh of the dense mapping; 
    }
    \label{fig: event-based-mapping}
\end{figure*}%

\subsection{TSDF-based Map Fusion}
\label{section: TSDF-based Map Fusion}


TSDF can construct the global environmental surface representation based on consecutive depth maps and associated camera poses. 
To merge the event-based local depth into a TSDF, we perform the ray-cast from the origin of the event camera to each depth point in the event-based local depth map. 
Meanwhile, we update the signed distance and weight values of TSDF voxels along this ray.
The merging of local depths is based on the current distance $D_{last}$ and the weight value $W_{last}$ of a TSDF voxel, as well as the new update values from a specific point observation in the event-based local depth map.

Given a center position of TSDF voxel ($\boldsymbol{V} \in \mathbb{R}^{3}$) is passed by a ray between a 3D depth point ($\boldsymbol{P} \in \mathbb{R}^{3}$), and the event camera origin ($\boldsymbol{O} \in \mathbb{R}^{3}$).
The updated TSDF distance value $D_{new}$ and the weight value $W_{new}$ for this voxel are described as follows:
\begin{equation}
\begin{aligned}
       &
        D_{new} =\frac{W_{last} \cdot D_{last}+wd}{W_{last}+w},\\
       &
        W_{new}=min(W_{last}+w,W_{max}),
\end{aligned}
\label{eq:tsdf-update}
\end{equation}
where $d$ is the distance from the TSDF voxel center $\boldsymbol{V}$ to the new coming 3D depth point $\boldsymbol{P}$, defined as follows:
\begin{equation}
    d=\lVert \boldsymbol{P}-\boldsymbol{V}\rVert sign[(\boldsymbol{P}-\boldsymbol{V}) \cdot (\boldsymbol{P}-\boldsymbol{O})]
\end{equation}

While for the weight $w$ of this new coming 3D depth point, we follow Ref.~\cite{voxblox} to define it with the truncation function as:
\begin{equation}
w=
\left\{
     \begin{array}{lr}
     \frac{1}{\rho^{2}} ,& -\varepsilon \leq d \\
     \frac{1}{\rho^{2}} \cdot \frac{1}{d_{t}-\varepsilon}(d+d_{t}) ,& -d_{t} \leq d < -\varepsilon \\
     0 ,& d <-d_{t}  
     \end{array}
\right.
\end{equation}
where $\rho$ is the depth value of the event-based local depth in the event frame;
$d_{t}=4 \varepsilon$ is the truncated distance for TSDF, and $\varepsilon$ represents the size of voxel ($\varepsilon=0.01$ in our implementation).

Following the creation and update of TSDF voxels, the global depth map will be refined based on Eq.~\eqref{eq:tsdf-update}.
For each depth point in the event-based local depth maps, we project its position onto the TSDF voxel grid and group it with all other depth points that map to the same voxel.
Subsequently, we compute the weighted mean of all points within each TSDF voxel and perform the ray-cast only once on this mean position.
The surface meshes can also be reconstructed from updated TSDF voxels (shown in Figs.~\ref{fig:fig4.f} and~\ref{fig:fig4.l}), which allow us to get a better assessment of the perception environment.

\section{Evaluation} 
\label{Evaluation}
This section comprehensively evaluates the performance of our EVI-SAM system in terms of pose tracking and 3D mapping, employing both quantitative and qualitative assessments through extensive experiments.

Firstly, we show the effectiveness of our event-based hybrid pose tracking module by comparing the estimated trajectories against the ground truth pose in challenging situations (Section~\ref{Tracking Performance in Aggressive Motion and HDR Scenarios}).

Secondly, we assess the mapping performance of EVI-SAM through four sets of experiments.
The first set (Section~\ref{Mapping Performance in Diversity Scenarios}) evaluates the mapping performance across diverse scenarios to show the generation ability of our event-based dense mapping method.
The second set of experiments compare our 3D mapping approach with different mapping baselines, including those relying on event-based (Section~\ref{Mapping Performance Comparison with Baselines}), image-based (APPENDIX~\ref{appendices:Mapping Performance Comparison with Traditional Image-based Methods}), learning-based (APPENDIX~\ref{appendices:Mapping Performance Comparison with Learning-based Methods}), and NeRF-based (APPENDIX~\ref{appendices:Mapping Performance Comparison with NeRF-based Methods}) methods.
Moreover, Section~\ref{Mapping Performance in Challenge Situations} evaluates our mapping performance under challenging situations, such as HDR (Section~\ref{Mapping Performance in HDR}), and the scenarios involving aggressive motion (Section~\ref{Mapping Performance in Aggressive Motion}).
Section~\ref{sectioin: Full System Onboard Evaluation} evaluates the overall system on the onboard platform, including comparing local mapping with a commercial depth camera and assessing the global event-based 3D dense reconstruction performance.

Finally, the analysis of the computational performance and the discussion of the limitations are completed in Section~\ref{Running Time Analysis} and Section~\ref{Discussion of Limitations}, respectively.

\begin{table*}[htbp]
        \begin{center}
        \caption{Accuracy Comparison of Our EVI-SAM with Other EIO/EVIO Works in DAVIS240c Dataset~\cite{GWPHKU:event-camera-dataset_davis240c}}
        \label{table:Localization_accuracy_comparison_1}
        \resizebox{\columnwidth*2}{!}
        { 
        \begin{threeparttable}
        \renewcommand{\arraystretch}{1.0}
        \setlength{\tabcolsep}{1.0mm}
        \begin{tabular}{cccccccccccc} 
        \hline  
        Sequence 
        & \makecell{Ref.~\cite{GWPHKU:Event-based-visual-inertial-odometry} \\(E+I)} 
        & \makecell{Ref.~\cite{GWPHKU:ETH-EVIO} \\(E+I)} 
        & \makecell{Ref.~\cite{GWPHKU:Ultimate-SLAM}\\(E+I)}
        & \makecell{Ref.~\cite{GWPHKU:Ultimate-SLAM}\\(E+F+I)}
        & \makecell{Ref.~\cite{HASTE-VIO}\\(E+I)}
        & \makecell{Ref.~\cite{EKLT-VIO}\\(E+F+I)}
        & \makecell{Ref.~\cite{dai2022tightly}\\(E+I)}
        & \makecell{EIO~\cite{GWPHKU:MyEVIO} \\(E+I)}
        & \makecell{PL-EVIO~\cite{GWPHKU:PL-EVIO} \\(E+F+I)} 
        & \makecell{Ref.~\cite{lee2023event} \\(E+F+I)} 
        & \makecell{ \textbf{Our EVI-SAM} \\(E+F+I)} \\
        \hline
        boxes\_translation      & 2.69 & 0.57 & 0.76  &0.27          & 2.55 & 0.48 &1.0 & 0.34  & \textbf{0.06} &0.74 &0.11\\
        hdr\_boxes              & 1.23 & 0.92 & 0.67  &0.37          & 1.75 & 0.46 &1.8 & 0.40  & \textbf{0.10} &0.69 &0.13\\
        boxes\_6dof             & 3.61 & 0.69 & 0.44  &0.30          & 2.03 & 0.84 &1.5 & 0.61  & 0.21          &0.77 &\textbf{0.16}\\
        dynamic\_translation    & 1.90 & 0.47 & 0.59  &\textbf{0.18} & 1.32 & 0.40 &0.9 & 0.26  & 0.24          &0.71 &0.30\\
        dynamic\_6dof           & 4.07 & 0.54 & 0.38  &\textbf{0.19} & 0.52 & 0.79 &1.5 & 0.43  & 0.48          &0.86 &0.27\\
        poster\_translation     & 0.94 & 0.89 & 0.15  &\textbf{0.12} & 1.34 & 0.35 &1.9 & 0.40  & 0.54          &0.28 &0.34\\
        hdr\_poster             & 2.63 & 0.59 & 0.49  &0.31          & 0.57 & 0.65 &2.8 & 0.40  & \textbf{0.12} &0.52 &0.15\\
        poster\_6dof            & 3.56 & 0.82 & 0.30  &0.28          & 1.50 & 0.35 &1.2 & 0.26  & \textbf{0.14} &0.59 &0.24\\
        \hline
        Average                 & 2.58 & 0.69 & 0.47  &0.25          & 1.45 & 0.54 &1.56& 0.39  & 0.24          &0.65 &\textbf{0.21}\\
        \hline       
        \end{tabular}
        \begin{tablenotes} 
        \item \textit{Unit:\%, 0.21 means the average error would be 0.21m for 100m motion; Aligning 5 seconds [0-5s] of the estimated trajectory with the ground truth; The notations E, F, and I stand for the use of event, frame, and IMU, respectively.} 
        \end{tablenotes} 
        \end{threeparttable} 
        }
        \end{center}
\end{table*}

\begin{table*}[htbp] 
        \renewcommand\arraystretch{1.2}
        \LARGE 
        \begin{center}
        \caption{Accuracy Comparison of Our EVI-SAM with Other Image-based or Event-based Methods}
        \label{table:Localization_accuracy_comparison_2}
        \resizebox{\columnwidth*2}{!}
        { 
        \begin{threeparttable}
        \begin{tabular}{c|c|ccccccc} 
        \hline  
    \multicolumn{2}{c|}{\multirow{2}*{Sequence}}  
    & \makecell{ORB-SLAM3~\cite{ORB-SLAM3} \\(F+F+I)} 
    & \makecell{VINS-Fusion~\cite{GWPHKU:VINS-Fusion} \\(F+F+I)}
    & \makecell{EVO~\cite{GWPHKU:EVO} \\(E)}
    & \makecell{ESVO~\cite{GWPHKU:ESVO} \\(E+E)} 
    & \makecell{Ultimate SLAM~\cite{GWPHKU:Ultimate-SLAM}\\(E+F+I)} 
    & \makecell{PL-EVIO~\cite{GWPHKU:PL-EVIO} \\(E+F+I)} 
    & \makecell{ \textbf{Our EVI-SAM} \\(E+F+I)}   \\
    \cline{3-9}
    \multicolumn{2}{c|}{} & MPE / MRE & MPE / MRE & MPE / MRE & MPE / MRE & MPE / MRE & MPE / MRE & MPE / MRE\\
\hline
\multirow{9}*{HKU dataset}
    & hku\_agg\_translation    & 0.15 / 0.075& 0.11 / \textbf{0.019} &\textit{failed} &\textit{failed} &0.59 / 0.020& \textbf{0.07} / 0.091 &0.17 / 0.056\\
~ & hku\_agg\_rotation        & 0.35 / 0.11& 1.34 / \textbf{0.024} &\textit{failed} &\textit{failed} & 3.14 / 0.026 &\textbf{0.23} / 0.12 &0.24 / 0.056\\
~ & hku\_agg\_flip           & 0.36 / \textbf{0.39}& 1.16 / 2.02 &\textit{failed} &\textit{failed} & 6.86 / 2.04 &0.39 / 2.23 & \textbf{0.32} / 2.08\\
~ & hku\_agg\_walk           & \textit{failed} & \textit{failed} &\textit{failed} &\textit{failed} & 2.00 / 0.16&0.42 / \textbf{0.14} &\textbf{0.26} / 0.22\\
~ & hku\_hdr\_circle         & 0.17 / \textbf{0.12}& 5.03 / 0.60 &\textit{failed} &\textit{failed} & 1.32 / 0.54 &0.14 / 0.62 & \textbf{0.13} / 0.56\\
~ & hku\_hdr\_slow           &  0.16 / 0.058& 0.13 / \textbf{0.026} &\textit{failed} &\textit{failed} & 2.80 / 0.099&0.13 / 0.068 &\textbf{0.11} / 0.033\\
~ & hku\_hdr\_tran\_rota     & 0.30 / 0.042& 0.11 / \textbf{0.021} &\textit{failed} &\textit{failed} & 2.64 / 0.13 &\textbf{0.10} / 0.064 &0.11 / 0.026\\
~ & hku\_hdr\_agg             & 0.29 / \textbf{0.085}& 1.21 / 0.27 &\textit{failed} &\textit{failed} & 2.47 / 0.27 &0.14 / 0.30 &\textbf{0.10} / 0.26\\
~ & hku\_dark\_normal        & \textit{failed}& 0.86 / \textbf{0.028} &\textit{failed} &\textit{failed} & 2.17 / 0.031 &1.35 / 0.081 &\textbf{0.85} / 0.052\\
\hline
\multirow{17}*{\makecell{VECtor~\cite{GWPHKU:VECtor}}}
  & corner-slow  & 1.49 / 14.28& 1.61 / \textbf{14.06}& 4.33 / 15.52 & 4.83 / 20.98& 4.83 / 14.42& \textbf{2.10} / 14.21&  2.50 / 14.82\\
~ & robot-normal  & 0.73 / 1.18& \textbf{0.58} / 1.18& 3.25 / 2.00 &\textit{failed}& 1.18 / 1.11& 0.68 / 1.25&  0.67 / \textbf{0.85}\\
~ & robot-fast  & 0.71 / 0.70& \textit{failed}& \textit{failed} &\textit{failed}& 1.65 / 0.56&  \textbf{0.17} / 0.74&  0.22 / \textbf{0.41}\\
~ & desk-normal  & \textbf{0.46} / 0.41& 0.47 / 0.36& \textit{failed} &\textit{failed}& 2.24 / 0.56 & 3.66 / 0.45&  1.45 / \textbf{0.28}\\
~ & desk-fast   & 0.31 / 0.41& 0.32 / 0.33& \textit{failed} &\textit{failed} &  1.08 / 0.38 & \textbf{0.14} / 0.48&  0.18 / \textbf{0.38}\\
~ & sofa-normal   & \textbf{0.15} / 0.41& 0.13 / 0.40& \textit{failed} &1.77 / 0.60& 5.74 / 0.39&  0.19 / 0.46&  0.19 / \textbf{0.20}\\
~ & sofa-fast   & 0.21 / 0.43& 0.57 / 0.34& \textit{failed} &\textit{failed}& 2.54 / 0.36 & \textbf{0.17} / 0.47&  0.98 / \textbf{0.31}\\
~ & mountain-normal   & \textbf{0.35} / 1.00& 4.05 / 1.05& \textit{failed} & \textit{failed}& 3.64 / 1.06& 4.32 / 0.76&  1.39 / \textbf{0.65}\\
~ & mountain-fast   & 2.11 / 0.64& \textit{failed} & \textit{failed} &\textit{failed}& 4.13 / 0.62& \textbf{0.13} / 0.56&  0.38 / \textbf{0.30}\\
~ & hdr-normal  & \textbf{0.64} / 1.20& 1.27 / 1.10& \textit{failed} &\textit{failed} & 5.69 / 1.65&  4.02 / 1.52 &  5.74 / \textbf{0.87}\\
~ & hdr-fast   & 0.22 / 0.45& 0.30 / 0.34& \textit{failed} &\textit{failed} & 2.61 / 0.34& \textbf{0.20} / 0.50&  0.67 / \textbf{0.26}\\
~ & corridors-dolly  & \textbf{1.03} / 1.37& 1.88 / 1.37& \textit{failed} &\textit{failed}  & \textit{failed} & 1.58 / \textbf{1.37}&  1.58 / 1.38\\
~ & corridors-walk   & 1.32 / 1.31& \textbf{0.50} / 1.31& \textit{failed} &\textit{failed}  & \textit{failed} & 0.92 / \textbf{1.31}&  1.27 / 1.35\\
~ & school-dolly      & 0.73 / 1.02 & \textbf{1.42} / 1.06 & \textit{failed} &10.87 / 1.08  & \textit{failed}  & 2.47 / 0.97& 1.53 / \textbf{0.89}\\
~ & school-scooter    & 0.70 / \textbf{0.49}& \textbf{0.52} / 0.61& \textit{failed} &9.21 / 0.63 & 6.40 / 0.61& 1.30 / 0.54& 1.46 / 0.53\\
~ & units-dolly     & 7.64 / 0.41& 4.39 / 0.42& \textit{failed} &\textit{failed}  & \textit{failed} & 5.84 / 0.44& \textbf{0.59} / \textbf{0.35}\\
~ & units-scooter   & 6.22 / \textbf{0.22}& 4.92 / 0.24& \textit{failed} &\textit{failed}  & \textit{failed} & 5.00 / 0.42& \textbf{0.83} / 0.38\\
\hline
        \end{tabular}
        \begin{tablenotes} 
        \item \textit{Unit: MPE(\%) / MRE(deg/m); Aligning the whole ground truth trajectory with estimated poses; The notations E, F, and I stand for the use of event, frame, and IMU, respectively.} 
        \end{tablenotes} 
        \end{threeparttable} 
        }
        \end{center}
\end{table*}

\begin{figure*}[htb]  
    \setlength{\abovecaptionskip}{-0.1em}
    \subfigtopskip=0pt 
    \subfigbottomskip=5pt 
    \subfigcapskip=-5pt 
    \captionsetup{justification=justified}
    \centering
    \subfigure[ boxes\_6dof ]{
                \begin{minipage}[t]{0.45\columnwidth}
                \centering
                \includegraphics[width=1.0\columnwidth]{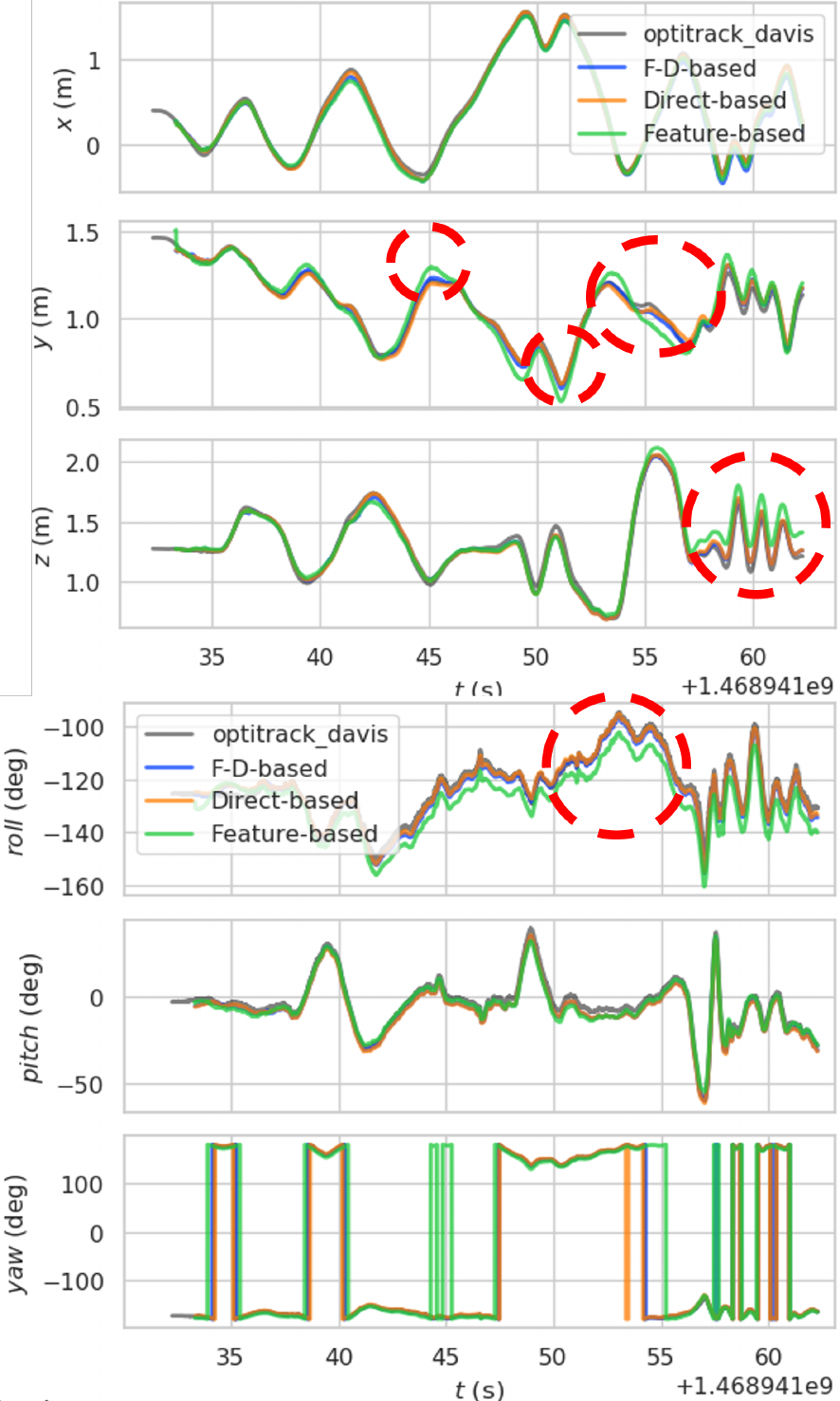}
                \label{fig:boxes_6dof_comparison}
                \end{minipage}%
        }   
        \subfigure[ hdr\_poster ]{
                \begin{minipage}[t]{0.45\columnwidth}
                \centering
                \includegraphics[width=1.0\columnwidth]{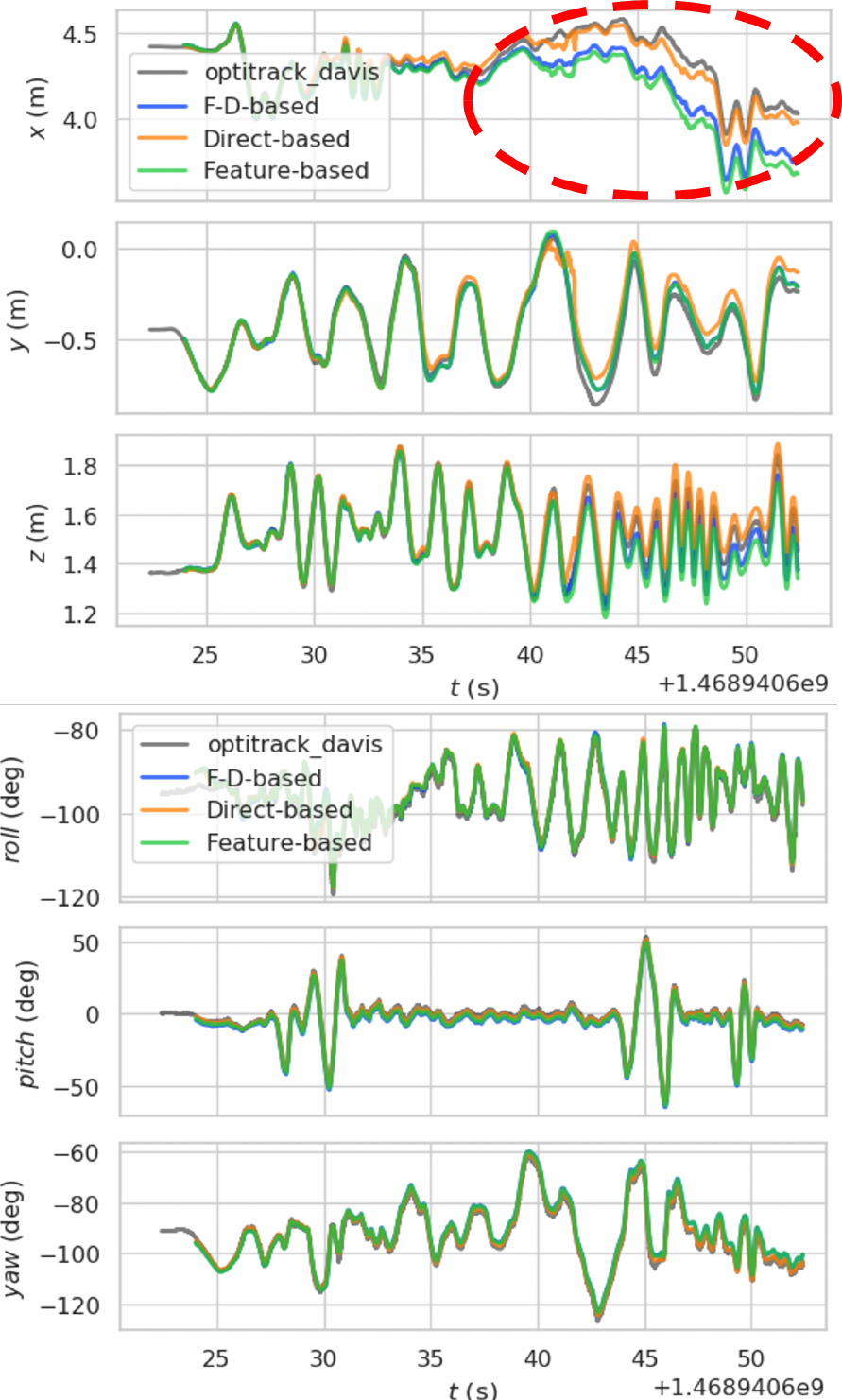}
                \label{fig:hdr_poster_comparison}
                \end{minipage}%
        }
        \subfigure[ hku\_agg\_walk ]{
                \begin{minipage}[t]{0.45\columnwidth}
                \centering
                \includegraphics[width=1.0\columnwidth]{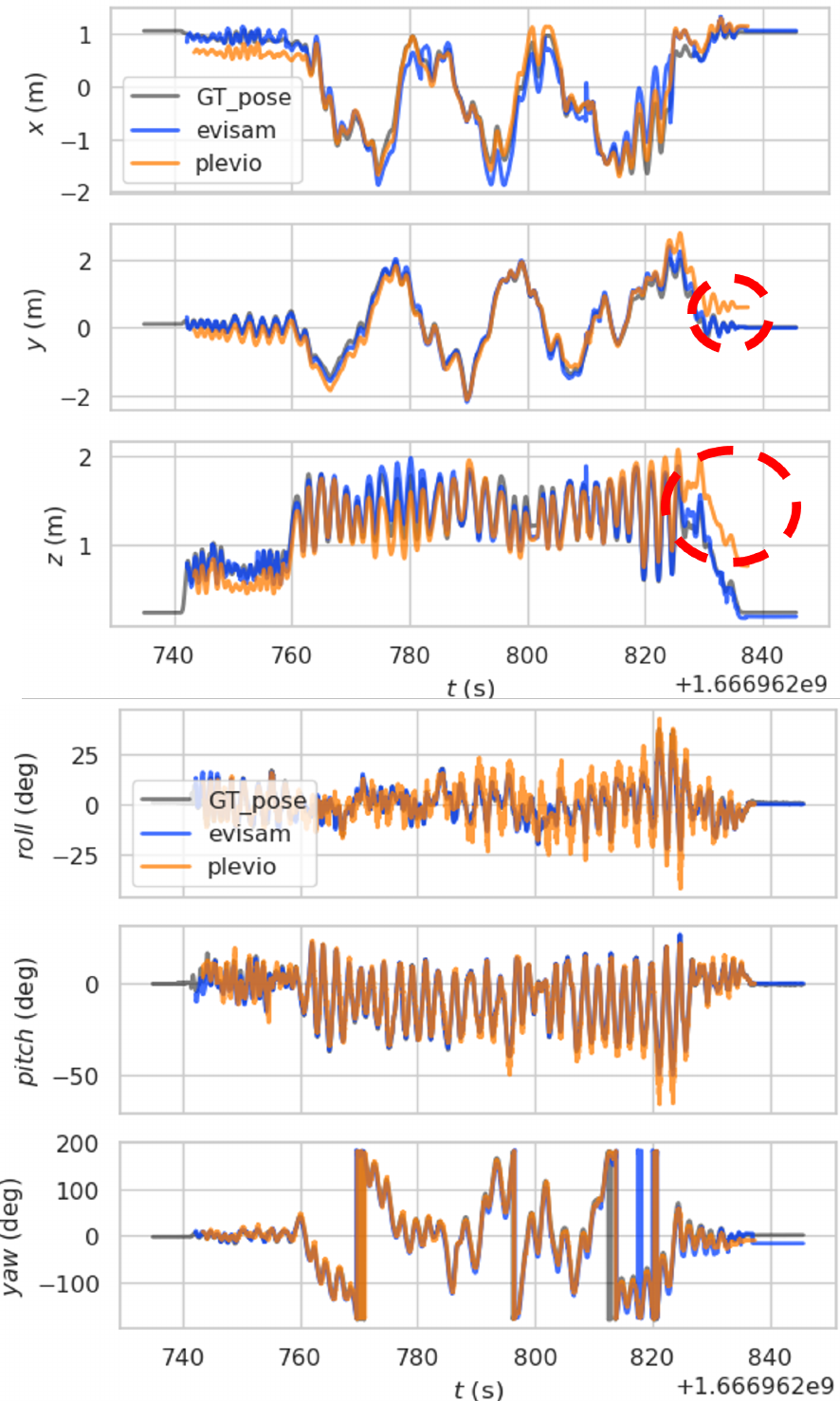}
                \label{fig:hku_agg_walk_comparison}
                \end{minipage}%
        }
        \subfigure[ mountain\_normal ]{
                \begin{minipage}[t]{0.45\columnwidth}
                \centering
                \includegraphics[width=1.0\columnwidth]{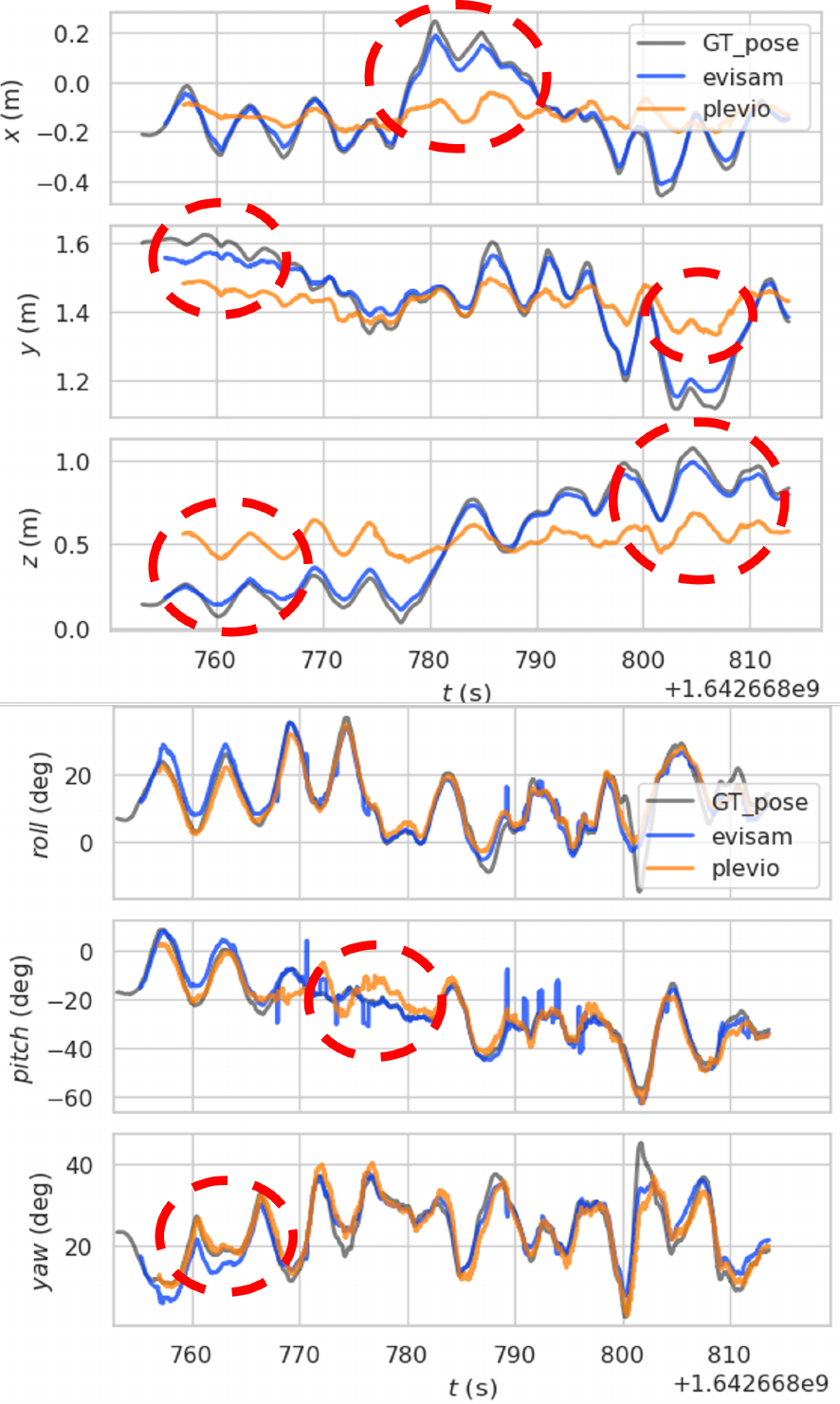}
                \label{fig:mountain_normal_comparison}
                \end{minipage}%
        }       
        
    \caption{
    Comparing the estimated trajectory in terms of translation and rotation produced by our EVI-SAM with the ground truth trajectory in DAVIS240C~\cite{GWPHKU:event-camera-dataset_davis240c}, HKU-dataset, and VECtor~\cite{GWPHKU:VECtor}.
    (a) and (b) show the comparison among our EVI-SAM with a feature-based event pipeline only, a direct-based event pipeline only, and the full event-based hybrid pipeline
    (feature-based + direct-based, F-D-based);
    (c) and (d) show the comparison between the PL-EVIO (feature-based) and our EVI-SAM (hybrid).
    The red dashed lines highlight the disparity between featured-based and direct-based methods.
    }
    \label{fig:pose_tracking_performance}
\end{figure*}%

\begin{figure*}[htbp]  
    \subfigtopskip=0pt 
    \subfigbottomskip=0pt 
    \subfigcapskip=-8pt 
    \captionsetup{justification=justified}
    \centering
        \subfigure[ poster\_translation ]{
                \begin{minipage}[t]{1.0\columnwidth}
                \centering
                \includegraphics[width=1.0\columnwidth]{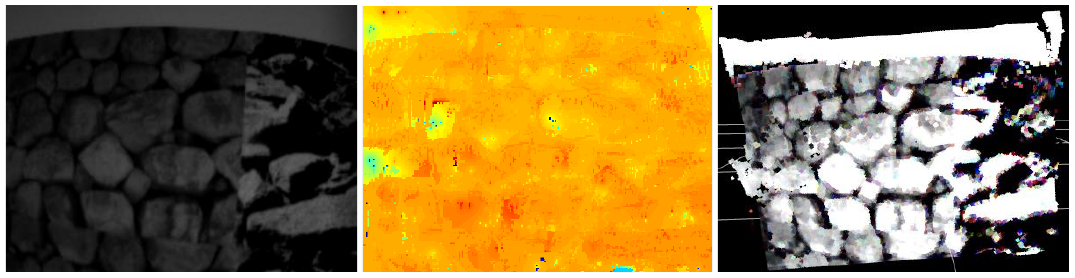}
                \label{fig:poster_translation}
                \end{minipage}%
        }
        \subfigure[ indoor\_flying\_3 ]{
            \begin{minipage}[t]{1.0\columnwidth}
            \centering
            \includegraphics[width=1.0\columnwidth]{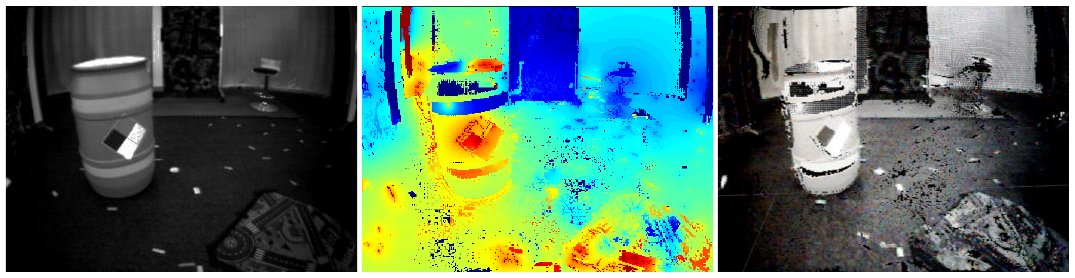}
            \label{fig:indoor_flying_2}
            \end{minipage}%
        }
        \subfigure[ HKU\_monitor ]{    
                \begin{minipage}[t]{1.0\columnwidth}
                \centering
                \includegraphics[width=1.0\columnwidth]{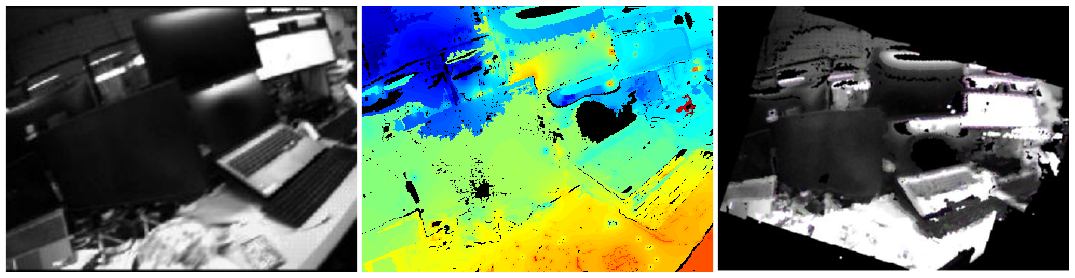}
                \label{fig:HKU_desk}
                \end{minipage}%
        }
        \subfigure[ HKU\_reader ]{
            \begin{minipage}[t]{1.0\columnwidth}
            \centering
            \includegraphics[width=1.0\columnwidth]{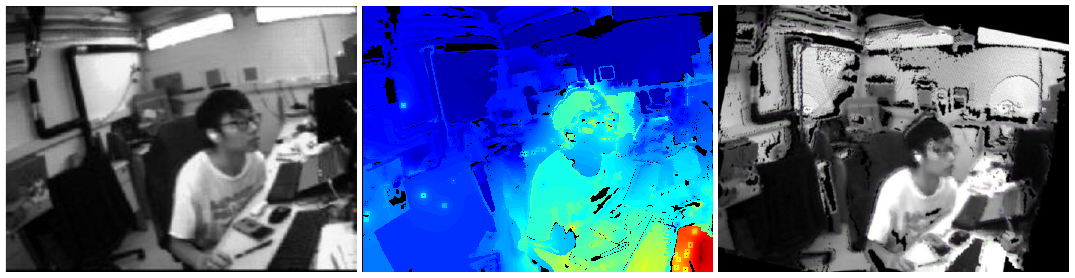}
            \label{fig:HKU_reader}
            \end{minipage}%
        }
        \subfigure[ HKU\_hdr\_box ]{
                \begin{minipage}[t]{1.0\columnwidth}
                \centering
                \includegraphics[width=1.0\columnwidth]{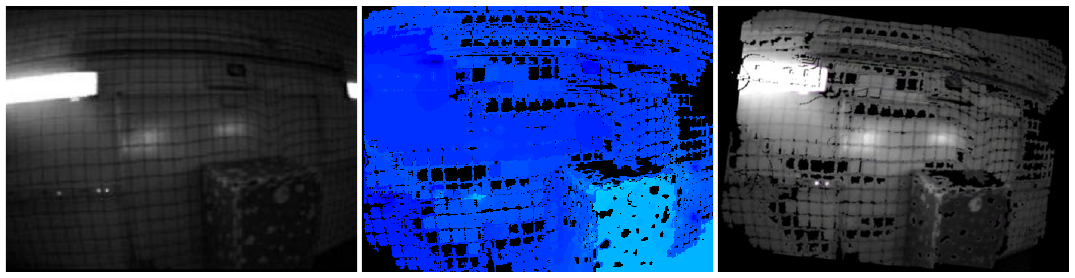}
                \label{fig:HKU_hdr_box}
                \end{minipage}%
        }
        \subfigure[ HKU\_reader\_1 ]{
            \begin{minipage}[t]{1.0\columnwidth}
            \centering
            \includegraphics[width=1.0\columnwidth]{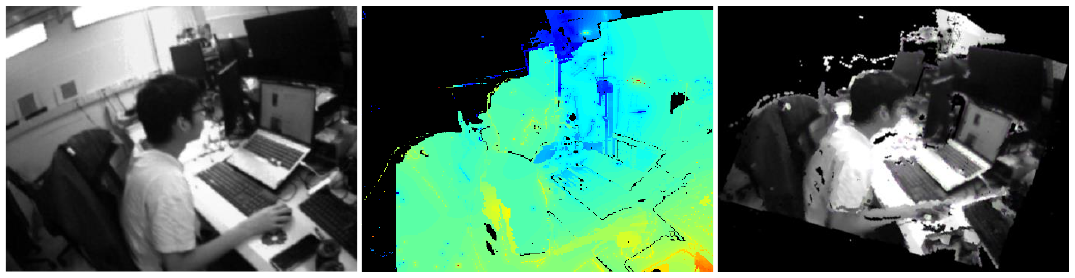}
            \label{fig:HKU_reader_1}
            \end{minipage}%
        }
         \subfigure[ HKU\_desk ]{    
                \begin{minipage}[t]{1.0\columnwidth}
                \centering
                \includegraphics[width=1.0\columnwidth]{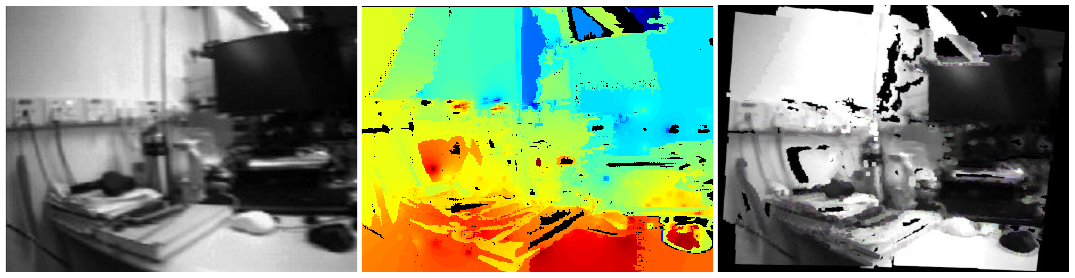}
                \label{fig:HKU_desk_1}
                \end{minipage}%
        }
        \subfigure[ dsec\_zurich\_city\_04\_a ]{    
                \begin{minipage}[t]{1.0\columnwidth}
                \centering
                \includegraphics[width=1.0\columnwidth]{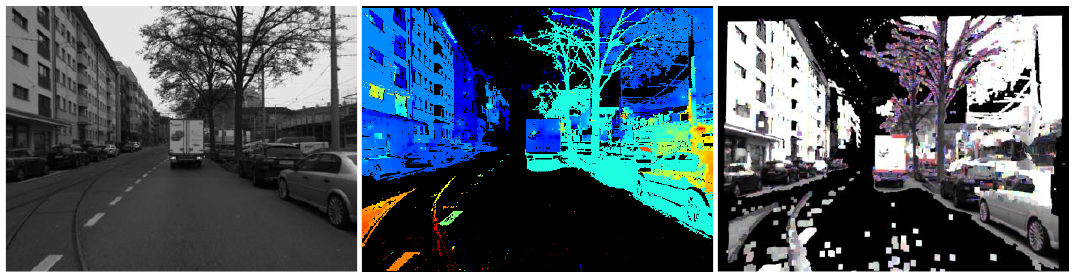}
                \label{fig:dsec_zurich_city_04_a}
                \end{minipage}%
        }
        \subfigure[ dsec\_zurich\_city\_04\_e ]{
            \begin{minipage}[t]{1.0\columnwidth}
            \centering
            \includegraphics[width=1.0\columnwidth]{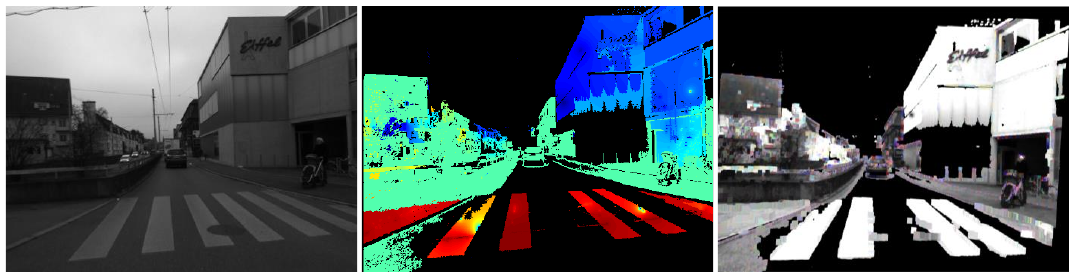}
            \label{fig:dsec_zurich_city_04_e}
            \end{minipage}%
        }
        \subfigure[ dsec\_zurich\_city\_04\_f ]{
                \begin{minipage}[t]{1.0\columnwidth}
                \centering
                \includegraphics[width=1.0\columnwidth]{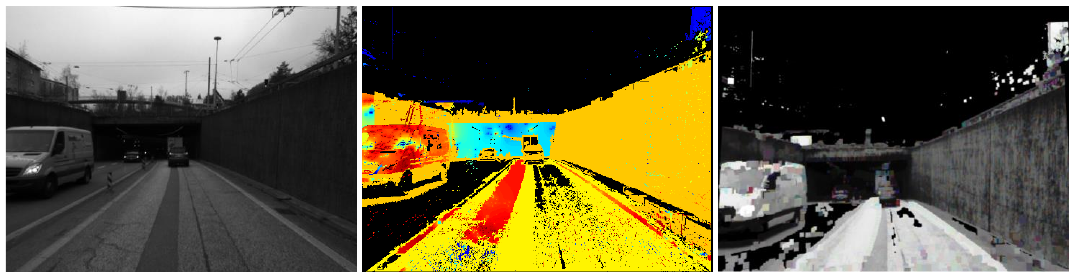}
                \label{fig:dsec_zurich_city_04_f}
                \end{minipage}%
        }
        
    \caption{
    Qualitative mapping result of our EVI-SAM in various data sequences.
    Each element includes the image view, dense depth, and the point cloud with texture information. 
    The depth is pseudo-colored, from red (close) to blue (far),
    in the range of 0.55-6.25 m for the rpg~\cite{zhou2018semi} and davis240c~\cite{GWPHKU:event-camera-dataset_davis240c};
    in the range 1.0-6.5 m for MVSEC~\cite{GWPHKU:MVSEC};
    in the range 0.5-4.0 m for HKU-dataset;
    in the range 4.0-200 m for DSEC~\cite{GWPHKU:DSEC}, respectively.
    }
    \label{fig:dense_mapping_performance}
\end{figure*}%

\subsection{Tracking Performance in Aggressive Motion and HDR Scenarios}
\label{Tracking Performance in Aggressive Motion and HDR Scenarios}

To evaluate the tracking performance of our event-based hybrid framework in challenging scenarios, we compare our system with other state-of-the-art image-based or event-based VO/VIO using publicly available datasets: DAVIS240c~\cite{GWPHKU:event-camera-dataset_davis240c}, HKU-dataset\footnote{\url{https://github.com/arclab-hku/Event_based_VO-VIO-SLAM}} and VECtor~\cite{GWPHKU:VECtor}.
These datasets offer events, grayscale frames, IMU, and ground truth poses, encompassing challenge scenarios such as HDR, aggressive motion, large-scale, as well as textureless environments.
The criterion used to assess tracking accuracy is the mean absolute trajectory error, which aligns the estimated trajectory with the ground-truth pose in the 6-DOF transformation (SE3).
The best results for each sequence are highlighted in bold.

In Table~\ref{table:Localization_accuracy_comparison_1}, we directly report raw results from the original papers we compared, as the same trajectory alignment protocol is utilized.
Our EVI-SAM achieves the best performance among event-based VIO, while with only slight improvement compared to PL-EVIO. 
Therefore, we further conduct Table~\ref{table:Localization_accuracy_comparison_2} to provide a more diverse evaluation.
Compared to pure direct event-based VO methods (e.g., EVO~\cite{GWPHKU:EVO}, ESVO~\cite{GWPHKU:ESVO}) or pure feature-based EVIO methods (e.g., Ultimate SLAM~\cite{GWPHKU:Ultimate-SLAM}), our EVI-SAM significantly surpasses them in both accuracy and robustness.
This underscores the effectiveness of our proposed hybrid tracking pipeline, which combines the robustness of the feature-based method with the high accuracy achieved by the direct-based method.
While compared with PL-EVIO~\cite{GWPHKU:PL-EVIO}, a feature-based method, our EVI-SAM also outperforms it in most sequences.
This highlights the effectiveness of our hybrid pose tracking pipeline, with the improvement stemming from the proposed event-based 2D-2D alignment.

An interesting phenomenon regarding the pure direct-based methods, the EVO~\cite{GWPHKU:EVO} and ESVO~\cite{GWPHKU:ESVO}, is that both of them crashed in most of the sequences listed in Table~\ref{table:Localization_accuracy_comparison_2}, with only a few sequences complete.
However, even in cases where they are completed successfully, these methods exhibit significant errors when compared to the ground truth poses.
This can be attributed to the challenges presented by our evaluation sequences, especially those containing aggressive motion.
Consequently, approaches relying on edge-map (2D-3D) model alignment struggle to provide reliable state estimations, as they heavily depend on the timely updates of the local 3D map.
Our proposed event-based 2D-2D alignment method does not suffer from these shortcomings.
Our event-based hybrid tracking module is robust and accurate enough to handle state estimation in such challenging evaluations.

We further analyze the performance comparison between feature-based and direct-based EVIO.
To ensure a fair comparison, we individually run the direct-based event tracking pipeline of EVI-SAM (Direct-based), the feature-based event tracking pipeline of EVI-SAM (Feature-based), and the entire EVI-SAM system (includes both the direct-based and feature-based pipelines, marks as F-D-based).
We visualize the estimated results for translation along the X-Y-Z axes and rotation around the roll-pitch-yaw axes, along with the ground truth in Fig.~\ref{fig:boxes_6dof_comparison} and~\ref{fig:hdr_poster_comparison}.
It is evident that the introduction of our event-based 2D-2D alignment significantly enhances the performance of pose tracking. 
For instance, in the X-axes of Fig.~\ref{fig:hdr_poster_comparison}, the Feature-based method deviates significantly from the ground truth between the 45th to 50th seconds, whereas the Direct-based method attains higher accuracy during this interval.
Regarding the F-D-based, it has some offset influenced by the feature-based pipeline, but its performance remains superior to that of the standalone feature-based event tracking pipeline.
On the other hand, it proves effective in handling irregular scene variations and large inter-frame motions, as evidenced by Table~\ref{table:Localization_accuracy_comparison_2}.
Our entire EVI-SAM system not only improves the accuracy compared to the standalone feature-based event tracking pipeline but also retains the advantages of the feature-based method.

Furthermore, we use PL-EVIO as a representative of the feature-based method and compare its 6-DoF estimation results with those of our EVI-SAM, as illustrated in Figs.~\ref{fig:hku_agg_walk_comparison} and~\ref{fig:mountain_normal_comparison}.
It is worth noting that although the numerical improvement in accuracy may not be pronounced in the tables due to statistical considerations, the effectiveness of our hybrid approach over a pure feature-based method is more evident when examining Fig.~\ref{fig:pose_tracking_performance}, particularly in local state estimation. 
This validates the improvement in event-based pose tracking and highlights the significance of introducing our event-based 2D-2D direct alignment method, particularly in conjunction with the feature-based method. 

\subsection{Mapping Performance in Diverse Scenarios}
\label{Mapping Performance in Diversity Scenarios}

In this section, we evaluate the mapping performance of our EVI-SAM across diverse platforms and scenarios, including handle-held devices (such as rpg~\cite{zhou2018semi}, DAVIS240C~\cite{GWPHKU:event-camera-dataset_davis240c}, HKU-dataset), flying drones (MVSEC~\cite{GWPHKU:MVSEC}), and driving vehicles (DSEC~\cite{GWPHKU:DSEC}).
These datasets provide event streams, grayscale frames, and IMU data.
The event stream has a resolution of $240\times180$ pixels for rpg~\cite{zhou2018semi} and DAVIS240C~\cite{GWPHKU:event-camera-dataset_davis240c}, $346\times260$ pixels for DAVIS346 in MVSEC~\cite{GWPHKU:MVSEC} and HKU-dataset, $640\times480$ pixels for DSEC~\cite{GWPHKU:DSEC}.
Additionally, These datasets span a wide range of scenarios, from indoor to outdoor autonomous driving, featuring scenes with varying texture richness, ranging from well-lit to dimly lit conditions, and encompassing both dynamic and static objects. 
We evaluate the full EVI-SAM system utilizing the event-based hybrid tracking module to provide pose feedback and demonstrate the generalization capability of our event-based dense mapping.

\begin{figure*}[htb]  
        \centering
        \includegraphics[width=2.0\columnwidth]{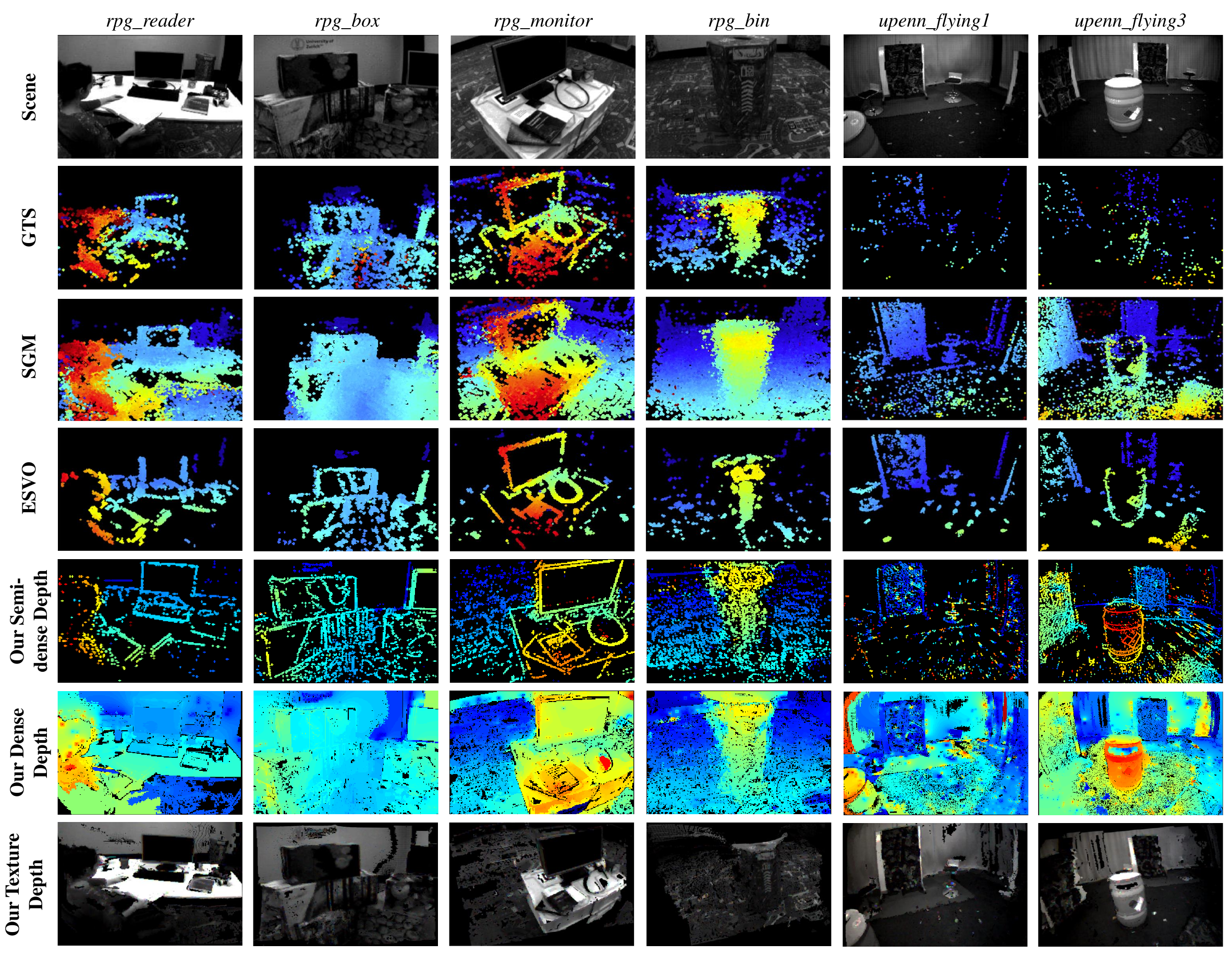}
        \captionsetup{justification=justified}
        \caption{Mapping result comparison.
        The first row shows the intensity frames from the event camera.
        Rows 2 to 4 show semi-depth estimation results from GTS~\cite{GTS,GWPHKU:ESVO}, SGM~\cite{SGM,GWPHKU:ESVO}, and ESVO~\cite{GWPHKU:ESVO}, respectively. 
        The last three rows show the estimated semi-dense depth, dense depth, and texture point cloud from our method.
        Depth maps are pseudo-colored, from red (close) to blue (far), in the range of 0.55-6.25 m for the rpg~\cite{zhou2018semi} and in the range 1.0-6.5 m for MVSEC~\cite{GWPHKU:MVSEC}.
        }  
        \label{fig:mapping_performance_comparison}
        \vspace{-1.0em}
\end{figure*}%

As some sequences lack ground truth depth, we only present the qualitative mapping results in Fig.~\ref{fig:dense_mapping_performance}.
The mapping results evident that our method does a notable job of reconstructing 3D dense structures with texture information in such diverse scenarios, thus demonstrating its excellence and generalization ability in real-world scenes.
This highlights the robustness and powerful generalization capability of our algorithm, distinguishing it from learning-based approaches~\cite{mahbub2023multimodal,chaney2019learning,hidalgo2020learning,gehrig2021combining,zhu2019unsupervised,tulyakov2019learning,mostafavi2021event,ahmed2021deep} that typically assess only a limited number of sequences within the same scene.
The key distinction lies in their usual requirement for scene-specific pre-training or even re-designing when applied across various scenes.
Additionally, as evident in Figs.~\ref{fig:dsec_zurich_city_04_a} -~\ref{fig:dsec_zurich_city_04_f},
the geometry of 3D scenes and the textures of buildings or vehicles have been effectively reconstructed.
Notably, in Fig.~\ref{fig:dsec_zurich_city_04_e}, the reconstruction of the crosswalk on the road is particularly well-executed.
In the dense depth, the black areas indicate instances where depth recovery was unsuccessful. 
For instance, the sky areas in the illustrations correspond to regions that are too far for depth recovery. 
Additionally, certain textureless regions of the road, lacking sufficient event data, may also pose challenges for recovering dense depth.
However, it is essential to recognize that such occurrences are caused by inherent characteristics of event cameras.
These situations would not occur in indoor environments. Our event-based dense mapping effectively recovers the complete dense depth along with the details of the structure.
For instance, as illustrated in Fig.~\ref{fig:poster_translation}, our method adeptly reconstructs the texture of the poster, and the consistency of the wall depth is excellent.
On the flat wall surface, most areas in our event-based dense map display similar colors, representing proximity and uniform distances across those regions.

\subsection{Mapping Performance Comparison with Baselines}
\label{Mapping Performance Comparison with Baselines}

\subsubsection{Qualitative Comparison}

In this section, we conduct comparisons with other event-based depth estimation approaches using sequences from MVSEC~\cite{GWPHKU:MVSEC} and rpg~\cite{zhou2018semi} datasets.
Fig.~\ref{fig:mapping_performance_comparison} shows inverse depth maps produced by our EVI-SAM and baseline methods.
The first row displays raw images of the selected reference views.
The second to the fourth row show inverse depth maps generated by GTS~\cite{GTS,GWPHKU:ESVO}, SGM~\cite{SGM,GWPHKU:ESVO}, and ESVO~\cite{GWPHKU:ESVO}, respectively.
The last three rows present the semi-dense, dense, and textured depths generated by our EVI-SAM. 
As anticipated, since event cameras solely respond to the apparent motion of edges, the methods that produce semi-dense depth maps only depict edges of the 3D scene.
In contrast, our EVI-SAM can recover texture and surface information to obtain dense depth maps.
The results of GTS~\cite{GTS,GWPHKU:ESVO}, SGM~\cite{SGM,GWPHKU:ESVO}, and ESVO~\cite{GWPHKU:ESVO} are borrowed from the respective papers.
For comparison, we select the reference views that are similar to those used in ESVO~\cite{GWPHKU:ESVO}.
It is worth noting that, unlike our method, the baseline methods exclusively rely on known camera poses provided by an external motion capture system to run the mapping process.
In contrast, our mapping approach relies on the pose feedback provided by our event-based hybrid tracking modules to achieve depth estimation.
Additionally, our EVI-SAM utilizes the monocular event camera, while the baseline methods employ stereo event cameras. 
However, it still can be observed that our event-based mapping method gives the best results in terms of the overall depth recovery performance and the density percentage.

\subsubsection{Quantitative Comparison}
\label{section: Quantitative Comparison}
Table~\ref{table:mapping_comparison} details the depth errors associated with various event-based depth estimation methods in MVSEC~\cite{GWPHKU:MVSEC} flying drone sequences.
We utilize mean depth error as our evaluation metric, computed by comparing estimated depth values with the ground truth depth obtained through 3D LiDAR.
For the semi-dense methods, we validate them using their open-source code and known camera poses provided by the dataset. 
We choose estimated depths from each baseline that closely match the same ground truth depth for assessment.
To ensure a fair comparison with the dense ground truth depth, we assign zero values to regions where depths fail to recover and calculate the density percentage of the semi-dense depth estimates.
As can be seen from Table~\ref{table:mapping_comparison}, our method performs better than these baselines in terms of dense depth recovery.
To our knowledge, this is the first successful attempt to compute monocular dense depth results for event cameras without using learning-based methods.
Note that the reported results of these event-based semi-dense methods are inferior to those presented in original papers. 
This discrepancy is attributed to the usage of ground truth dense depth rather than ground truth sparse depth.
Meanwhile, it should be noted that the ground truth dense depth provided by MVSEC may not be sufficiently accurate.
Firstly, the low frequency of the ground truth depth poses a challenge to time alignment, particularly during camera motion.
Secondly, errors in the ground truth depth could be introduced during the generation process, stemming from the odometry procedure. 
This may cause objects in the ground truth depth to appear inflated compared to their original versions in the image/event space.
These factors constrain the ability to achieve accurate validation of mapping performance.

\begin{table}[htbp]
        \begin{center}
        \captionsetup{justification=justified}
        \caption{Quantitative comparison of mapping performance in terms of mean depth error (m) and the density percentage of the recovery (\%), which is assessed against the dense ground truth depth.}
        \label{table:mapping_comparison}
        \resizebox{\columnwidth*1}{!}
        { 
        \begin{threeparttable}
        \renewcommand{\arraystretch}{1.0}
        \setlength{\tabcolsep}{1.0mm}
        \begin{tabular}{ccccc} 
        \hline  
        \multicolumn{1}{c}{\multirow{2}*{\textbf{Methods}}}
        & \multicolumn{1}{c}{\multirow{2}*{\textbf{Types}}}
        & \multicolumn{1}{c}{\multirow{1}*{\textbf{indoor\_flying1}}}
        & \multicolumn{1}{c}{\multirow{1}*{\textbf{indoor\_flying2}}}
        & \multicolumn{1}{c}{\multirow{1}*{\textbf{indoor\_flying3}}}\\
        \cline{3-5}
        & &  \multicolumn{3}{c}{\multirow{1}*{\textbf{Mean Error(m)} / \textbf{Density Percentage (\%)}}}\\
        \hline
        EMVS~\cite{GWPHKU:EMVS}     & Mono       & 2.38 / 6.01\%  & 1.82 / 4.90\%  & 2.82 / 4.93\%\\
        Ref.~\cite{ghosh2022multi}  & Stereo     & 2.46 / 1.79\%  & 1.85 / 1.35\%  & 2.88 / 1.85\% \\
        ESVO~\cite{GWPHKU:ESVO}     & Stereo     & 2.43 / 3.29\%  & 1.80 / 4.00\%  & 2.88 / 2.34\% \\
        ~ & Event-only stereo     & \textit{failed}  & \textit{failed}  & 0.26 / 100\% \\
        Ref.~\cite{mostafavi2021event}  & Image-only stereo     & \textit{failed}  & \textit{failed}  & 0.24 / 100\% \\
        ~ & Event-Image stereo     & \textit{failed}  & \textit{failed}  & 0.22 / 100\% \\
        \textbf{Ours}              & Mono       & 0.78 / 100\% & 0.82 / 100\% & 0.86 / 100\% \\
        \hline
        \end{tabular}
        \end{threeparttable} 
        }
        \end{center}
\end{table}

Furthermore, Table~\ref{table:mapping_comparison} also provides the depth error of a learning-based method~\cite{mostafavi2021event}.
Since we followed the same evaluation protocol as Ref.~\cite{mostafavi2021event}, we can directly report the raw results from the original paper.
Despite this method being numerically superior to the other non-learning methods, it requires scene-specific pre-training for each data sequence and high-level hardware resources such as GPU.
It is worth noting that when evaluating the indoor\_flying3, both indoor\_flying1 and indoor\_flying2 sequences are used to train the network.
Additionally, only around 1000 frames from the entire rosbag, containing over 3000 frames, can be used for testing.
In contrast, our EVI-SAM can process the entire rosbag of indoor\_flying1 to indoor\_flying3 without any pre-training. 
Therefore, we only report the raw result of indoor\_flying3 from Ref.~\cite{mostafavi2021event}, and label the result of indoor\_flying1 and indoor\_flying2 as failed since it only works under specified pre-training conditions.
The qualitative comparison with this learning-based method is available in APPENDIX~\ref{appendices:Mapping Performance Comparison with Learning-based Methods}.

\subsection{Mapping Performance in Challenge Situations}
\label{Mapping Performance in Challenge Situations}
The tracking performance of our EVI-SAM in the scenarios involving aggressive motion and HDR has been effectively evaluated in Section~\ref{Tracking Performance in Aggressive Motion and HDR Scenarios}.
In this section, we specifically evaluate the mapping performance of our EVI-SAM in challenging conditions.

\subsubsection{Mapping Performance in HDR}
\label{Mapping Performance in HDR}

\begin{figure}[htbp]  
        \subfigbottomskip=0pt 
        \subfigcapskip=-3pt 
        \centering
        \captionsetup{justification=justified}
        \subfigure[Dense street]{
            \centering
            \includegraphics[width=0.32\columnwidth]{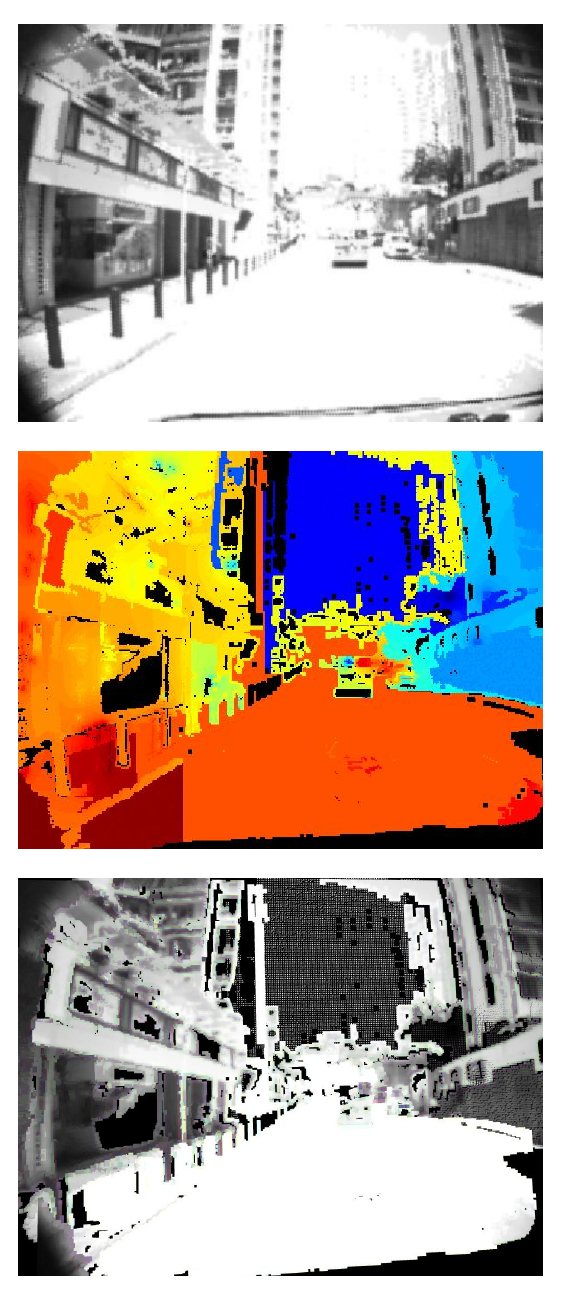}
            \label{hdr_mapping_dense_street}
        }
        \hspace{-0.5cm}
        \subfigure[Tunnel]{
            \centering
            \includegraphics[width=0.32\columnwidth]{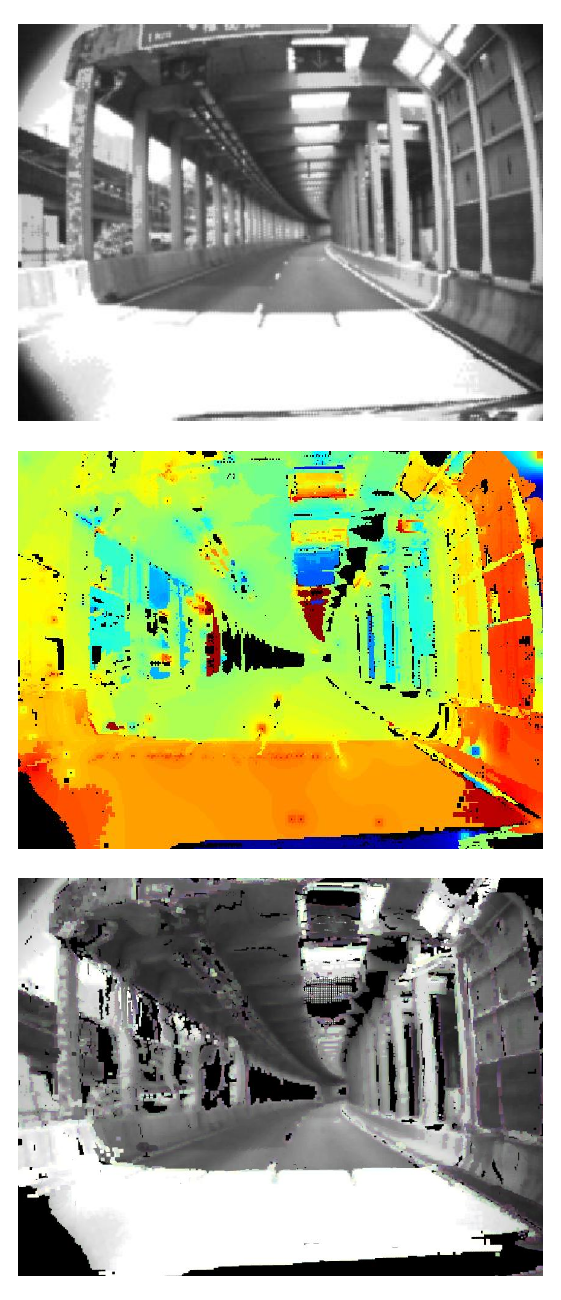}
            \label{hdr_mapping_tunnel}
        }
        \hspace{-0.5cm}
        \subfigure[Suburban road]{
            \centering
            \includegraphics[width=0.32\columnwidth]{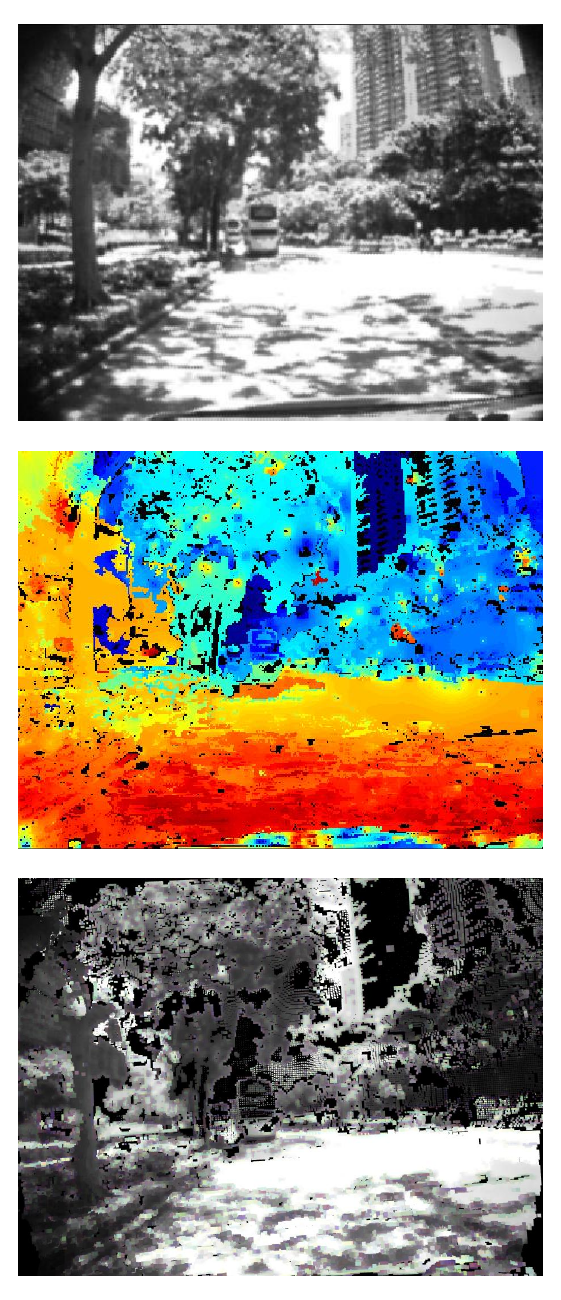}
            \label{hdr_mapping_suburban_road}
        }
        \caption{
         Mapping result of EVI-SAM under HDR scenes.
        The first row shows the intensity frames from the event camera. 
        The second row shows the event-based dense maps, and the last row is the event-based dense point clouds with texture information.
        } 
        \label{fig:hdr_mapping}
\end{figure}

In this experiment, we demonstrate the effectiveness of our event-based dense mapping in HDR scenarios using an event-based driving dataset (ECMD~\cite{GWPHKU:ECMD}). 
We specifically choose three sequences (Dense\_urban, Tunnel, and Suburban\_road) that exhibit HDR scenes caused by the intense illumination change.
Fig.~\ref{fig:hdr_mapping} illustrates the successful recovery of dense depth by our event-based dense mapping under challenging conditions, including intense sunlight and HDR encountered in tunnels.
Since the texture information is adopted from the image measurements, the event-based dense point clouds exhibit overexposure caused by strong sunlight.
While our event-based dense map would not be affected by the illumination.
Besides, as shown in Fig.~\ref{fig:HKU_hdr_box}, our event-based dense mapping is capable of recovering dense depth even in low-light scenarios.
Despite these achievements, addressing nighttime driving scenarios proves to be a significant challenge. 
This challenge arises from the expansive and dark characteristics of nighttime driving environments.
Our event-based dense point cloud derives texture information from image measurements, resulting in the point cloud that appears pitch-black in such conditions.
In conclusion, our event-based dense mapping remains effective in recovering depth under HDR scenarios. 
Nevertheless, it is crucial to acknowledge that HDR conditions may impact the texture of our dense point cloud. 

\begin{figure*}[htb]  
        \centering
        \includegraphics[width=2.0\columnwidth]{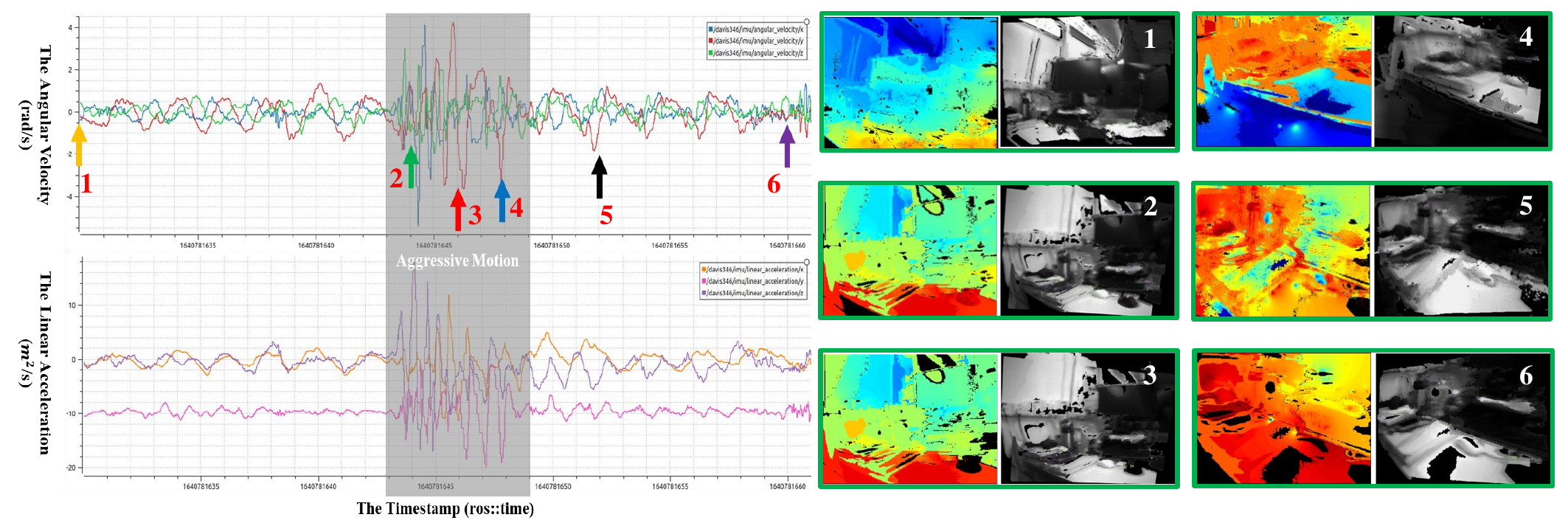}
        \captionsetup{justification=justified}
        \caption{
        Mapping results of EVI-SAM under aggressive motions.
        We visualize the results of our event-based dense mapping along with the timestamps (ROS time), and the raw gyroscope and acceleration reading from the IMU. 
        The shaded area in grey represents the phase of aggressive motions.
        }  
        \label{fig:mapping under aggressive motion}
        \vspace{-1.0em}
\end{figure*}%

\subsubsection{Mapping Performance in Aggressive Motions}
\label{Mapping Performance in Aggressive Motion}

In this experiment, we examine the robustness of our event-based dense mapping in scenarios involving aggressive motions (see Fig.~\ref{fig:mapping under aggressive motion}), where the maximum angular velocity reaches up to 5 $rad/s$ (approximately 290$\degree /s$), and the linear acceleration reaches up to 18 $m^{2}/s$ (see the gyroscope and acceleration readings in the left part of Fig.~\ref{fig:mapping under aggressive motion}).
The results of our estimated event-based dense depths and texture point clouds are also presented alongside timestamps, with the aggressive motion phase shaded in grey blocks.
It can be observed that our event-based dense mapping consistently and stably reconstructs the 3D structure of the scene even in the presence of intense rotation and translation motions.
These results underscore the robustness of our algorithm and demonstrate its capacity to handle scenarios with aggressive motions. 
For visual demonstrations of our mapping results under aggressive motions, please refer to the video demo.

\subsection{Full System Onboard Evaluation}
\label{sectioin: Full System Onboard Evaluation}
Although the data sequence utilized in the previous section is appropriate for assessing the performance of pose tracking and mapping in challenging situations, it does not encompass real-time onboard evaluation and global mapping.
Moreover, we lack a dataset for comparing our event-based dense mapping with the depth camera.
To fill this gap, we design an event-based handheld device and evaluate the onboard performance of our EVI-SAM.
Please note that the depth maps provided by the depth camera only serve as references for qualitative comparison.

\begin{figure}[htb]  
    \captionsetup{justification=justified}
    \centering
    \includegraphics[width=0.80\columnwidth]{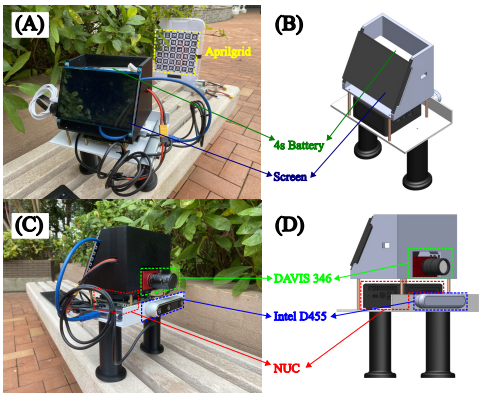}
    \caption{
    The event-based handheld device with the schematics model for onboard evaluation. 
    Please note that the RGB-D camera (Intel D455) is only used for reference, the complete system of our EVI-SAM operates exclusively with the monocular event camera (DAVIS346).
    }
    \label{fig:handle_device}
    \vspace{-1.0em}
\end{figure}%

\subsubsection{Event-based Handheld Device}
\label{section:Event-based Handheld Device}
Our handheld device is shown in Fig.~\ref{fig:handle_device}, which includes a power supply unit, an onboard computer NUC (equipped with Intel i7-1260P, 32GB RAM, and Ubuntu 20.04 operation system), a DAVIS346 event camera, and a Intel® RealSense™ D455 RGB-D camera.
All mechanical modules of this device are designed for 3D printing, and both design schematics and the collected data sequence are available on our GitHub repository:\url{https://github.com/arclab-hku/Event_based_VO-VIO-SLAM/blob/main/EVI-SAM/data_srcipt.md}

\begin{figure}[htb]  
        \subfigcapskip=-10pt 
        \centering
        \captionsetup{justification=justified}
        \subfigure[Depth from RGB-D camera]{
            \begin{minipage}[t]{0.47\columnwidth}
            \centering
            \includegraphics[width=1.0\columnwidth]{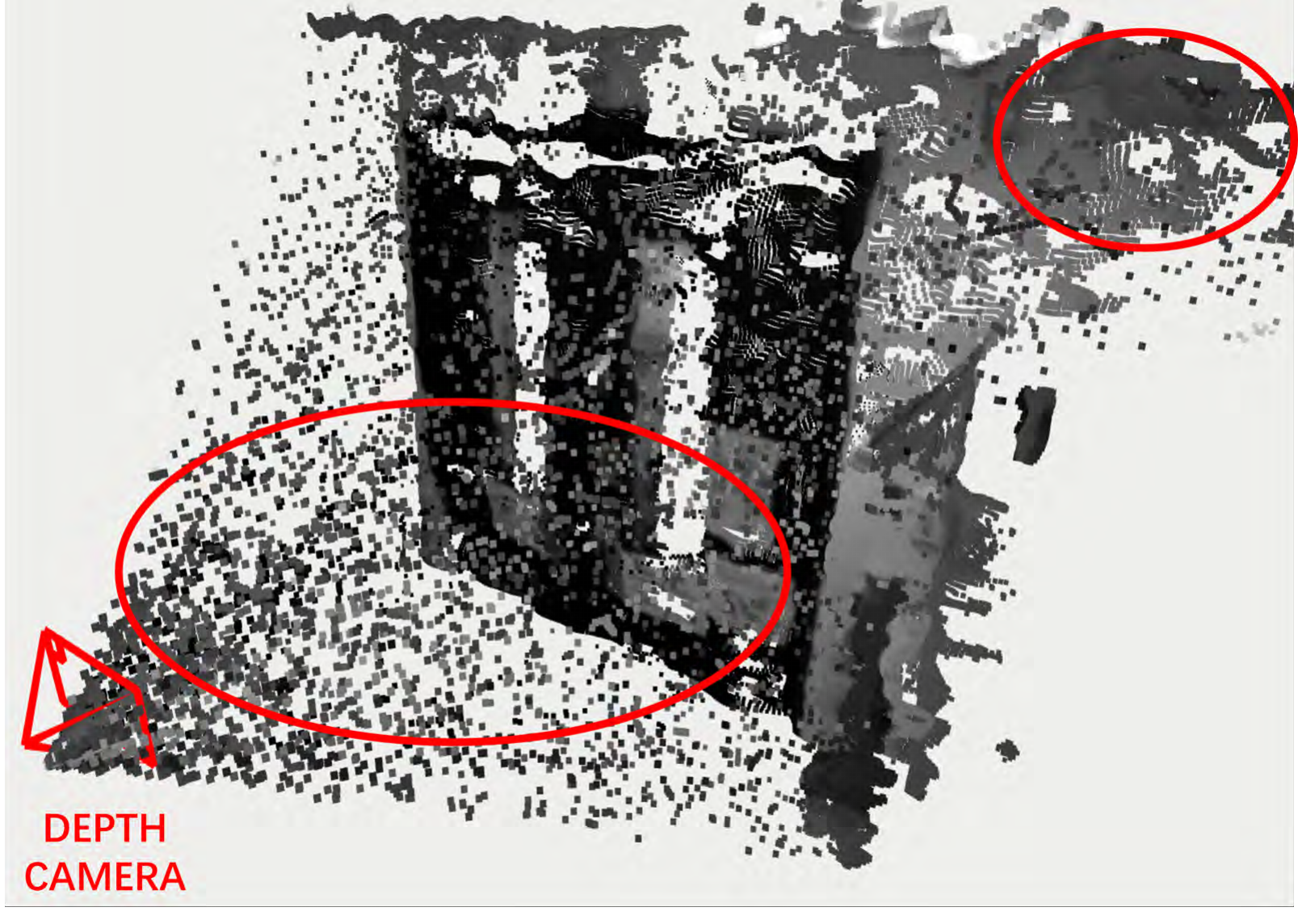}
            \label{fig:depth_from_rgbd}
            \end{minipage}%
        }        
        \subfigure[Depth from our EVI-SAM]{
            \begin{minipage}[t]{0.47\columnwidth}
            \centering
            \includegraphics[width=1.0\columnwidth]{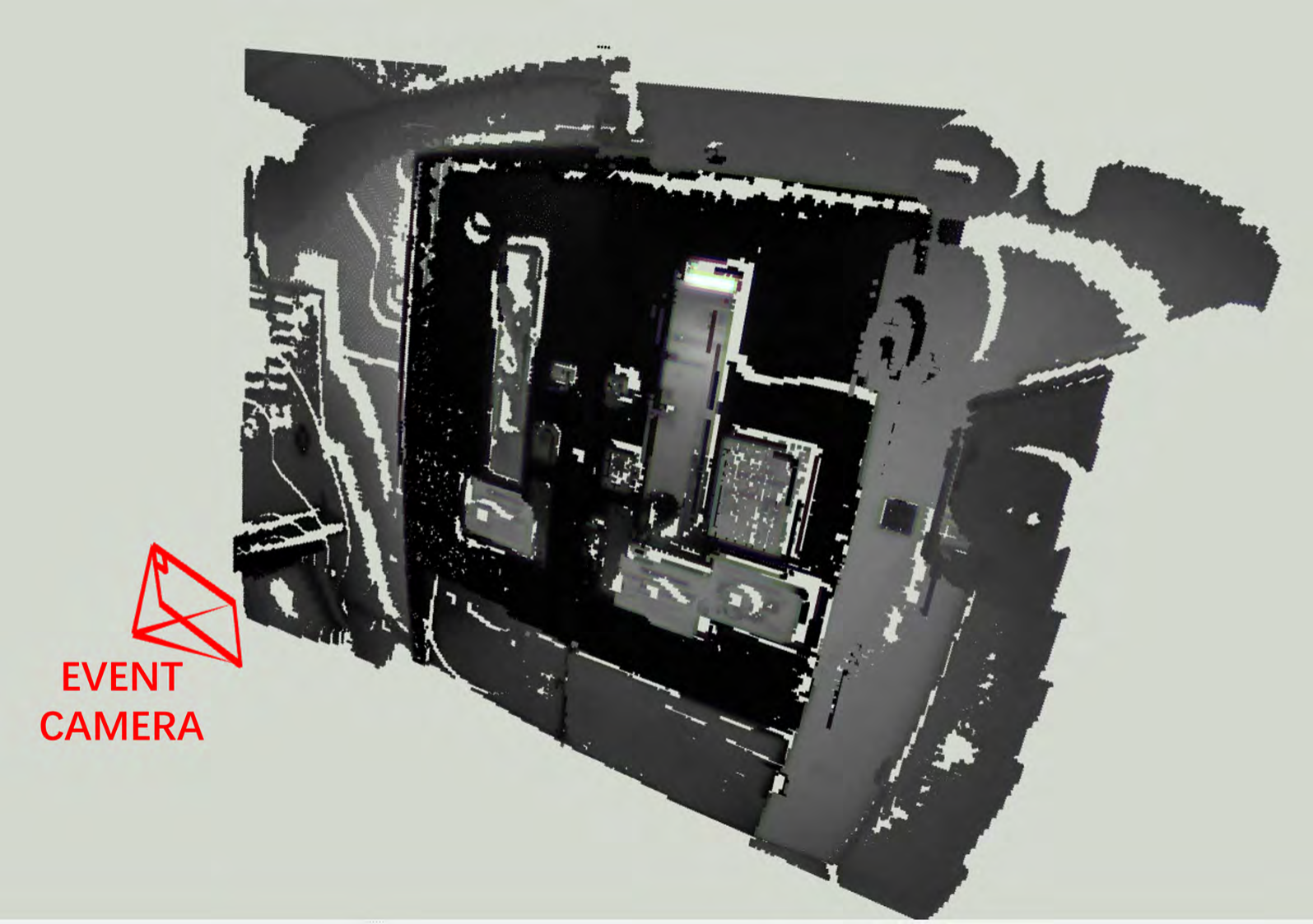}
            \label{fig:depth_from_dvs}
            \end{minipage}%
        }
        \caption{
        Comparing the local texture depth generated by EVI-SAM with the one produced by the RGB-D camera.
        }  
        \label{fig:depth_comparison}
\end{figure}%

\subsubsection{Local Mapping Performance}
\label{section: Local Mapping Performance}
\begin{figure*}[htb]  
        \centering
        \includegraphics[width=2.0\columnwidth]{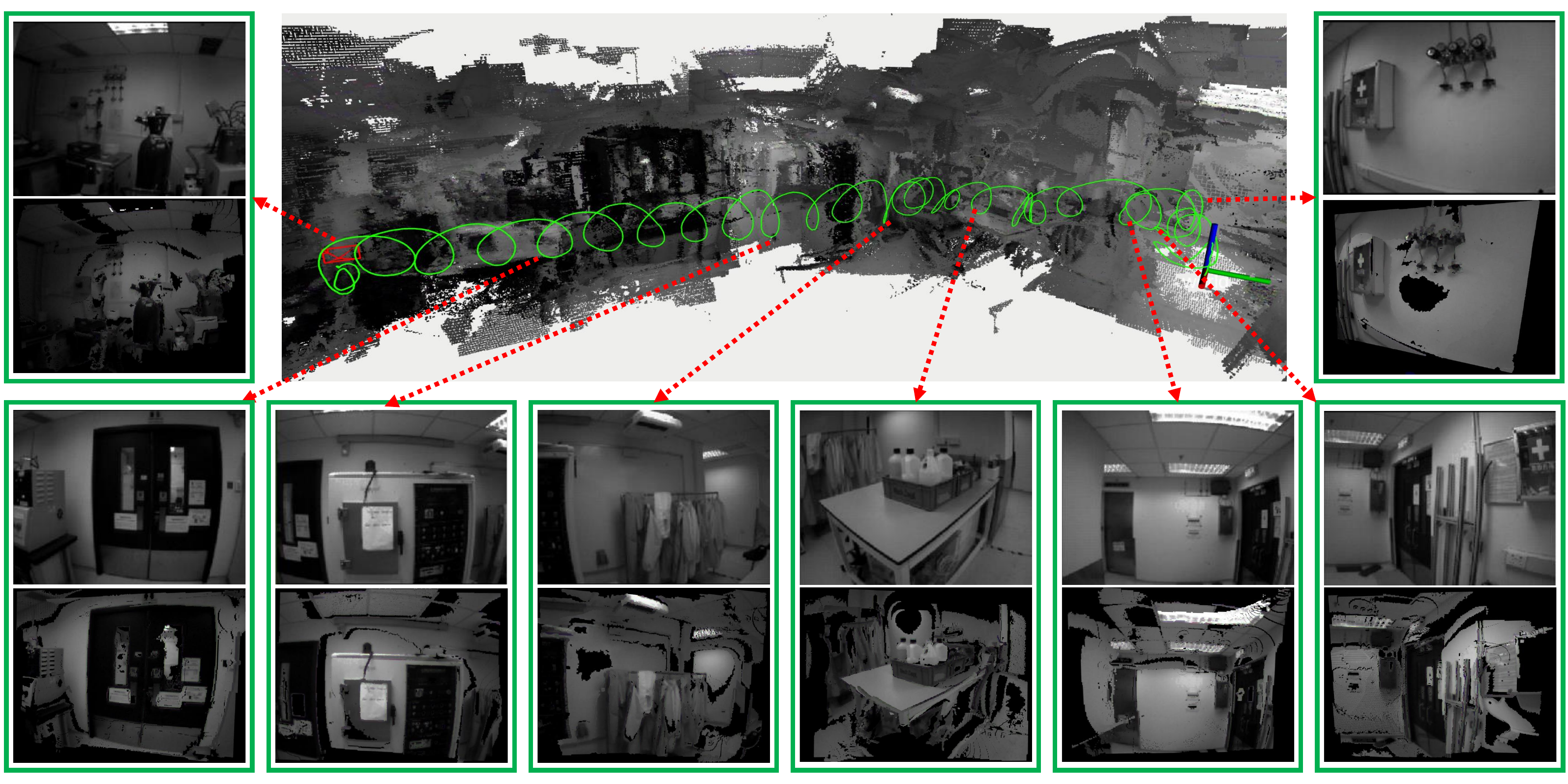}
        \captionsetup{justification=justified}
        \caption{
        Visualization of the estimated camera trajectory and global 3D reconstruction (surface mesh) of our EVI-SAM. 
        Sequentially display from right to left includes the event-based dense point clouds with texture information and intensity images, at selected viewpoints.
        }  
        \label{fig:global_mapping}
\end{figure*}

We conduct real-time testing of EVI-SAM using the handheld device within the LG01\_lab of the Haking Wong Building at The University of Hong Kong.
The results, including event-based dense mapping at selected viewpoints and the estimated trajectory, are illustrated in Fig.~\ref{fig:global_mapping}. 
Three live demonstrations are also available at~\url{https://b23.tv/cwetpxL}.
Additionally, we conduct qualitative comparisons between the dense mapping results from our EVI-SAM and the depth images from the RGB-D camera.
To ensure a fair comparison, we render the texture information from the intensity image of DAVIS346 onto the depth image of the RGB-D camera, maintaining the same resolution as the DAVIS346.
The depth images are directly acquired from the RGB-D camera, with its infrared ranging sensor obscured.
We directly align the texture point clouds from the RGB-D camera along with the estimated pose of our tracking module. 
As observed in the highlighted area within the red circle in Fig.~\ref{fig:depth_comparison}, it is evident that our proposed event-based dense mapping approach demonstrates performance comparable to commercial depth cameras.
The reconstructed depths from the RGB-D camera are notably noisy and cluttered, lagging far behind the results obtained from our EVI-SAM.
Moreover, it exhibits superior noise reduction and texture recovery capabilities.
More dense mapping performance comparisons between our EVI-SAM and the RGB-D camera can be found in the APPENDIX~\ref{appendices:Mapping Performance Comparison with Traditional Image-based Methods}.

\subsubsection{Global Mapping Performance}
The global mapping performance of our EVI-SAM is also illustrated in Fig.~\ref{fig:global_mapping}, showing the surface mesh generated through TSDF-based map fusion.
Our global event-based dense mapping exhibits excellent global consistency. 
The video demo in Section~\ref{section: Local Mapping Performance} also illustrates the incremental reconstruction of the event-based surface mesh from updated TSDF voxels. 
This process enables on-demand meshing for visualization, allowing flexibility in generating the mesh at any time.
Our TSDF-based map fusion for global mapping is designed to generate surface meshes that enable humans to assess the 3D reconstructed environment more effectively. 
This capability supports high-level mission goals, such as collision-free motion planning.
During evaluation, an onboard computer NUC is utilized to support real-time pose estimation and local event-based dense mapping.
However, the NUC lacks sufficient computational power to support a real-time meshing process.
Therefore, we utilize a personal computer (Intel i7-11800H, 32GB RAM) without GPU to output the global mesh of EVI-SAM for the global mapping evaluation.
Further details regarding the time-consuming aspects of different modules will be discussed in Section~\ref{Running Time Analysis}. 

\subsection{Running Time Analysis}
\label{Running Time Analysis}
In this section, we investigate the average time consumption of our proposed system on the CPU-only NUC.
We calculate the average time consumption during the evaluation of Section~\ref{sectioin: Full System Onboard Evaluation}, whose results are shown in Table~\ref{table:time_cost}.
The hybrid tracking module takes an average of 55.24 ms with an additional 3.75 ms for event-corner feature tracking, and the dense mapping module takes an average of 137.45 ms processing time.
The estimated pose frequency can be effortlessly boosted to IMU-rate by directly integrating the latest estimation with IMU measurements in a loosely-coupled manner.
Meanwhile, the dense map generation frequency is around 7 Hz, which should be real-time enough for tasks like collision-free motion planning, e.g. obstacle avoidance.

\begin{table}[htbp]
        \begin{center}
        \caption{The average time consumption of different modules in our EVI-SAM}
        \label{table:time_cost}
        \resizebox{\columnwidth*1}{!}
        { 
        \begin{threeparttable}
        \renewcommand{\arraystretch}{1.0}
        \setlength{\tabcolsep}{1.0mm}
        \begin{tabular}{cccc} 
        \hline  
        Sequence 
        & \makecell{\textbf{Event-corner tracking} \\ \textbf{Mean} / \textbf{Std (ms)}}
        & \makecell{\textbf{Hybrid tracking} \\ \textbf{Mean} / \textbf{Std (ms)}}
        & \makecell{\textbf{Dense mapping} \\ \textbf{Mean} / \textbf{Std (ms)}}\\
        \hline
        Logo\_wall\_1      & 3.50 / 1.17 & 54.39 / 23.35 & 138.21 / 79.49\\
        Logo\_wall\_2     &  3.75 / 1.03 & 55.11 / 23.01 & 123.8 / 49.24  \\
        LG\_Factory      & 4.49 / 1.66 & 67.29 / 54.90 & 129.63 / 44.54\\
        LG\_office      &  3.89 / 1.37& 55.10 / 31.88 & 112.09 / 40.96 \\
        Fountain\_1      & 3.75 / 1.15 & 49.78 / 15.76 & 141.46 / 48.59 \\
        Fountain\_2      & 3.11 / 1.14 & 49.79 / 8.57 & 179.49 / 106.68 \\
        \hline
        \textbf{Average} & 3.75 / 1.25& 55.24 / 26.24& 137.45 / 61.57\\
        \hline
        \end{tabular}
        \end{threeparttable} 
        }
        \end{center}
        \vspace{-2.0em}
\end{table}

\subsection{Limitations and Discussions}
\label{Discussion of Limitations}

Regarding our event-based dense mapping, one limitation lies in the difficulty of attaining comprehensive and precise recovery of dense depth across various environments.
This is caused by two primary factors.
The first factor is the inherent nature of event cameras, which produce sparse pixels and only respond to moving edges.
The second factor is attributed to scenes being too distant from the camera, such as the sky in a driving scenario.
Conversely, when the event camera operates in structured environments where depth patterns exhibit regularity, such as many planar surfaces with limited edges, or scenarios with sharp object boundaries, our method can leverage this regularity to infer good dense depth from a small number of event measurements.
Besides, our event-based dense mapping is a kind of color-guided depth enhancement method.
This kind of approach is under the assumption that the edges of the event-based semi-dense depth and the color edges at the corresponding locations are consistent.
Therefore, the incorrect guidance from the companion color image will lead to texture-copy artifacts and blurring depth edges on the reconstructed dense depth.
While various methods~\cite{choi2014consensus,xiang2015no} exist to address this issue, we intend to defer this challenge to future works.

Another limitation of our event-based dense mapping is the inevitable occurrence of artifacts.
The term "artifacts" pertains to specific errors in scale recovery, such as cases where small areas in distant regions are mistakenly reconstructed as closer ones.
This may be attributed to the implementation of the monocular MVS problem in our approach, resulting in a loss of scale. 
Consequently, in future work, we may extend our algorithm to incorporate a stereo setup to address this issue.

Certainly, image-based dense reconstruction has been thoroughly studied and achieved promising results. 
However, the exceptional advantages of event cameras underscore the importance of exploring event-based dense reconstruction, especially considering that non-learning methods for event-based dense reconstruction remain an unexplored territory.
Our method generates dense depths from event streams, with images only serving as guidance. 
Thus the reconstruction results may not match the impressive performance of some state-of-the-art RGB-only or RGB-D methods. 
Besides, our method does not rely on images to generate dense depth, instead, dense depths are generated from the event stream with images only serving as guidance.
Therefore, the reconstruction results may not be as impressive as some state-of-the-art RGB-based methods.
Nevertheless, our focus lies in addressing dense reconstruction using event cameras, which offer a potential solution for challenging scenarios not adequately addressed by standard images.
This work not only aims to achieve optimal mapping performance in challenging situations but also serves as an inspiration for the research community encouraging future research in event-based dense reconstruction. 


\section{CONCLUSIONS} 
\label{CONCLUSIONS}

In this paper, we have presented EVI-SAM, a framework designed for 6-DoF pose tracking and 3D dense mapping using the monocular event camera. 
To enhance the robustness and accuracy of 6-DoF pose tracking, we have proposed the first event-based hybrid pose tracking framework which combines the robustness of feature-based event pose tracking with the relatively high accuracy achieved through event-based direct alignment.
To achieve dense mapping, an image-guided and segmentation-based approach was proposed to reconstruct dense maps from sparse and incomplete event measurements.
To create a comprehensive representation of the 3D environment, we have designed the TSDF-based map fusion to construct both the textured global map and the surface mesh.
Our framework balances accuracy and robustness against computational efficiency towards strong tracking and mapping performance in challenging scenarios such as HDR or fast motion.
Notably, EVI-SAM demonstrates computational efficiency for real-time execution, making it suitable for onboard localization, mapping, and navigation.
Meanwhile, the generated event-based dense maps can be directly applied to collision-free motion planning, ensuring safe navigation.
In future work, we aim to explore a full event-based navigation and obstacle avoidance system based on our EVI-SAM.

\begin{appendices}
\setcounter{table}{0} 
\setcounter{figure}{0} 
\setcounter{equation}{0} 
\renewcommand{\thetable}{\thesection-\Roman{table}} 
\renewcommand{\theequation}{\thesection-\arabic{equation}} 
\renewcommand{\thefigure}{\thesection-\arabic{figure}} 

\section{More Mapping Performance Comparisons}
\label{appendices:Local Mapping Performance Comparison}

Section~\ref{Mapping Performance Comparison with Baselines} provides a quantitative and qualitative comparison of our event-based dense mapping performance with other event-based mapping baselines.
This APPENDIX further extends the mapping performance comparison with traditional image-based methods (APPENDIX~\ref{appendices:Mapping Performance Comparison with Traditional Image-based Methods}), learning-based mapping method (APPENDIX~\ref{appendices:Mapping Performance Comparison with Learning-based Methods}), and NeRF-based methods (APPENDIX~\ref{appendices:Mapping Performance Comparison with NeRF-based Methods}), respectively.

\begin{figure*}[htbp]  
        \centering
        \includegraphics[width=2.0\columnwidth]{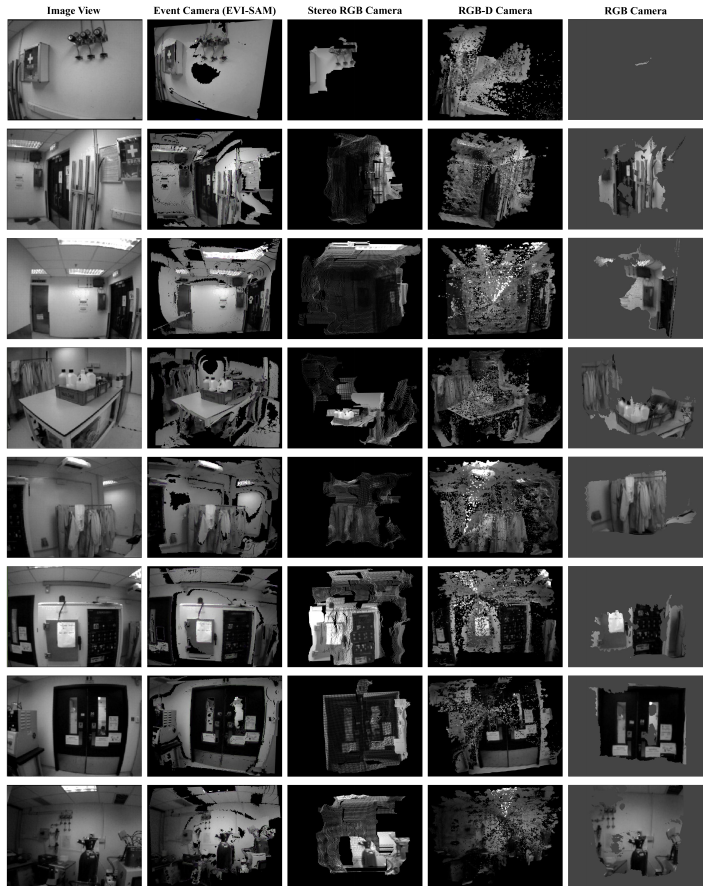}
        \captionsetup{justification=justified}
        \caption{
        Local mapping performance comparison. 
        The first column shows the intensity image of the selected view.
        Column 2 to 5 show the color point cloud results from our EVI-SAM, stereo RGB camera~\cite{image_undistort}, RGB-D camera~\cite{intelrealsense}, and monocular camera~\cite{shenshaojie:quadtree}, respectively.
        }  
        \label{fig:cameras_comparison}
\end{figure*}%

\begin{figure*}[htb]  
        \subfigbottomskip=-1pt 
        \subfigcapskip=-10pt 
        \centering
        \captionsetup{justification=justified}
        \subfigure[ ]{
            \begin{minipage}[t]{1.0\columnwidth}
            \centering
            \includegraphics[width=1.0\columnwidth]{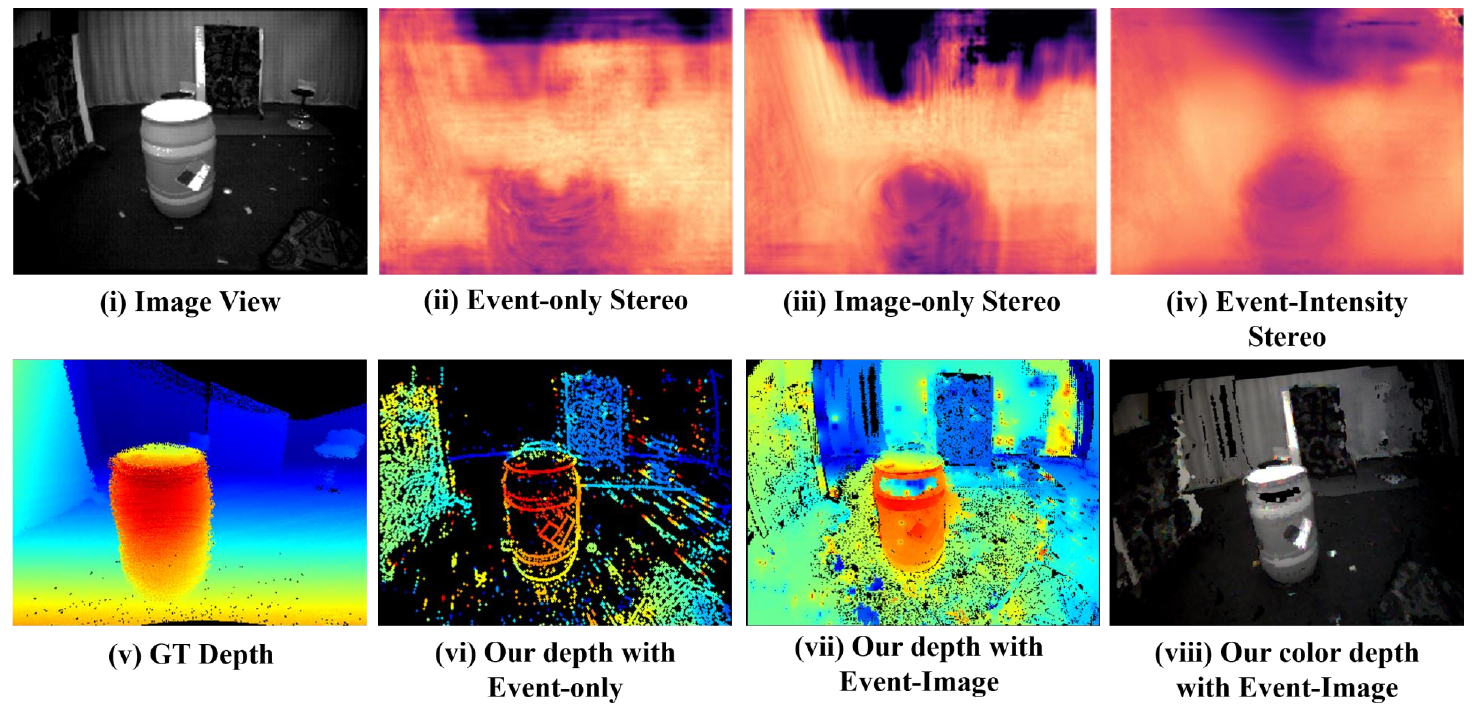}
            \label{fig:compare_with_learning1}
            \end{minipage}%
        }        
        \subfigure[ ]{
            \begin{minipage}[t]{1.0\columnwidth}
            \centering
            \includegraphics[width=1.0\columnwidth]{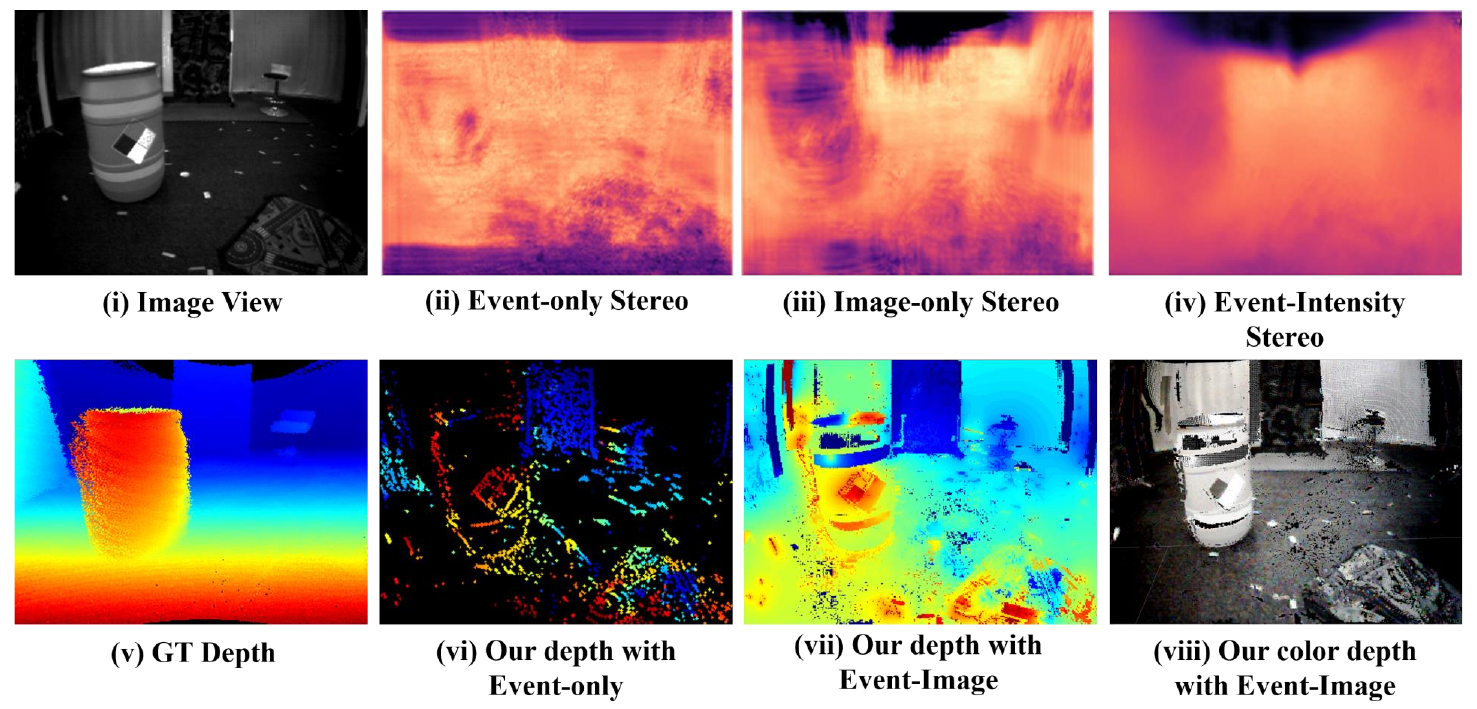}
            \label{fig:compare_with_learning2}
            \end{minipage}%
        }
        \caption{
       Qualitative comparison of our EVI-SAM and the learning-based method~\cite{mostafavi2021event} in MVSEC~\cite{GWPHKU:MVSEC} dataset.
       For each element, the first row is the (i) image view, and the estimated depth of learning-based method, which utilizes the (ii) event-only, the (iii) image-only, and the (iv) event-image methods;
       The second row is the (v) ground truth depth obtained through 3D LiDAR, the estimated depth of our EVI-SAM using (vi) event-only and (vii) event-image, and our estimated depth with texture information.
        }  
        \label{fig:compare_with_mapping}
        \vspace{-1.0em}
\end{figure*}%

\subsection{Mapping Performance Comparison with Traditional Image-based Methods}
\label{appendices:Mapping Performance Comparison with Traditional Image-based Methods}

We conduct qualitative comparisons of the depth estimation results presented as texture point clouds obtained from our EVI-SAM, with the depth estimation results from stereo RGB camera, monocular RGB camera, and RGB-D camera serving as the baseline.
We used the data sequence "LG\_office", which features fast motion, from Section~\ref{section:Event-based Handheld Device}.
To ensure a fair comparison, firstly, we only employ lightweight and real-time methods to estimate depth from standard cameras, as opposed to offline or complex approaches that rely on GPU acceleration. 
Since our EVI-SAM operates in real-time solely on the CPU.
Secondly, we resize the resolution of the depth generated from the baselines to match the resolution of our event camera (346 
$\times$ 260). 
Thirdly, we render the texture information from the intensity image of the DAVIS346 onto the estimated depth to generate the textured point cloud.
More details these baseline methods are as follows:
\begin{itemize}
\item RGB-D camera. The results are directly obtained from the depth image of Realsense D455~\cite{intelrealsense}, with its infrared ranging sensor obscured. While the texture information is transformed from the intensity image of the DAVIS346.
\item Stereo RGB cameras. The results are generated by the image\_undistort tool~\cite{image_undistort}.
\item Monocular RGB camera. The results are generated by using the monocular dense mapping algorithm in Ref.~\cite{shenshaojie:quadtree} using GPU acceleration for real-time running.
\item All of these baseline methods use the monocular image+IMU (VINS-MONO~\cite{GWPHKU:VINS-MONO}) to provide pose estimation.
\end{itemize}

As can be seen from Fig.~\ref{fig:cameras_comparison}, our EVI-SAM performs better than these baselines in both the accuracy of depth recovery and the integrity of texture information.
The textured point cloud obtained from the RGB-D camera is directly captured from a commercial Realsense camera.
However, its infrared ranging sensor is obstructed, as infrared light can interfere with the perception of event cameras.
When the infrared ranging sensor is blocked, the depth map produced by the Realsense camera tends to contain a considerable amount of noise, as illustrated in column 4 of Fig.~\ref{fig:cameras_comparison}. 
This observation matches our expectations, as demonstrated in Section~\ref{section: Local Mapping Performance}, where it is evident that the depth map generated by the RGB-D camera exhibits more noise compared to that of our EVI-SAM.

Regarding the results from monocular and stereo RGB cameras, as shown in columns 3 and 5 of Fig.~\ref{fig:cameras_comparison}, although their estimated depths exhibit less noise compared to those estimated by RGB-D cameras, the completeness of depth recovery falls far behind that of RGB-D and our EVI-SAM.
This might be caused by the challenging testing conditions characterized by varying illumination and fast motion. 
Additionally, the regions with limited texture and the drawbacks such as the narrow dynamic range of image sensor perception lead to the poor depth recovery performance of monocular and stereo RGB cameras on this data sequence.
While our EVI-SAM still can provide reliable depth perception under such challenging situations.

\subsection{Mapping Performance Comparison with Learning-based Mapping Method}
\label{appendices:Mapping Performance Comparison with Learning-based Methods}

In Section~\ref{section: Quantitative Comparison}, we conduct the quantitative comparisons with learning-based approach~\cite{mostafavi2021event}.
As shown in Fig.~\ref{fig:compare_with_mapping}, we further demonstrate the qualitative comparisons with this learning-based approach using the sequences of indoor\_flying3 from the MVSEC dataset~\cite{GWPHKU:MVSEC}.
Since the source code and pre-training model of Ref.~\cite{mostafavi2021event} are not released, the estimated depth results are borrowed from the original paper.
While the ground truth depth and our estimated depth are pseudo-colored, from red (close) to blue (far).
For comparison, we select the reference views that are similar to those in Ref.~\cite{mostafavi2021event}.
It is worth noting that this learning-based approach employs a complex network with many structures, such as recycling, deformable, and multiscale architecture.
Although its numerical results in Table~\ref{table:mapping_comparison} may appear superior to our EVI-SAM, a thorough examination of the qualitative results indicates otherwise. 
There are instances of significant noise in their estimated results, particularly when employing stereo images or stereo event setups.
Conversely, our EVI-SAM not only demonstrates superior mapping accuracy (both quantitatively and qualitatively) but also can operate without any scene-specific pre-training. 

\begin{figure*}[htbp]  
        \centering
        \captionsetup{justification=justified}
        \subfigure[ ]{
            \begin{minipage}[t]{2.0\columnwidth}
            \centering
            \includegraphics[width=1.0\columnwidth]{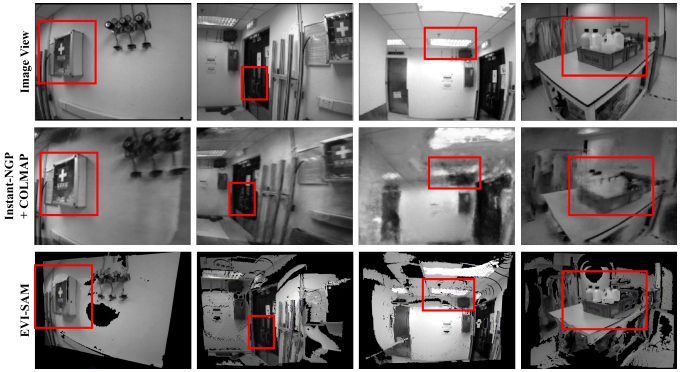}
            \label{fig:comparison_nerf_1}
            \end{minipage}%
        }
        
        \subfigure[ ]{
            \begin{minipage}[t]{2.0\columnwidth}
            \centering
            \includegraphics[width=1.0\columnwidth]{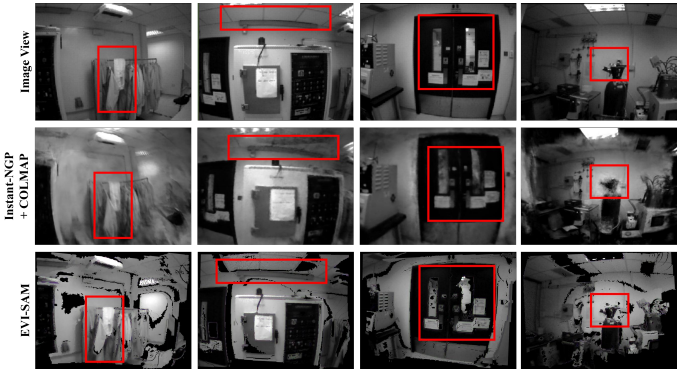}
            \label{fig:comparison_nerf_2}
            \end{minipage}%
        }
        \caption{
       Qualitative comparison of our EVI-SAM, the Instant-NGP~\cite{instant-ngp}, and the ground truth image view.
       The red box highlights the difference of the reconstruction quality and the motion blur caused by image blur in NeRF-based methods.
       While our event-based dense mapping demonstrates performance comparable to state-of-the-art NeRF-based mapping.
        }  
        \label{fig:compare_with_nerf}
\end{figure*}%

Additionally, our EVI-SAM consistently achieves good mapping results across diverse and challenging scenarios, as evidenced in Sections~\ref{Mapping Performance in Diversity Scenarios} and~\ref{Mapping Performance in Challenge Situations}.
In particular, thorough the evaluations in Section~\ref{Mapping Performance in Aggressive Motion}, our event-based dense mapping reliably reconstructs the 3D structure of scenes even under fast rotation (up to 290$\degree /s$) and translation (up to 18 $m^{2}/s$) motions.
While most of the event-based mapping works, especially for those relying on learning-based methods, often neglect to assess their performance under aggressive motion.

\subsection{Mapping Performance Comparison with NeRF-based Methods}
\label{appendices:Mapping Performance Comparison with NeRF-based Methods}

We evaluate the NeRF-based methods using the LG\_office sequence of the EVI-SAM dataset in Section~\ref{section:Event-based Handheld Device}.
Neural radiance field (NeRF) is a fully-connected neural network that can generate novel views of complex 3D scenes, based on a partial set of 2D images. 
It operates by processing input images representing a scene and interpolating between them to render one complete scene. 
Therefore, each scene must be retrained and can only be tested in the same scene. 
Failure to retrain the model when the scene changes would result in entirely incorrect outputs.
We employ the Instant-NGP~\cite{instant-ngp} for the novel view synthesis from the image of DAVIS346 and use the COLMAP~\cite{COLMAP} to estimate the camera parameters, including both intrinsics and extrinsics (pose), for each input image.
For comparing the performance of our event-based dense mapping, we visualize the view synthesis of similar viewpoints of Fig.~\ref{fig:global_mapping} and Fig.~\ref{fig:cameras_comparison}.

As highlighted with red-box in Fig.~\ref{fig:compare_with_nerf}, significant blurring effects can be observed in the reconstruction results of the NeRF-based method, which is caused by image blurring in rapid motion.
In contrast, our event-based dense mapping method effortlessly overcomes this challenge without encountering motion blur.
Additionally, the memory consumption of Instant-NGP is extremely high, making it impossible for our GPU (NVIDIA GeForce RTX 4070 Ti) to complete training with the entire sequence.
Therefore, we had to select 8 corresponding viewpoints for 8 separate training sessions. 
In other words, the result of the Instan-NGP in Fig.~\ref{fig:compare_with_nerf} stems from 8 entirely distinct training sessions, whereas our EVI-SAM can operate in real-time without the need for segmentation processing or prior training.

\begin{figure}[htbp]  
    \captionsetup{justification=justified}
    \centering
    \includegraphics[width=1.0\columnwidth]{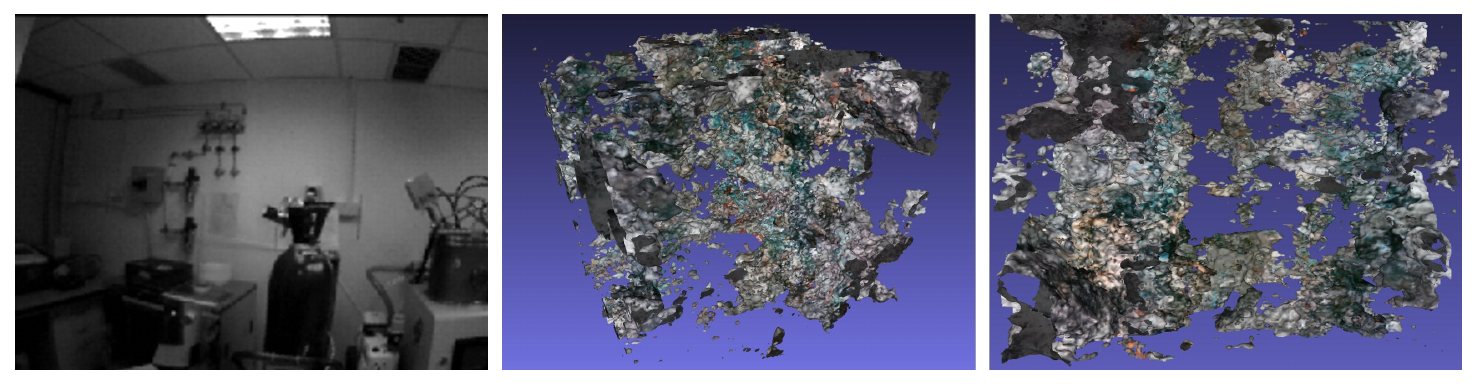}
    \caption{
    The mapping result of Co-SLAM~\cite{Co-slam} and the corresponding selected viewpoint.
    }
    \label{fig:mapping_of_nerfslam}
    \vspace{-1.0em}
\end{figure}

Furthermore, we also utilized the RGB-D data from sequence to train two state-of-the-art NeRF-based RGB-D SLAM~\cite{Nice-slam,Co-slam}.
However, due to significant noise in the depth camera and the lack of an infrared ranging sensor, depth information cannot be captured in low-texture scenes (i.e. the white wall in Fig.~\ref{fig:compare_with_nerf}).
This resulted in inadequate depth information for training both Nice-SLAM~\cite{Nice-slam} and Co-SLAM~\cite{Co-slam}.
Besides, the rapid and aggressive motion in the testing sequence scenes is too challenging for these methods, resulting in inadequate pose estimation and ultimately leading to mapping failures.
The mapping results of Co-SLAM are shown in Fig.~\ref{fig:mapping_of_nerfslam}.

\section{Ablation Study of the Pose Tracking}
\label{appendices:Ablation Study of the event-based Hybrid Pose Tracking}

\subsection{Ablation Study of Event and Image-based VIO}
\label{appendices:Ablation Study of Event and Image-based VIO}

As indicated in Section~\ref{Tracking Performance in Aggressive Motion and HDR Scenarios}, compared to standard cameras, event cameras are capable of providing reliable visual perception during high-speed motion and HDR scenarios. 
However, when event cameras have restricted relative motion, such as in static states, they may produce limited information. 
Although the image camera encounters difficulties under high-speed motion or HDR scenarios, it can provide rich-intensity textures of the scenes under uniform motion or favorable lighting conditions.
Therefore, leveraging the complementary advantages of both standard and event cameras is significant and promising, as emphasized in many previous works~\cite{GWPHKU:PL-EVIO,ESVIO,GWPHKU:Ultimate-SLAM,mostafavi2021event,gehrig2021combining,EDS,lee2023event,EKLT-VIO}.

Although the effectiveness of tightly integrating events, images, and IMU for pose estimation has been thoroughly assessed in various event-based studies, we still conduct this ablation study of various combinations within our EVI-SAM framework including event-IMU-based (EIO), image-IMU-based (VIO), and event-image-IMU-based (EVIO) configurations, as shown in Table~\ref{table: Ablation Study of event, image}.
In the first sequence, our feature-based EIO and VIO demonstrate comparable performance, thanks to our well-designed framework and the HDR level is not serious.
Conversely, in the second sequence, our feature-based EIO significantly outperforms the feature-based VIO, attributed to the high motion intensity in the testing scene. 
While our featured-based EVIO performs satisfactory results and surpasses the solely image-based VIO in both testing sequences.

\begin{table}[htbp]
        \captionsetup{justification=justified}
        \setlength{\abovecaptionskip}{-0.02cm}
        \renewcommand\arraystretch{1.2}
        \tiny 
        \begin{center}
        \caption{Accuracy Comparison of various combinations within EVI-SAM framework, including Event+IMU, Image+IMU, Event+Image+IMU, direct-based, and our EVI-SAM}
        \label{table: Ablation Study of event, image}
        \resizebox{\columnwidth*1}{!}
        { 
        \begin{threeparttable}
        \begin{tabular}{c|ccccc} 
        \hline  
        \multicolumn{1}{c|}{\multirow{2}*{Sequence}}
        & \multicolumn{3}{c}{\multirow{1}*{\textbf{Feature-based}}}
    & \multicolumn{1}{c}{\multirow{1}*{\textbf{Direct-based}}}
    & \multicolumn{1}{c}{\multirow{1}*{\textbf{Hybrid }}}\\
    & \multicolumn{1}{c}{\multirow{1}*{EIO}}  & \multicolumn{1}{c}{\multirow{1}*{VIO}} & \multicolumn{1}{c}{\multirow{1}*{EVIO}} & EVIO & EVIO\\
        \hline
        vicon\_hdr4 & 0.115 & 0.116 & 0.089& 0.080&0.092  \\
        vicon\_aggressive\_hdr& 0.572 &1.266& 0.921&0.594 & 0.683\\
        \hline        
        \end{tabular}
        \begin{tablenotes} 
        \item \textit{Unit:\%, 0.115 means the average error would be 0.115m for 100m motion; Aligning the whole ground truth trajectory with estimated poses.} 
        \end{tablenotes} 
        \end{threeparttable} 
        }
        \end{center}
        \vspace{-1.0em}
\end{table}

\subsection{Ablation Study of Direct-based and Feature-based EVIO}
\label{appendices:Ablation Study of Direct-based and Feature-based EVIO}

Section~\ref{Tracking Performance in Aggressive Motion and HDR Scenarios} provides the analysis of the performance comparison between feature-based and direct-based EIO/EVIO, both in the same framework (Fig.~\ref{fig:boxes_6dof_comparison},~\ref{fig:hdr_poster_comparison}) and across different frameworks (Fig.~\ref{fig:hku_agg_walk_comparison},~\ref{fig:mountain_normal_comparison}, and Table~\ref{table:Localization_accuracy_comparison_1},~\ref{table:Localization_accuracy_comparison_2}).
In this APPNDIX, we further demonstrates more performance comparisons among different combinations within the same framework (our EVI-SAM).
We used the publicly available monocular HKU-Dataset~\cite{GWPHKU:MyEVIO} (346$\times$260, event, image, IMU, GT pose) and the DAVIS240C Dataset~\cite{GWPHKU:event-camera-dataset_davis240c} (240$\times$180, event, image, IMU, GT pose).
Since the DAVIS240C dataset typically exhibits limited displacement in the last 30 seconds, this is not conducive to explicitly evaluating the differences in position accuracy. 
Therefore, we edited the rosbag files to include only the [0-30s] segment of the total duration of the 60s.
The sequences in Table~\ref{table: Ablation Study of the feature-based and direct-based} are renamed with "edited" to distinguish them from the complete data in Table~\ref{table:Localization_accuracy_comparison_1}.

\begin{table}[htbp]
        \captionsetup{justification=justified}
        \setlength{\abovecaptionskip}{-0.02cm}
        \renewcommand\arraystretch{1.2}
        \tiny 
        \begin{center}
        \caption{Accuracy comparison of various combinations within EVI-SAM framework, including feature-based, direct-based, and our hybrid (feature-based + direct-based) tracking pipeline}
        \label{table: Ablation Study of the feature-based and direct-based}
        \resizebox{\columnwidth*1}{!}
        { 
        \begin{threeparttable}
        \begin{tabular}{c|ccc} 
        \hline  
        Sequence & \textbf{Feature-based} & \textbf{Direct-based} & \makecell{\textbf{Hybrid} \\\textbf{EVI-SAM}}\\
        \hline
        vicon\_hdr4                  & 0.09 & 0.08 & 0.09\\
        vicon\_aggressive\_hdr       & 0.92 & 0.59 & 0.68\\
        boxes\_translation\_edited   & 0.15 & 0.10 & 0.14 \\
        hdr\_boxes\_edited           & 0.21 & 0.12 & 0.21  \\
        boxes\_6dof\_edited          & 0.34 & 0.16 & 0.20 \\
        dynamic\_translation\_edited & 0.40 & 0.32 & 0.34 \\
        dynamic\_6dof\_edited        & 0.19 & 0.18 & 0.23 \\
        poster\_translation\_edited  & 0.12 & 0.15 & 0.35 \\
        hdr\_poster\_edited          & 0.55 & 0.32 & 0.44 \\
        poster\_6dof\_edited         & 0.10 & 0.08 & 0.09 \\
        \hline        
        \end{tabular}
        \begin{tablenotes} 
        \item \textit{Unit:\%, 0.10 means the average error would be 0.10m for 100m motion; Aligning the whole ground truth trajectory with estimated poses.} 
        \end{tablenotes} 
        \end{threeparttable} 
        }
        \end{center}
        \vspace{-1.0em}
\end{table}

As can be seen from the result, the proposed direct-based pose tacking outperforms the feature-based component in most of the sequences.
This result is consistent with the theoretical analysis in Section~\ref{Hybrid Optimization for Feature-based EVIO and Event-based Direct Tracking}, as well as the common viewpoint in direct-based methods, i.e. sub-pixel alignment of the direct-based methods normally achieves better accuracy than the feature-based methods.
On the other hand, as what has been well evaluated in our previous works~\cite{GWPHKU:PL-EVIO,ESVIO}, our feature-based EVIO can operate robustly in large-scale outdoor environments and flight scenarios.
This proves that it is suitable for handling irregular scene variations and significant inter-frame motions.
Therefore, leveraging the complementary advantages of direct-based and feature-based methods is significant and promising.

\subsection{More Tracking Performance Comparison in Challenge Situations}
\label{appendices:More Tracking Performance Comparison with Image-based VIO in Challenge Situations}

Section~\ref{Tracking Performance in Aggressive Motion and HDR Scenarios} provides the performance comparison of our event-based hybrid pose tracking with the state-of-the-art EIO, VIO, and EVIO works in challenging situations.
We further conduct the comparison of our EVI-SAM with two latest works in the field of image-based VIO~\cite{envio,MSOC-S-IKF}.
We used the publicly available stereo HKU-dataset~\cite{ESVIO} which features rapid motion and HDR scenarios.
Following the trajectory alignment protocol of Table~\ref{table:Localization_accuracy_comparison_2}, we computed the mean position as percentages of the total traveled distance, while the estimated trajectories and ground-truth were aligned in SE3 with all alignments.

\begin{table}[htb]
        \renewcommand\arraystretch{1.2}
        \begin{center}
        \caption{Accuracy Comparison of Our EVI-SAM with image-based VIO on HKU-Dataset}
        \label{table:Localization_accuracy_comparison_3}
        \resizebox{\columnwidth}{!}
        { 
        \begin{threeparttable}
        \begin{tabular}{c|cccc} 
        \hline  
        \multicolumn{1}{c|}{Sequence} 
    &\makecell{EnVIO~\cite{envio} \\Stereo VIO} 
    &\makecell{MSOC-S-IKF~\cite{MSOC-S-IKF} \\Stereo VIO}
    &\makecell{MSOC-S-IKF~\cite{MSOC-S-IKF} \\Mono VIO}
    &\makecell{\textbf{Our EVI-SAM} \\\textbf{Mono EVIO}}\\  
\hline
hku\_agg\_translation            & \textit{failed} & \textit{failed} & 0.51 & \textbf{0.17} \\
hku\_agg\_rotation             & 2.12 & 1.52 & 3.50& \textbf{0.24}\\
hku\_agg\_flip           & 2.94& \textit{failed} & \textit{failed}& \textbf{0.32}\\
hku\_agg\_walk            & 3.38 & \textit{failed} & 2.07 &\textbf{0.26}\\
hku\_hdr\_circle    & 0.85 & \textit{failed} & \textit{failed} &\textbf{0.13}\\
hku\_hdr\_slow    & 0.43& \textit{failed} & \textit{failed}&\textbf{0.11}\\
hku\_hdr\_tran\_rota   & 0.37 & \textit{failed} & 5.48&\textbf{0.11}\\
hku\_hdr\_agg    & 0.50& \textit{failed} & \textit{failed} &\textbf{0.10}\\
hku\_dark\_normal           & \textit{failed}& \textit{failed} & \textit{failed} &\textbf{0.85}\\
\hline      
        \end{tabular}
        \begin{tablenotes} 
        \item \textit{Unit:\%, 0.17 means the average error would be 0.17m for 100m motion; Aligning the whole ground truth trajectory with estimated poses.} 
        \end{tablenotes} 
        \end{threeparttable} 
        }
        \end{center}\
        \vspace{-1.0em}
\end{table}

Table~\ref{table:Localization_accuracy_comparison_3} demonstrates that our EVI-SAM outperforms these image-based VIO methods in all sequences.
Given the challenge of these sequences, which involve rapid movement, low-light conditions, and significant illumination change, it is difficult for the traditional image-based methods~\cite{ORB-SLAM3,GWPHKU:VINS-Fusion,envio,MSOC-S-IKF} to handle such high-speed motions and HDR scenarios.
For instance, both monocular and stereo versions of the MSOC-S-IKF~\cite{MSOC-S-IKF} perform well at the beginning of most of these sequences.
However, when confronted with fast motion or sudden illumination changes, these VIO systems are prone to failure or significant drift due to the unreliable perception of images. 
Certainly, the performance may vary among different image-based VIO methods.
For example, although the results of ORB-SLAM3~\cite{ORB-SLAM3} in Table~\ref{table:Localization_accuracy_comparison_2} and the EnVIO~\cite{envio} also exhibit large errors under such challenging conditions, they tend to avoid serious drift or failure, possibly owing to differences in framework stability.

\end{appendices}


\bibliographystyle{IEEEtran} 
\bibliography{references.bib} 

\begin{thebibliography}{10}
\providecommand{\url}[1]{#1}
\csname url@rmstyle\endcsname
\providecommand{\newblock}{\relax}
\providecommand{\bibinfo}[2]{#2}
\providecommand\BIBentrySTDinterwordspacing{\spaceskip=0pt\relax}
\providecommand\BIBentryALTinterwordstretchfactor{4}
\providecommand\BIBentryALTinterwordspacing{\spaceskip=\fontdimen2\font plus
\BIBentryALTinterwordstretchfactor\fontdimen3\font minus \fontdimen4\font\relax}
\providecommand\BIBforeignlanguage[2]{{%
\expandafter\ifx\csname l@#1\endcsname\relax
\typeout{** WARNING: IEEEtran.bst: No hyphenation pattern has been}%
\typeout{** loaded for the language `#1'. Using the pattern for}%
\typeout{** the default language instead.}%
\else
\language=\csname l@#1\endcsname
\fi
#2}}

\bibitem{GWPHKU:EVENT-SURVEY}
G.~Gallego, T.~Delbr{\"u}ck, G.~Orchard, C.~Bartolozzi, B.~Taba, A.~Censi, S.~Leutenegger, A.~J. Davison, J.~Conradt, K.~Daniilidis, \emph{et~al.}, ``Event-based vision: A survey,'' \emph{IEEE transactions on pattern analysis and machine intelligence}, vol.~44, no.~1, pp. 154--180, 2020.

\bibitem{GWPHKU:EMVS}
H.~Rebecq, G.~Gallego, E.~Mueggler, and D.~Scaramuzza, ``Emvs: Event-based multi-view stereo—3d reconstruction with an event camera in real-time,'' \emph{International Journal of Computer Vision}, vol. 126, no.~12, pp. 1394--1414, 2018.

\bibitem{zhou2018semi}
Y.~Zhou, G.~Gallego, H.~Rebecq, L.~Kneip, H.~Li, and D.~Scaramuzza, ``Semi-dense 3d reconstruction with a stereo event camera,'' in \emph{Proceedings of the European conference on computer vision (ECCV)}, 2018, pp. 235--251.

\bibitem{ghosh2022multi}
S.~Ghosh and G.~Gallego, ``Multi-event-camera depth estimation and outlier rejection by refocused events fusion,'' \emph{Advanced Intelligent Systems}, vol.~4, no.~12, p. 2200221, 2022.

\bibitem{zhu2019unsupervised}
A.~Z. Zhu, L.~Yuan, K.~Chaney, and K.~Daniilidis, ``Unsupervised event-based learning of optical flow, depth, and egomotion,'' in \emph{Proceedings of the IEEE/CVF Conference on Computer Vision and Pattern Recognition}, 2019, pp. 989--997.

\bibitem{EventNeRF}
V.~Rudnev, M.~Elgharib, C.~Theobalt, and V.~Golyanik, ``Eventnerf: Neural radiance fields from a single colour event camera,'' in \emph{Proceedings of the IEEE/CVF Conference on Computer Vision and Pattern Recognition}, 2023, pp. 4992--5002.

\bibitem{tulyakov2019learning}
S.~Tulyakov, F.~Fleuret, M.~Kiefel, P.~Gehler, and M.~Hirsch, ``Learning an event sequence embedding for dense event-based deep stereo,'' in \emph{Proceedings of the IEEE/CVF International Conference on Computer Vision}, 2019, pp. 1527--1537.

\bibitem{nam2022stereo}
Y.~Nam, M.~Mostafavi, K.-J. Yoon, and J.~Choi, ``Stereo depth from events cameras: Concentrate and focus on the future,'' in \emph{Proceedings of the IEEE/CVF Conference on Computer Vision and Pattern Recognition}, 2022, pp. 6114--6123.

\bibitem{ahmed2021deep}
S.~H. Ahmed, H.~W. Jang, S.~N. Uddin, and Y.~J. Jung, ``Deep event stereo leveraged by event-to-image translation,'' in \emph{Proceedings of the AAAI Conference on Artificial Intelligence}, vol.~35, no.~2, 2021, pp. 882--890.

\bibitem{mostafavi2021event}
M.~Mostafavi, K.-J. Yoon, and J.~Choi, ``Event-intensity stereo: Estimating depth by the best of both worlds,'' in \emph{Proceedings of the IEEE/CVF International Conference on Computer Vision}, 2021, pp. 4258--4267.

\bibitem{GWPHKU:EVO}
H.~Rebecq, T.~Horstsch{\"a}fer, G.~Gallego, and D.~Scaramuzza, ``Evo: A geometric approach to event-based 6-dof parallel tracking and mapping in real time,'' \emph{IEEE Robotics and Automation Letters}, vol.~2, no.~2, pp. 593--600, 2016.

\bibitem{GWPHKU:ESVO}
Y.~Zhou, G.~Gallego, and S.~Shen, ``Event-based stereo visual odometry,'' \emph{IEEE Transactions on Robotics}, 2021.

\bibitem{EDS}
J.~Hidalgo-Carri{\'o}, G.~Gallego, and D.~Scaramuzza, ``Event-aided direct sparse odometry,'' in \emph{Proceedings of the IEEE/CVF Conference on Computer Vision and Pattern Recognition}, 2022, pp. 5781--5790.

\bibitem{GWPHKU:Ultimate-SLAM}
A.~R. Vidal, H.~Rebecq, T.~Horstschaefer, and D.~Scaramuzza, ``Ultimate slam? combining events, images, and imu for robust visual slam in hdr and high-speed scenarios,'' \emph{IEEE Robotics and Automation Letters}, vol.~3, no.~2, pp. 994--1001, 2018.

\bibitem{GWPHKU:PL-EVIO}
W.~Guan, P.~Chen, Y.~Xie, and P.~Lu, ``Pl-evio: Robust monocular event-based visual inertial odometry with point and line features,'' \emph{IEEE Transactions on Automation Science and Engineering}, 2023.

\bibitem{ESVIO}
P.~Chen, W.~Guan, and P.~Lu, ``Esvio: Event-based stereo visual inertial odometry,'' \emph{IEEE Robotics and Automation Letters}, vol.~8, no.~6, pp. 3661--3668, 2023.

\bibitem{GWPHKU:Event-based-visual-inertial-odometry}
A.~Zihao~Zhu, N.~Atanasov, and K.~Daniilidis, ``Event-based visual inertial odometry,'' in \emph{Proceedings of the IEEE Conference on Computer Vision and Pattern Recognition}, 2017, pp. 5391--5399.

\bibitem{GWPHKU:ETH-EVIO}
H.~Rebecq, T.~Horstschaefer, and D.~Scaramuzza, ``Real-time visual-inertial odometry for event cameras using keyframe-based nonlinear optimization,'' in \emph{British Machine Vision Conference (BMVC)}, 2017.

\bibitem{GWPHKU:Continuous-time-visual-inertial-odometry-for-event-cameras}
E.~Mueggler, G.~Gallego, H.~Rebecq, and D.~Scaramuzza, ``Continuous-time visual-inertial odometry for event cameras,'' \emph{IEEE Transactions on Robotics}, vol.~34, no.~6, pp. 1425--1440, 2018.

\bibitem{EKLT-VIO}
F.~Mahlknecht, D.~Gehrig, J.~Nash, F.~M. Rockenbauer, B.~Morrell, J.~Delaune, and D.~Scaramuzza, ``Exploring event camera-based odometry for planetary robots,'' \emph{IEEE Robotics and Automation Letters (RA-L)}, 2022.

\bibitem{GWPHKU:EKLT}
D.~Gehrig, H.~Rebecq, G.~Gallego, and D.~Scaramuzza, ``Eklt: Asynchronous photometric feature tracking using events and frames,'' \emph{International Journal of Computer Vision}, vol. 128, no.~3, pp. 601--618, 2020.

\bibitem{GWPHKU:MyEVIO}
W.~Guan and P.~Lu, ``Monocular event visual inertial odometry based on event-corner using sliding windows graph-based optimization,'' in \emph{2022 IEEE/RSJ International Conference on Intelligent Robots and Systems (IROS)}.\hskip 1em plus 0.5em minus 0.4em\relax IEEE, 2022, pp. 2438--2445.

\bibitem{IROS2022_EVIO}
B.~Dai, C.~L. Gentil, and T.~Vidal-Calleja, ``A tightly-coupled event-inertial odometry using exponential decay and linear preintegrated measurements,'' in \emph{2022 IEEE/RSJ International Conference on Intelligent Robots and Systems (IROS)}, 2022, pp. 9475--9482.

\bibitem{liu2022asynchronous}
D.~Liu, A.~Parra, Y.~Latif, B.~Chen, T.-J. Chin, and I.~Reid, ``Asynchronous optimisation for event-based visual odometry,'' in \emph{2022 International Conference on Robotics and Automation (ICRA)}.\hskip 1em plus 0.5em minus 0.4em\relax IEEE, 2022, pp. 9432--9438.

\bibitem{GWPHKU:IDOL}
C.~Le~Gentil, F.~Tschopp, I.~Alzugaray, T.~Vidal-Calleja, R.~Siegwart, and J.~Nieto, ``Idol: A framework for imu-dvs odometry using lines,'' in \emph{2020 IEEE/RSJ International Conference on Intelligent Robots and Systems (IROS)}.\hskip 1em plus 0.5em minus 0.4em\relax IEEE, 2020, pp. 5863--5870.

\bibitem{chamorro2023event}
W.~Chamorro, J.~Sol{\`a}, and J.~Andrade-Cetto, ``Event-imu fusion strategies for faster-than-imu estimation throughput,'' in \emph{Proceedings of the IEEE/CVF Conference on Computer Vision and Pattern Recognition}, 2023, pp. 3975--3982.

\bibitem{chamorro2022event}
W.~Chamorro, J.~Sol{\`a}, and J.~Andrade-Cetto, ``Event-based line slam in real-time,'' \emph{IEEE Robotics and Automation Letters}, vol.~7, no.~3, pp. 8146--8153, 2022.

\bibitem{GWPHKU:kim2016real}
H.~Kim, S.~Leutenegger, and A.~J. Davison, ``Real-time 3d reconstruction and 6-dof tracking with an event camera,'' in \emph{European Conference on Computer Vision}.\hskip 1em plus 0.5em minus 0.4em\relax Springer, 2016, pp. 349--364.

\bibitem{gallego2017event}
G.~Gallego, J.~E. Lund, E.~Mueggler, H.~Rebecq, T.~Delbruck, and D.~Scaramuzza, ``Event-based, 6-dof camera tracking from photometric depth maps,'' \emph{IEEE transactions on pattern analysis and machine intelligence}, vol.~40, no.~10, pp. 2402--2412, 2017.

\bibitem{bryner2019event}
S.~Bryner, G.~Gallego, H.~Rebecq, and D.~Scaramuzza, ``Event-based, direct camera tracking from a photometric 3d map using nonlinear optimization,'' in \emph{2019 International Conference on Robotics and Automation (ICRA)}.\hskip 1em plus 0.5em minus 0.4em\relax IEEE, 2019, pp. 325--331.

\bibitem{DEVO}
Y.-F. Zuo, J.~Yang, J.~Chen, X.~Wang, Y.~Wang, and L.~Kneip, ``Devo: Depth-event camera visual odometry in challenging conditions,'' in \emph{2022 International Conference on Robotics and Automation (ICRA)}.\hskip 1em plus 0.5em minus 0.4em\relax IEEE, 2022, pp. 2179--2185.

\bibitem{dong2021standard}
Y.~Dong, ``Standardand event cameras fusion for dense mapping,'' \emph{arXiv preprint arXiv:2102.03567}, 2021.

\bibitem{EOMVS}
H.~Cho, J.~Jeong, and K.-J. Yoon, ``Eomvs: Event-based omnidirectional multi-view stereo,'' \emph{IEEE Robotics and Automation Letters}, vol.~6, no.~4, pp. 6709--6716, 2021.

\bibitem{gallego2018unifying}
G.~Gallego, H.~Rebecq, and D.~Scaramuzza, ``A unifying contrast maximization framework for event cameras, with applications to motion, depth, and optical flow estimation,'' in \emph{Proceedings of the IEEE conference on computer vision and pattern recognition}, 2018, pp. 3867--3876.

\bibitem{chiavazza2023low}
S.~Chiavazza, S.~M. Meyer, and Y.~Sandamirskaya, ``Low-latency monocular depth estimation using event timing on neuromorphic hardware,'' in \emph{Proceedings of the IEEE/CVF Conference on Computer Vision and Pattern Recognition}, 2023, pp. 4070--4079.

\bibitem{chaney2019learning}
K.~Chaney, A.~Zihao~Zhu, and K.~Daniilidis, ``Learning event-based height from plane and parallax,'' in \emph{Proceedings of the IEEE/CVF Conference on Computer Vision and Pattern Recognition Workshops}, 2019, pp. 0--0.

\bibitem{hidalgo2020learning}
J.~Hidalgo-Carri{\'o}, D.~Gehrig, and D.~Scaramuzza, ``Learning monocular dense depth from events,'' in \emph{2020 International Conference on 3D Vision (3DV)}.\hskip 1em plus 0.5em minus 0.4em\relax IEEE, 2020, pp. 534--542.

\bibitem{gehrig2021combining}
D.~Gehrig, M.~R{\"u}egg, M.~Gehrig, J.~Hidalgo-Carri{\'o}, and D.~Scaramuzza, ``Combining events and frames using recurrent asynchronous multimodal networks for monocular depth prediction,'' \emph{IEEE Robotics and Automation Letters}, vol.~6, no.~2, pp. 2822--2829, 2021.

\bibitem{Ev-NeRF}
I.~Hwang, J.~Kim, and Y.~M. Kim, ``Ev-nerf: Event based neural radiance field,'' in \emph{Proceedings of the IEEE/CVF Winter Conference on Applications of Computer Vision}, 2023, pp. 837--847.

\bibitem{E-nerf}
S.~Klenk, L.~Koestler, D.~Scaramuzza, and D.~Cremers, ``E-nerf: Neural radiance fields from a moving event camera,'' \emph{IEEE Robotics and Automation Letters}, vol.~8, no.~3, pp. 1587--1594, 2023.

\bibitem{mahbub2023multimodal}
S.~Mahbub, B.~Feng, and C.~Metzler, ``Multimodal neural surface reconstruction: Recovering the geometry and appearance of 3d scenes from events and grayscale images,'' in \emph{NeurIPS 2023 Workshop on Deep Learning and Inverse Problems}, 2023.

\bibitem{ieng2018neuromorphic}
S.-H. Ieng, J.~Carneiro, M.~Osswald, and R.~Benosman, ``Neuromorphic event-based generalized time-based stereovision,'' \emph{Frontiers in Neuroscience}, vol.~12, p. 442, 2018.

\bibitem{T-ESVO}
Z.~Liu, D.~Shi, R.~Li, Y.~Zhang, and S.~Yang, ``T-esvo: Improved event-based stereo visual odometry via adaptive time-surface and truncated signed distance function,'' \emph{Advanced Intelligent Systems}, p. 2300027, 2023.

\bibitem{ghosh2022event}
S.~Ghosh and G.~Gallego, ``Event-based stereo depth estimation from ego-motion using ray density fusion,'' \emph{European Conf. on Computer Vision Workshops (ECCVW) Ego4D}, 2022.

\bibitem{GWPHKU:VINS-MONO}
T.~Qin, P.~Li, and S.~Shen, ``Vins-mono: A robust and versatile monocular visual-inertial state estimator,'' \emph{IEEE Transactions on Robotics}, vol.~34, no.~4, pp. 1004--1020, 2018.

\bibitem{GWPHKU:VINS-MONO-initialization}
T.~Qin and S.~Shen, ``Robust initialization of monocular visual-inertial estimation on aerial robots,'' in \emph{2017 IEEE/RSJ International Conference on Intelligent Robots and Systems (IROS)}.\hskip 1em plus 0.5em minus 0.4em\relax IEEE, 2017, pp. 4225--4232.

\bibitem{dso}
J.~Engel, V.~Koltun, and D.~Cremers, ``Direct sparse odometry,'' \emph{IEEE transactions on pattern analysis and machine intelligence}, vol.~40, no.~3, pp. 611--625, 2017.

\bibitem{event-slam-survey}
K.~Huang, S.~Zhang, J.~Zhang, and D.~Tao, ``Event-based simultaneous localization and mapping: A comprehensive survey,'' \emph{arXiv preprint arXiv:2304.09793}, 2023.

\bibitem{svo}
C.~Forster, Z.~Zhang, M.~Gassner, M.~Werlberger, and D.~Scaramuzza, ``Svo: Semidirect visual odometry for monocular and multicamera systems,'' \emph{IEEE Transactions on Robotics}, vol.~33, no.~2, pp. 249--265, 2016.

\bibitem{envio}
J.~H. Jung, Y.~Choe, and C.~G. Park, ``Photometric visual-inertial navigation with uncertainty-aware ensembles,'' \emph{IEEE Transactions on Robotics}, vol.~38, no.~4, pp. 2039--2052, 2022.

\bibitem{sibley2010sliding}
G.~Sibley, L.~Matthies, and G.~Sukhatme, ``Sliding window filter with application to planetary landing,'' \emph{Journal of field robotics}, vol.~27, no.~5, pp. 587--608, 2010.

\bibitem{leutenegger2013keyframe}
S.~Leutenegger, P.~Furgale, V.~Rabaud, M.~Chli, K.~Konolige, and R.~Siegwart, ``Keyframe-based visual-inertial slam using nonlinear optimization,'' \emph{Proceedings of Robotis Science and Systems (RSS) 2013}, 2013.

\bibitem{space-sweep}
R.~T. Collins, ``A space-sweep approach to true multi-image matching,'' in \emph{Proceedings CVPR IEEE Computer Society Conference on Computer Vision and Pattern Recognition}.\hskip 1em plus 0.5em minus 0.4em\relax Ieee, 1996, pp. 358--363.

\bibitem{ma2019sparse}
F.~Ma, L.~Carlone, U.~Ayaz, and S.~Karaman, ``Sparse depth sensing for resource-constrained robots,'' \emph{The International Journal of Robotics Research}, vol.~38, no.~8, pp. 935--980, 2019.

\bibitem{zhang2018probability}
H.-T. Zhang, J.~Yu, and Z.-F. Wang, ``Probability contour guided depth map inpainting and superresolution using non-local total generalized variation,'' \emph{Multimedia Tools and Applications}, vol.~77, pp. 9003--9020, 2018.

\bibitem{telea2004image}
A.~Telea, ``An image inpainting technique based on the fast marching method,'' \emph{Journal of graphics tools}, vol.~9, no.~1, pp. 23--34, 2004.

\bibitem{shenshaojie:quadtree}
K.~Wang, W.~Ding, and S.~Shen, ``Quadtree-accelerated real-time monocular dense mapping,'' in \emph{2018 IEEE/RSJ International Conference on Intelligent Robots and Systems (IROS)}.\hskip 1em plus 0.5em minus 0.4em\relax IEEE, 2018, pp. 1--9.

\bibitem{voxblox}
H.~Oleynikova, Z.~Taylor, M.~Fehr, R.~Siegwart, and J.~Nieto, ``Voxblox: Incremental 3d euclidean signed distance fields for on-board mav planning,'' in \emph{2017 IEEE/RSJ International Conference on Intelligent Robots and Systems (IROS)}.\hskip 1em plus 0.5em minus 0.4em\relax IEEE, 2017, pp. 1366--1373.

\bibitem{GWPHKU:event-camera-dataset_davis240c}
E.~Mueggler, H.~Rebecq, G.~Gallego, T.~Delbruck, and D.~Scaramuzza, ``The event-camera dataset and simulator: Event-based data for pose estimation, visual odometry, and slam,'' \emph{The International Journal of Robotics Research}, vol.~36, no.~2, pp. 142--149, 2017.

\bibitem{HASTE-VIO}
I.~Alzugaray and M.~Chli, ``Asynchronous multi-hypothesis tracking of features with event cameras,'' in \emph{2019 International Conference on 3D Vision (3DV)}.\hskip 1em plus 0.5em minus 0.4em\relax IEEE, 2019, pp. 269--278.

\bibitem{dai2022tightly}
B.~Dai, C.~Le~Gentil, and T.~Vidal-Calleja, ``A tightly-coupled event-inertial odometry using exponential decay and linear preintegrated measurements,'' in \emph{2022 IEEE/RSJ International Conference on Intelligent Robots and Systems (IROS)}.\hskip 1em plus 0.5em minus 0.4em\relax IEEE, 2022, pp. 9475--9482.

\bibitem{lee2023event}
M.~S. Lee, J.~H. Jung, Y.~J. Kim, and C.~G. Park, ``Event-and frame-based visual-inertial odometry with adaptive filtering based on 8-dof warping uncertainty,'' \emph{IEEE Robotics and Automation Letters}, 2023.

\bibitem{ORB-SLAM3}
C.~Campos, R.~Elvira, J.~J.~G. Rodr{\'\i}guez, J.~M. Montiel, and J.~D. Tard{\'o}s, ``Orb-slam3: An accurate open-source library for visual, visual--inertial, and multimap slam,'' \emph{IEEE Transactions on Robotics}, 2021.

\bibitem{GWPHKU:VINS-Fusion}
T.~Qin, J.~Pan, S.~Cao, and S.~Shen, ``A general optimization-based framework for local odometry estimation with multiple sensors,'' \emph{arXiv preprint arXiv:1901.03638}, 2019.

\bibitem{GWPHKU:VECtor}
L.~Gao, Y.~Liang, J.~Yang, S.~Wu, C.~Wang, J.~Chen, and L.~Kneip, ``Vector: A versatile event-centric benchmark for multi-sensor slam,'' \emph{IEEE Robotics and Automation Letters}, 2022.

\bibitem{GWPHKU:MVSEC}
A.~Z. Zhu, D.~Thakur, T.~{\"O}zaslan, B.~Pfrommer, V.~Kumar, and K.~Daniilidis, ``The multivehicle stereo event camera dataset: An event camera dataset for 3d perception,'' \emph{IEEE Robotics and Automation Letters}, vol.~3, no.~3, pp. 2032--2039, 2018.

\bibitem{GWPHKU:DSEC}
M.~Gehrig, W.~Aarents, D.~Gehrig, and D.~Scaramuzza, ``Dsec: A stereo event camera dataset for driving scenarios,'' \emph{IEEE Robotics and Automation Letters}, vol.~6, no.~3, pp. 4947--4954, 2021.

\bibitem{GTS}
S.-H. Ieng, J.~Carneiro, M.~Osswald, and R.~Benosman, ``Neuromorphic event-based generalized time-based stereovision,'' \emph{Frontiers in neuroscience}, vol.~12, p. 442, 2018.

\bibitem{SGM}
H.~Hirschmuller, ``Stereo processing by semiglobal matching and mutual information,'' \emph{IEEE Transactions on pattern analysis and machine intelligence}, vol.~30, no.~2, pp. 328--341, 2007.

\bibitem{GWPHKU:ECMD}
P.~Chen, W.~Guan, F.~Huang, Y.~Zhong, W.~Wen, L.-T. Hsu, and P.~Lu, ``Ecmd: An event-centric multisensory driving dataset for slam,'' \emph{IEEE Transactions on Intelligent Vehicles}, 2023.

\bibitem{choi2014consensus}
O.~Choi and S.-W. Jung, ``A consensus-driven approach for structure and texture aware depth map upsampling,'' \emph{IEEE transactions on image processing}, vol.~23, no.~8, pp. 3321--3335, 2014.

\bibitem{xiang2015no}
S.~Xiang, L.~Yu, and C.~W. Chen, ``No-reference depth assessment based on edge misalignment errors for t+ d images,'' \emph{IEEE Transactions on Image Processing}, vol.~25, no.~3, pp. 1479--1494, 2015.

\bibitem{image_undistort}
Z.~Taylor, ``A compact package for undistorting images directly from kalibr calibration files,'' \url{https://github.com/ethz-asl/image\_undistort}.

\bibitem{intelrealsense}
``intelrealsense,'' \url{https://www.intelrealsense.com/depth-camera-d455/}.

\bibitem{instant-ngp}
T.~M{\"u}ller, A.~Evans, C.~Schied, and A.~Keller, ``Instant neural graphics primitives with a multiresolution hash encoding,'' \emph{ACM transactions on graphics (TOG)}, vol.~41, no.~4, pp. 1--15, 2022.

\bibitem{COLMAP}
J.~L. Schonberger and J.-M. Frahm, ``Structure-from-motion revisited,'' in \emph{Proceedings of the IEEE conference on computer vision and pattern recognition}, 2016, pp. 4104--4113.

\bibitem{Nice-slam}
Z.~Zhu, S.~Peng, V.~Larsson, W.~Xu, H.~Bao, Z.~Cui, M.~R. Oswald, and M.~Pollefeys, ``Nice-slam: Neural implicit scalable encoding for slam,'' in \emph{Proceedings of the IEEE/CVF Conference on Computer Vision and Pattern Recognition}, 2022, pp. 12\,786--12\,796.

\bibitem{Co-slam}
H.~Wang, J.~Wang, and L.~Agapito, ``Co-slam: Joint coordinate and sparse parametric encodings for neural real-time slam,'' in \emph{Proceedings of the IEEE/CVF Conference on Computer Vision and Pattern Recognition}, 2023, pp. 13\,293--13\,302.

\bibitem{MSOC-S-IKF}
Z.~Zhang, Y.~Song, S.~Huang, R.~Xiong, and Y.~Wang, ``Toward consistent and efficient map-based visual-inertial localization: Theory framework and filter design,'' \emph{IEEE Transactions on Robotics}, 2023.

\end{thebibliography}

\begin{IEEEbiography}[{\includegraphics[width=1in,height=1.25in,clip,keepaspectratio]{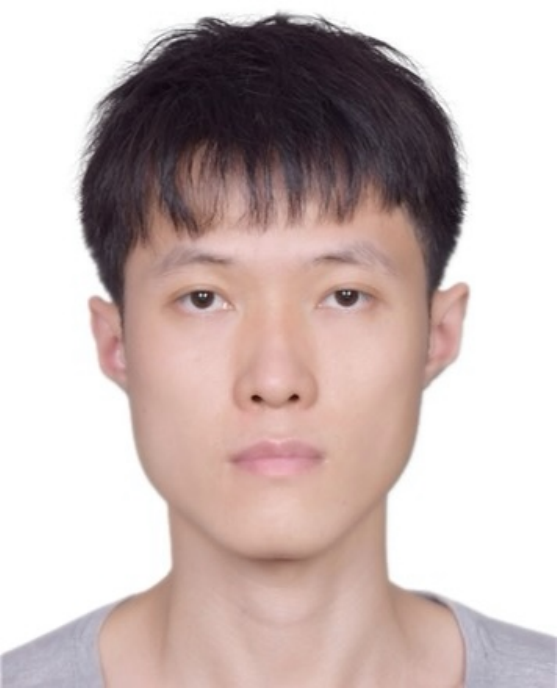}}]{Guan Weipeng}
obtained his Bachelor degree in Electronic Science \& Technology, as well as Master degree in Control Theory and Control Engineering from the South China University of Technology. 
He is currently pursuing his PhD degree in Robotics at the University of Hong Kong. 
He has worked with several reputable organizations, including: Samsung Electronics, Huawei Technologies, The Chinese Academy of Sciences, The Chinese University of Hong Kong, The Hong Kong University of Science and Technology, etc.  
He has also served as a technical consultant for multiple companies, such as TCL.
Moreover, he has published over 60 research articles in prestigious international journals and conferences, as well as holds more than 40 authorized patents. 
His research interests primarily focus on robotics, event-based VO/VIO/SLAM, visible light positioning, etc.
\end{IEEEbiography}

\begin{IEEEbiography}[{\includegraphics[width=1in,height=1.25in,clip,keepaspectratio]{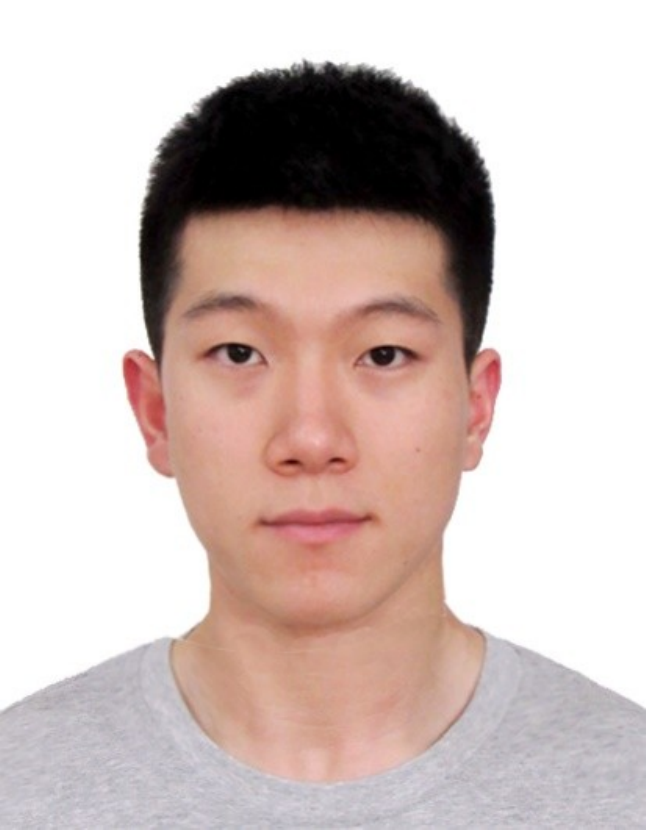}}]{Chen Peiyu}
received the BSc degrees in automation from the Nanjing University of Science and Technology, China, and MSc degrees in computer control \& automation from the Nanyang Technological University, Singapore, in 2020 and 2022, respectively. 
He is currently working forward the Ph.D. degree at the University of Hong Kong.  
His research interests include robotics, visual-inertial simultaneous localization and mapping, nonlinear control, and so on.
\end{IEEEbiography}

\begin{IEEEbiography}[{\includegraphics[width=1in,height=1.25in,clip,keepaspectratio]{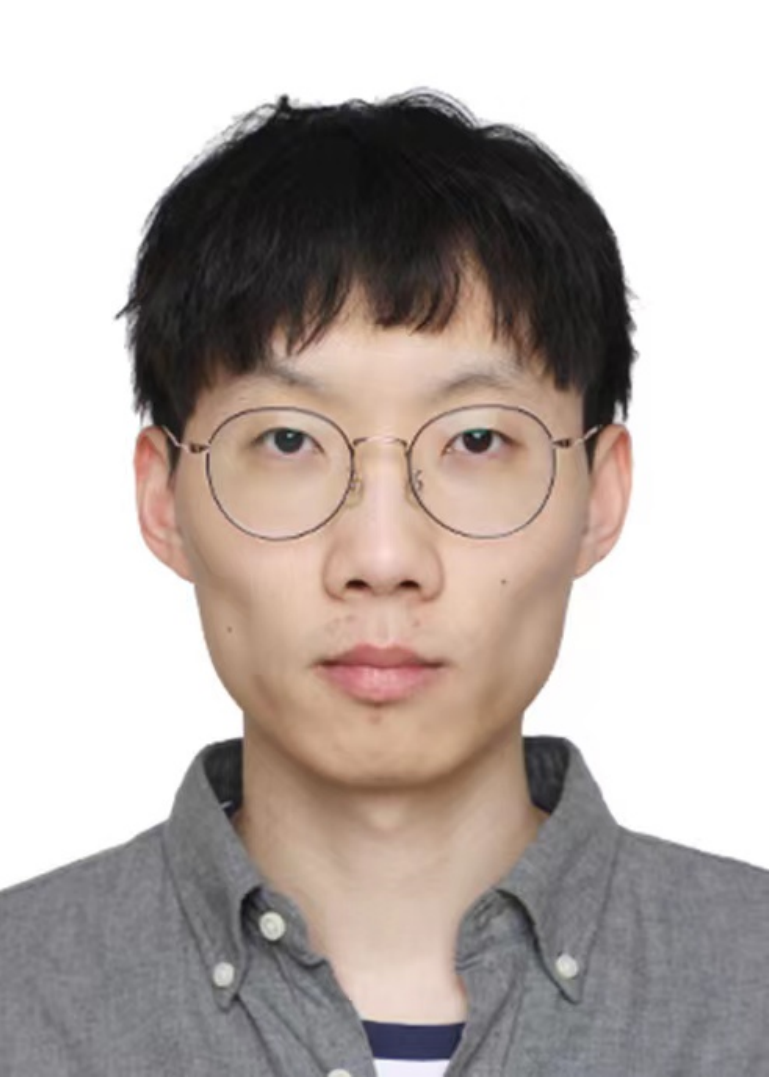}}]{Zhao Huibin}
received the BSc degrees in aerospace engineering from Beihang University, China, and MSc degrees in Robotics from the University of Bristol, UK, in 2018 and 2020, respectively. 
He is currently working forward the Ph.D. degree at the University of Hong Kong.  
His research interests include robotics, learning-based 3D reconstruction, and so on.

\end{IEEEbiography}

\begin{IEEEbiography}[{\includegraphics[width=1in,height=1.25in,clip,keepaspectratio]{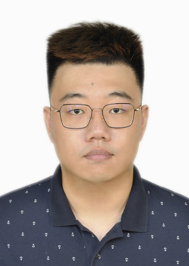}}]{Wang Yu}
received the BSc degrees in Mechanical Engineering from Auburn University, USA, and MSc degrees in Mechanical Engineering from the University of Hong Kong, in 2021 and 2023, respectively. 
He is currently working forward the MPhil degree at the University of Hong Kong.  

\end{IEEEbiography}

\begin{IEEEbiography}[{\includegraphics[width=1in,height=1.25in,clip,keepaspectratio]{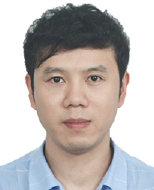}}]{Lu Peng}
obtained his BSc degree in automatic control and MSc degree in nonlinear flight control both from Northwestern Polytechnical University (NPU). He continued his journey on flight control at Delft University of Technology (TU Delft) where he received his PhD degree in 2016. After that, he shifted a bit from flight control and started to explore control for ground/construction robotics at ETH Zurich (ADRL lab) as a Postdoc researcher in 2016. He also had a short but nice journey at University of Zurich \& ETH Zurich (RPG group) where he was working on vision-based control for UAVs as a Postdoc researcher. He was an assistant professor in autonomous UAVs and robotics at Hong Kong Polytechnic University prior to joining the University of Hong Kong in 2020.

Prof. Lu has received several awards such as 3rd place in 2019 IROS autonomous drone racing competition and best graduate student paper finalist in AIAA GNC (top conference in aerospace). He serves as an associate editor for 2020 IROS (top conference in robotics) and session chair/co-chair for conferences like IROS and AIAA GNC for several times. He also gave a number of invited/keynote speeches at multiple conferences, universities and research institutes.
\end{IEEEbiography}

\vfill 
\end{document}